\documentclass[letterpaper]{article} 
\usepackage{aaai25}  
\usepackage{times}  
\usepackage{helvet}  
\usepackage{courier}  
\usepackage[hyphens]{url}  
\usepackage{graphicx} 
\usepackage{natbib}  
\usepackage{caption} 
\usepackage{algorithm}
\usepackage{algorithmic}

\usepackage{amsmath,amssymb,amsthm,amsfonts}
\usepackage{mathtools}
\usepackage{booktabs}
\usepackage{subfigure}
\usepackage{makecell}
\usepackage{multirow}
\usepackage{float}

\def\cL{{\cal L}}

\def\cL{{\cal L}}

\newcommand{\by}{{\bf y}}

\newcommand{\bx}{{\bf x}}

\newcommand{\bm}{{\bf m}}

\newcommand{\bbr}{{\bf r}}

\newcommand{\mbR}{\mathbb{R}}
\newcommand{\mbE}{\mathbb{E}}

\newcommand{\bc}{\begin{center}}
\newcommand{\ec}{\end{center}}
\newcommand{\be}{\begin{equation}}
\newcommand{\ee}{\end{equation}}
\newcommand{\ba}{\begin{array}}
\newcommand{\ea}{\end{array}}
\newcommand{\bean}{\begin{eqnarray*}}
\newcommand{\eean}{\end{eqnarray*}}
\newcommand{\bea}{\begin{eqnarray}}
\newcommand{\eea}{\end{eqnarray}}
\newcommand{\ben}{\begin{enumerate}}
\newcommand{\een}{\end{enumerate}}
\newcommand{\bed}{\begin{itemize}}
\newcommand{\eed}{\end{itemize}}

\newcommand{\indep}{\perp \!\!\! \perp}

\newtheorem{prop}{Proposition}
\newtheorem{definition}{Definition}
\newtheorem{assumption}{Assumption}
\newtheorem{remark}{Remark}

\usepackage{newfloat}
\usepackage{listings}
\DeclareCaptionStyle{ruled}{labelfont=normalfont,labelsep=colon,strut=off} 
\floatstyle{ruled}
\newfloat{listing}{tb}{lst}{}
\floatname{listing}{Listing}
\title{
Masked Language Modeling Becomes Conditional Density Estimation for Tabular Data Synthesis
}
\author {
    Seunghwan An\textsuperscript{\rm 1},
    Gyeongdong Woo\textsuperscript{\rm 1},
    Jaesung Lim\textsuperscript{\rm 1},\\
    ChangHyun Kim\textsuperscript{\rm 1},
    Sungchul Hong\textsuperscript{\rm 2},
    Jong-June Jeon\textsuperscript{\rm 2}\thanks{Corresponding author.}
}
\affiliations {
    \textsuperscript{\rm 1}Department of Statistical Data Science, University of Seoul, S. Korea\\
    \textsuperscript{\rm 2}Department of Statistics, University of Seoul, S. Korea\\
    dkstmdghks79@uos.ac.kr, dngudxor23@uos.ac.kr, wotjd1410@uos.ac.kr,\\
    hahaha503@uos.ac.kr, shong@uos.ac.kr, jj.jeon@uos.ac.kr
}

\begin{document}

\maketitle

\begin{abstract}
In this paper, our goal is to generate synthetic data for heterogeneous (mixed-type) tabular datasets with high machine learning utility (MLu). Since the MLu performance depends on accurately approximating the conditional distributions, we focus on devising a synthetic data generation method based on conditional distribution estimation. We introduce MaCoDE by redefining the consecutive multi-class classification task of Masked Language Modeling (MLM) as histogram-based non-parametric conditional density estimation. Our approach enables the estimation of conditional densities across arbitrary combinations of target and conditional variables. We bridge the theoretical gap between distributional learning and MLM by demonstrating that minimizing the orderless multi-class classification loss leads to minimizing the total variation distance between conditional distributions. To validate our proposed model, we evaluate its performance in synthetic data generation across 10 real-world datasets, demonstrating its ability to adjust data privacy levels easily without re-training. Additionally, since masked input tokens in MLM are analogous to missing data, we further assess its effectiveness in handling training datasets with missing values, including multiple imputations of the missing entries.
\end{abstract}

%

\section{Introduction}

There are two main objectives in synthetic tabular data generation: (1) preserving the statistical characteristics of the original dataset and (2) achieving comparable machine learning utility (MLu) to the original dataset. In this paper, our focus is on generating synthetic data with high MLu performance. Note that achieving high statistical fidelity does not guarantee high MLu performance \cite{hansen2023reimagining}. 

Given that MLu performance depends on accurately approximating conditional distributions, we focus on developing a synthetic data generation method based on conditional distribution estimation. However, two important properties of tabular data must be considered: ($i$) tabular data can consist of mixed types of data \cite{Borisov2021DeepNN, SHWARTZZIV202284}, and ($ii$) the tabular data does not have an intrinsic ordering among columns \cite{gulati2023tabmt}.

($i$) 
Considering the heterogeneous nature of tabular data and aiming to develop a method that addresses the challenges of modeling diverse distributions of continuous columns, we employ histogram-based non-parametric conditional density estimation through a multi-class classification task \cite{Li2019DeepDR}. This approach enables us to apply the classification loss uniformly across all types of columns. Since the histogram-based approach is theoretically valid only when continuous variables have bounded supports \cite{wasserman2006all, Li2019DeepDR}, we transform continuous columns using the Cumulative Distribution Function (CDF) and constrain their values to the interval $[0, 1]$ \cite{Li2021ImprovingGW, Fang2022OvercomingCO}.

($ii$)
To learn the arbitrary generation ordering of columns, we utilize the Masked Language Modeling (MLM) approach \cite{Devlin2019BERTPO}. By employing a masking scheme and the BERT model architecture, our proposed model enables the estimation of conditional densities across arbitrary combinations of target and conditional variables \cite{Ghazvininejad2019MaskPredictPD, ivanov2018variational, nazabal2020hivae}. \cite{gulati2023tabmt} proposed a similar method called TabMT, however, TabMT faces challenges in distributional learning because it relies on predicting the K-means cluster index of masked entries. Our approach contrasts with existing auto-regressive density estimators, which generate data in a fixed column order \cite{Hansen1994AutoregressiveCD, Kamthe2021CopulaFF, Letizia2022CopulaDN}. Additionally, \cite{germain2015made, papamakarios2017masked} are also able to estimate conditional densities but differ from our approach by masking the model weights rather than the input.

Therefore, our proposed method redefines the consecutive multi-class classification task of MLM as histogram-based non-parametric conditional density estimation. We term our proposed model MaCoDE (\underline{Ma}sked \underline{Co}nditional \underline{D}ensity \underline{E}stimation). The main contribution of our work is bridging the theoretical gap between distributional learning and the consecutive minimization of the multi-class classification loss within the MLM approach. 

Specifically, we demonstrate that minimizing the \textit{orderless} multi-class classification loss, when combined with the CDF transformation, provides theoretical validity for minimizing the discrepancy between conditional distributions in terms of total variation distance. This implies that we do not need to consider the ordering of bins, which could otherwise serve as a useful inductive bias. Note that, in the natural language domain, previous attempts to interpret MLM as distributional learning have been somewhat limited, relying on pseudo-likelihood or Markov random fields \cite{Ghazvininejad2019MaskPredictPD, wang2019bert, Salazar2019MaskedLM, ng2020ssmba, TorrobaHennigen2023DerivingLM}.

We substantiate the effectiveness of our proposed method by evaluating its performance in synthetic data generation across 10 real-world tabular datasets, demonstrating its capability to adjust data privacy levels easily without re-training. Given that masked input tokens in MLM can be viewed as missing data, we also assess our model’s effectiveness in handling training datasets with missing values. Moreover, as our proposed model estimates the conditional distribution while accommodating arbitrary conditioning sets, it can address various missingness patterns - an essential capability for generating samples and performing multiple imputations \cite{Buuren2012FlexibleIO, ivanov2018variational, nazabal2020hivae}. Consequently, we further validate our method's effectiveness by evaluating its performance in multiple imputations across various missing data mechanisms.

\section{Related Works}

Existing methods using deep generative models aim to directly minimize the discrepancy between the multivariate ground-truth distribution and the generative model. These include CTGAN \cite{xu2019ctgan}, TVAE \cite{xu2019ctgan}, CTAB-GAN \cite{zhao2021ctabgan}, CTAB-GAN+ \cite{Zhao2023CTABGANET}, DistVAE \cite{an2023distributional}, and TabDDPM \cite{kotelnikov2023tabddpm}. 

In the realm of transformer-based synthesizers, methods such as TabPFGen \cite{ma2023tabpfgen}, TabMT \cite{gulati2023tabmt}, and REaLTabFormer \cite{Solatorio2023REaLTabFormerGR} have been proposed. TabMT utilizes an MLM-based approach to generate synthetic data by predicting cluster indices from K-means clustering. TabPFGen is an energy-based model that leverages the Bayesian inference of TabPFN \cite{muller2022transformers} framework. REaLTabFormer employs an autoregressive Transformer architecture akin to GPT-2 \cite{radford2019language}, applying natural language generation techniques to tabular data. Additionally, methods such as \cite{Kamthe2021CopulaFF, Letizia2022CopulaDN, Drouin2022TACTiSTC, Ashok2023TACTiS2BF} use transformers for copula density estimation, specifically in time-series datasets.



\section{Proposal}
\label{sec:proposal}

\begin{figure}
    \centering
    \includegraphics[width=0.9\linewidth]{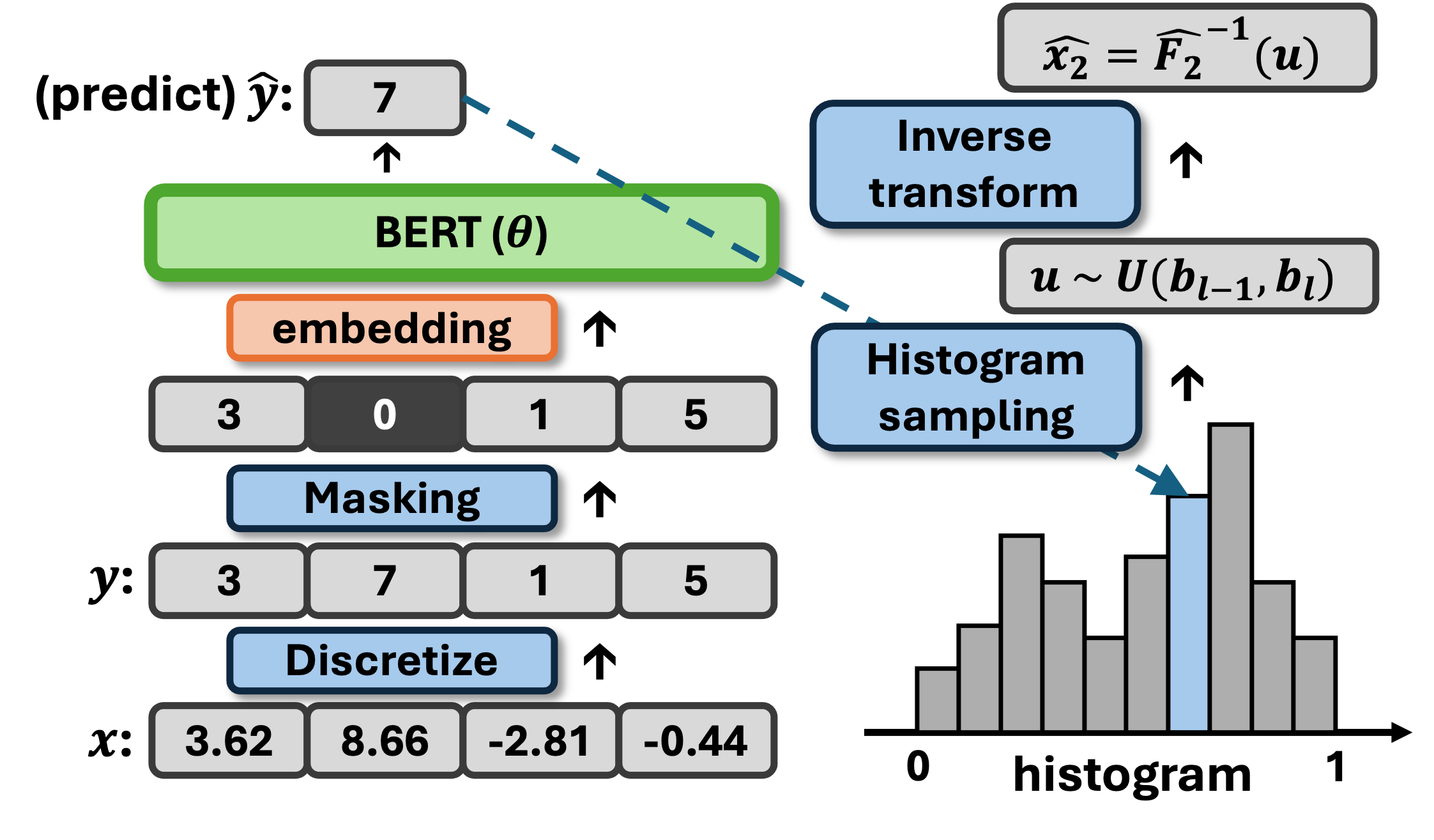}
    \caption{Overall structure of MaCoDE. In this case, the value of the second column is masked (replaced with `0') and predicted.}
    \label{fig:overall}
\end{figure}


\textbf{Notations.} 
Let $\bx \in \mbR^p$ denote an observation consisting of continuous and categorical (discrete) variables, and the $j$th variable (column) is denoted as $\bx_j$. Here, subscript $j$ refers to the $j$th element. $I_C$ and $I_D$ represent the index sets for continuous and categorical variables, where $I_C \cup I_D = \{1,\cdots,p\}$. The observed dataset is denoted as $\{\bx^{(i)}\}_{i=1}^n$. $\bm \in \{0, 1\}^p$ is a binary vector indicating masked values, with $\bm_j = 0$ indicating the $j$th column is masked (if $\bm_j=1$, then the $j$th column is not masked). 
$F^*_j$ indicates the ground-truth CDF of the $j$th column, and $\hat{F}_j$ is an estimator of $F^*_j$. 

\textbf{Overview.}
Without loss of generality, we can consider an arbitrary conditional density function: for $j \in I_C$,
\bea \label{eq:primary}
p_j^*(\bx_j | \bx_1, \cdots, \bx_{j-1}).
\eea
Our primary objective is to estimate \eqref{eq:primary}. However, there are two major challenges in estimating \eqref{eq:primary}:
\ben
    \item Modeling non-uniform distributions of continuous columns, $\bx_j$.
    \item Handling arbitrary combinations of conditional variables, $\bx_1, \cdots, \bx_{j-1}$.
\een

\begin{assumption} \label{assump:CDF}
For all $j \in I_C$, there exists $\hat{F}_j: \mbR \mapsto [0, 1]$ such that $\hat{F}_j$ is invertible and differentiable.
\end{assumption}

In this paper, we address these major challenges by unifying a histogram-based conditional density estimation and the MLM approach. And we assume that $\hat{F}_j$ satisfying Assumption \ref{assump:CDF} is given. For $j \in I_C$,
\bea \label{eq:changeofvariable}
p_j^*(\bx_j | \bx_{-j}) &=& c^*_j(\hat{F}_j(\bx_j) | \bx_{-j}) \cdot \hat{p}_j(\bx_j)
\eea
by the change of variable in terms of $\hat{F}_j$, where $c^*_j$ is the conditional density of $\hat{F}_j(\bx_j)$, $\bx_{-j} \coloneqq (\bx_1, \cdots, \bx_{j-1})$, and $\hat{p}_j$ denotes the density of $\hat{F}_j$. In particular, we estimate $c^*_j$ in \eqref{eq:changeofvariable} using a histogram-based approach.

\subsection{Classification Target (Discretization)}

The discretization is essential since the MLM-based approach hinges on multi-class classification tasks. In this section, we will describe how to transform the columns of a tabular dataset into classification targets. We denote the classification target (i.e., label) of the $j$th column as $\by_j$.

\textbf{Continuous column.} 
Firstly, since the histogram-based approach requires continuous columns to have bounded supports, we transform them into random variables within the $[0, 1]$ range using their marginal CDFs. Then, we partition the $[0, 1]$ interval with $L+1$ cut-points, $b_0, b_1, \cdots, b_L$, where $0 = b_0 < b_1 < b_2 < \cdots < b_{L-1} < b_L = 1$, resulting in $L$ bins. We define the classification target $\by_{j}$ based on the interval within which $\hat{F}_j(\bx_{j})$ falls, as follows:
\bean
\by_j \coloneqq \sum_{s=0}^{L-1} \mathbb{I}\Big(b_{s} \leq \hat{F}_j(\bx_j)\Big),
\eean
indicating that the classification target is the bin index, where $\mathbb{I}(\cdot)$ represents the indicator function, and $\by_{j} \in [L] \coloneqq \{1,2,\cdots,L\}$. Thus, if $b_{l-1} \leq \hat{F}_j(\bx_{j}) < b_l$, then $\by_j = l$. 

Our discretization procedure presents an advantage: since the lower and upper bounds are naturally defined as 0 and 1 within our approach, determining the number of bins becomes more intuitive. For example, finer bins near the boundaries of 0 and 1 can be employed to achieve more accurate tail density estimation.


\textbf{Categorical column.}
For categorical variables, without transformation procedures like those for continuous columns, we define $\by_j \coloneqq \bx_j$ for $j \in I_D$. To simplify notation, we denote the number of levels for all categorical variables as $L$. Allowing categorical variables with different numbers of categories is straightforward.

\begin{definition} \label{def:preprocess}
The discretization function $g: \mbR^p \mapsto [L]^p$ is defined as 
\bean
g(\bx; \hat{F})_j &\coloneqq& \begin{cases}
\sum_{s=0}^{L-1} \mathbb{I}(b_{s} \leq \hat{F}_j(\bx_j)), & \text{if $j \in I_C$}\\
\bx_j, & \text{if $j \in I_D$}
\end{cases},
\eean
for $j=1,\cdots,p$, where $\hat{F} \coloneqq \{\hat{F}_j: j \in I_C\}$.
\end{definition}

Finally, we transform the observations $\bx$ into the discretized label $\by$ using the discretization function $g$ of Definition \ref{def:preprocess}, as $\by_j = g(\bx; \hat{F})_j$ for all $j$.

\subsection{Masked Conditional Density Estimation}
\label{sec:cde}

\textbf{Target distribution.}
For all $j \in \{1,2,\cdots,p\}$, the ground-truth conditional probability of $g(\bx; \hat{F})_j = l$ given the other observed variables, i.e., $\{\bx_k: \bm_k=1, k \neq j\}$, is defined as
\bean
&& \Pr \left( g(\bx; \hat{F})_j = l \Big\vert \{\bx_k: \bm_k=1, k \neq j\} \right) \\
&\eqqcolon& \pi^*_{jl}\Big(\{\bx_k: \bm_k=1, k \neq j\}\Big),
\eean
which also corresponds to the conditional probability of $\hat{F}_j(\bx_j)$ in the $l$th bin, $[b_{l-1}, b_l)$. 

Then, the ground-truth conditional distribution of $g(\bx; \hat{F})_j$, which is our target distribution, is written accordingly:
\bea \label{eq:target_prob}
&& p_j^* \Big(g(\bx; \hat{F})_j | \{\bx_k: \bm_k=1, k \neq j\} \Big) \nonumber\\
&=& \prod_{l=1}^L \pi^*_{jl}\Big(\{\bx_k: \bm_k=1, k \neq j\}\Big)^{\mathbb{I}(g(\bx; \hat{F})_j = l)}.
\eea

\textbf{MaCoDE.}
For $j$ such that $\bm_j = 0$, we parameterize \eqref{eq:target_prob} as follows:
\bean
&& p_j \left(g(\bx; \hat{F})_j | g(\bx; \hat{F}) \odot \bm ; \theta \right) \\
&=& \prod_{l=1}^L \pi_{jl}\left(g(\bx; \hat{F}) \odot \bm ; \theta \right)^{\mathbb{I}(g(\bx; \hat{F})_j = l)},
\eean
where $\odot$ denotes element-wise multiplication and $\sum_{l=1}^L \pi_{jl}(g(\bx; \hat{F}) \odot \bm; \theta ) = 1$ for all $j$. In this paper, we use the empirical CDF for $\hat{F}_j$. Here, $\theta$ represents all the parameters of the transformer encoder-based classifier, allowing us to process inputs of arbitrary lengths and accommodate different combinations of observed and masked variables. Additionally, we denote $(\pi_{j1}(\cdot; \theta ), \cdots, \pi_{jL}(\cdot; \theta ))$ as $\pi_{j}(\cdot; \theta )$.

Then, the objective function for a single observation is defined as the negative log-likelihood of masked entries:
\bea \label{eq:obj1}
\cL(\by, \bm;\theta) &\coloneqq& - \sum_{j:\bm_j=0} \log p_j(\by_j | \by \odot \bm; \theta) \\
&=& - \sum_{j:\bm_j=0} \sum_{l=1}^L \mathbb{I}(\by_{j} = l) \cdot \log \pi_{jl}(\by \odot \bm; \theta), \nonumber
\eea
where the label $\by_j$ is defined as $\by_j = g(\bx; \hat{F})_j$ for all $j$. Our objective function \eqref{eq:obj1} estimates the conditional distribution of $\hat{F}_j(\bx_j)$ in each bin, where its target distribution is \eqref{eq:target_prob}.
In other words, our objective is to approximate the true conditional probability $\pi^*_{jl}$ by our estimated probability $\pi_{jl}$.

Similar to MLM, the bin index $(\by \odot \bm)_j$ serves as a `word' index aligned with the $j$th column, encompassing its own vocabulary set of $L+1$ words (i.e., $\{0, 1, 2, \cdots, L\}$), including `0' for the masked input. And $\pi_{jl}$ represents the probability that the model outputs the word index $l$ from the $j$th column. 

For each $j$, the embedding layer preprocesses of $(\by \odot \bm)_j$ through \texttt{one-hot} encoding, with each bin index being assigned a learnable embedding vector. Note that our approach to handling masked entries shares similarities with the existing \textit{zero imputation} technique, where all masked entries are replaced with zeros \cite{ivanov2018variational, Mattei2019MIWAEDG, nazabal2020hivae, ipsen2021notmiwae}. However, in our proposed method, although all masked entries are replaced with the same bin index, `0', each imputed zero value is embedded through distinct embedding vectors for each column. 


\begin{definition}[Mask distribution] \label{def:mask}
The distribution of mask vector $\bm$ is defined as:
\bean
p(\bm) \coloneqq \int_0^1 p(\bm|u) p(u) du,
\eean
where $p(u)$ is the uniform distribution density, and $\bm_1, \bm_2, \cdots, \bm_p$ follow conditionally independent Bernoulli distributions with probability $u$ given $u$. 
\end{definition}

Finally, we minimize the following objective function with respect to $\theta$:
\bea \label{eq:obj2}
\min_\theta \quad \sum_{i=1}^n \mbE_{p(\bm)} \big[ \cL(\by^{(i)}, \bm;\theta) \big],
\eea
where the discretized training dataset is $\{\by^{(i)}\}_{i=1}^n = \{g(\bx^{(i)}; \hat{F})\}_{i=1}^n$, and $\mbE_{p(\bm)}$ is approximated by Monte-Carlo and ancestral sampling. The distribution $p(\bm)$ can be defined by the user based on the specific problem requirements \cite{ivanov2018variational}. As in BERT \cite{Devlin2019BERTPO}, the task of our objective function \eqref{eq:obj2} is to predict the original label for all masked inputs. The overall structure of MaCoDE is outlined in Figure \ref{fig:overall}.

\begin{remark} \label{remark1}
We want to emphasize that our objective function \eqref{eq:obj2} does not imply an assumption of conditional independence among masked entries given the observed data. Instead, by ensuring that $p(\bm)$ has full support over $\{0, 1\}^p$, sampling $\bm$ from $p(\bm)$ allows us to learn conditional densities encompassing all possible combinations of conditioning sets and target variables \cite{ivanov2018variational, gulati2023tabmt}. Furthermore, this training scheme remains invariant to the missing data scenario.
\end{remark}

\begin{algorithm}[t] 
\caption{Synthetic Data Generation} 
\textbf{Initialize:} For all $j$, $\hat{\by}_j \leftarrow 0$ and $\hat{\bm}_j \leftarrow 0$. (all columns are masked) \\
\textbf{Output:} A synthetic sample $\hat{\bx}$.
\begin{algorithmic}[1]
    \FOR[tabular data lacks the inherent ordering between columns]{$j = randperm\{1,2,\cdots,p\}$} 
    \STATE $\hat{\by}_j \sim Cat\big(\pi_j(\hat{\by} \odot \hat{\bm}; \theta) / \tau \big)$
    \COMMENT{choosing a histogram bin (index) according to $\pi_{j}$}
    \STATE $\hat{\bm}_j \leftarrow 1$
    \COMMENT{un-mask the $j$th column}
    \ENDFOR 
    \FOR {$j = 1,2,\cdots,p$}
        \IF {$j \in I_C$}
            \STATE $u \sim U(b_{\hat{\by}_j-1}, b_{\hat{\by}_j})$
            \COMMENT{uniform sampling a quantile level from the bin interval}
            \STATE $\hat{\bx}_j \leftarrow \hat{F}^{-1}_j(u)$
            \COMMENT{inverse transform sampling of $\hat{p}_{j}$}
        \ENDIF
        \IF {$j \in I_D$}
        \STATE $\hat{\bx}_j \leftarrow \hat{\by}_j$
        \ENDIF
    \ENDFOR 
\end{algorithmic} \label{alg:syndata}
\end{algorithm}

\textbf{Synthetic data generation.} 
Tabular data lacks the inherent ordering between columns, unlike natural language. Therefore, as outlined in Algorithm \ref{alg:syndata}, MaCoDE randomly generates one column at a time, conditioned on masked subset sizes from $p$ to $1$, in descending order ($p \rightarrow p-1 \rightarrow \cdots \rightarrow 2 \rightarrow 1$). 

\textbf{Controllable privacy level.} 
Adjusting the privacy level during synthetic data generation is crucial in tabular domain \cite{Park2018DataSB}. Similar to TabMT \cite{gulati2023tabmt}, we can also regulate the privacy level using a single hyper-parameter $\tau$ without the need for re-training, while other existing synthesizers have fixed trade-offs between synthetic data quality and privacy level \cite{xu2019ctgan, Zhao2023CTABGANET, kotelnikov2023tabddpm} or require re-training \cite{an2023distributional}. 

By increasing the temperature parameter $\tau$ in $\hat{\by}_j \sim Cat\big(\pi_j(\hat{\by} \odot \hat{\bm}; \theta) / \tau \big)$ at line 2 of Algorithm \ref{alg:syndata}, we can mitigate the risk of privacy leakage. Figure \ref{fig:privacy_control} shows that MaCoDE allows for a trade-off between synthetic data quality (feature selection performance) and privacy preservability (DCR, Distance to Closest Record) as $\tau$ increases (see the Appendix for detailed results). 

\begin{figure}[t]
    \centering    \includegraphics[width=0.9\linewidth]{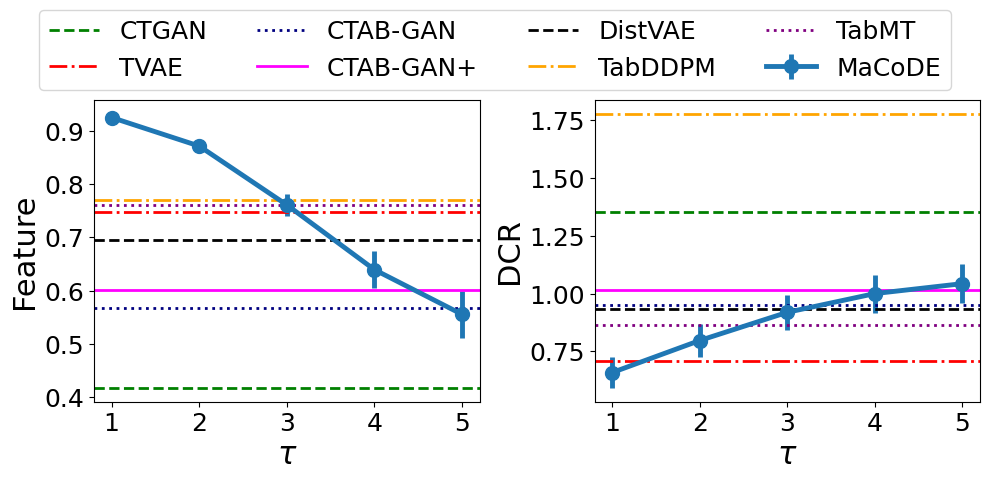}
    \caption{\textbf{Trade-off between quality and privacy.} Left: feature selection performance. Right: DCR. Error bars represent standard errors. See the Appendix for detailed results.}
    \label{fig:privacy_control}
\end{figure}

\subsection{Theoretical Results}

In this section, we aim to provide theoretical insights into MaCoDE's capabilities in conditional density estimation.
Firstly, we consider a factorization of the ground-truth joint PDF $p^*(\bx_1,\cdots,\bx_p)$ according to an arbitrary permutation $\sigma = [\sigma_1, \sigma_2, \cdots, \sigma_p]$ of the indices $\{1,2,\cdots,p\}$:
\bean
&& p^*_{\sigma_1}(\bx_{\sigma_1}) \cdot p^*_{\sigma_2}(\bx_{\sigma_2} | \bx_{\sigma_1}) \cdot p^*_{\sigma_3}(\bx_{\sigma_3} | \bx_{\sigma_1}, \bx_{\sigma_2}) \\
&& \times \cdots \times p^*_{\sigma_p}(\bx_{\sigma_p} | \bx_{\sigma_1}, \cdots, \bx_{\sigma_{p-1}}).
\eean
As discussed in Remark \ref{remark1}, our objective function \eqref{eq:obj2} facilitates the estimation of $p_{\sigma_j}^*(\bx_{\sigma_j} | \bx_{\sigma_1}, \cdots, \bx_{\sigma_{j-1}})$ for any $\sigma$ and $j$ by utilizing the mask distribution of Definition \ref{def:mask}. 

Without loss of generality, in this section, let $\sigma$ be an identity permutation. Since $c^*_j$ is the conditional density of $\hat{F}_j(\bx_j)$, the conditional probability of $\hat{F}_j(\bx_j)$ in the $l$th bin, $\pi_{jl}^{*}(\bx_{-j})$, is defined as
\bean
\pi_{jl}^{*}(\bx_{-j}) = \int_{b_{l-1}}^{b_l} c^*_j(v | \bx_{-j}) dv,
\eean
where $\bx_{-j} \coloneqq (\bx_1, \cdots, \bx_{j-1})$.

\begin{assumption} \label{assump:data} 
For all $j$, there exists $K_j \geq 0$ such that $c^*_j$ is $K_j$-Lipschitz.
\end{assumption}

Assumption \ref{assump:data} implies that $c^*_j$ is constrained in its rate of change, which allows us to estimate the conditional density using a piece-wise constant function \cite{wasserman2006all, tsybakov2009nonparametric}.

Based on \eqref{eq:changeofvariable}, by the change of variable under Assumption \ref{assump:CDF}, we define our conditional density estimator for $p^*_j$ as follows:
\bea \label{eq:estimator}
&& \hat{p}_j(\bx_j | \bx_{-j}; \theta) = \hat{c}_j(\hat{F}_j(\bx_j) | \bx_{-j}; \theta) \cdot \hat{p}(\bx_j) \nonumber\\
&=& \left( \sum_{l=1}^{L} \frac{\mathbb{I}(\hat{F}_{j}(\bx_{j}) \in [b_{l-1}, b_l))}{1/L} \cdot \pi_{jl} \right) \cdot \hat{p}(\bx_j),
\eea
where $\pi_{jl}$ denotes $\pi_{jl}(g(\bx;\hat{F}) \odot \bm^{(j)}; \theta)$ and $\bm^{(j)}$ is a masking vector such that
\bean
\bm^{(j)}_k &\coloneqq&
\begin{cases}
1, & \text{if $k \in \{1, \cdots, j-1\}$}\\
0, & \text{otherwise}
\end{cases}.
\eean

Note that $\hat{c}_j$ is the histogram-based conditional density estimator. Synthetic samples can be generated from \eqref{eq:estimator} as follows: Firstly, perform histogram sampling from $\hat{c}_{j}$ by choosing a histogram bin according to $\pi_{jl}$ and uniformly sampling from that bin interval. Then, apply inverse transform sampling of $\hat{p}_{j}$ with respect to the output of histogram sampling. This procedure is outlined in Algorithm \ref{alg:syndata}.



\begin{prop} \label{prop:main}
Under Assumption \ref{assump:CDF} and \ref{assump:data},
\bean
\text{TV}\Big( p^*_j(\cdot | \bx_{-j}), \hat{p}_j(\cdot | \bx_{-j}; \theta) \Big) &\leq& \frac{K_j}{2L} + \frac{\sqrt{Bias(\theta)}}{\sqrt{2}/L}
\eean
for all $j \in I_C$, and $\bx \in \mbR^p$. Here, $\text{TV}(\cdot, \cdot)$ denotes the total variation distance, and $Bias(\theta)$ is defined as:
\bean
&& Bias(\theta) = \sum_{l=1}^L \pi_{jl}^{*}(\bx_{-j}) \log \pi_{jl}^{*}(\bx_{-j}) \\
&-& \mbE_{\by_j|\bx_{-j}} \left[ \sum_{l=1}^L \mathbb{I}(\by_j = l) \log \pi_{jl}(g(\bx; \hat{F}) \odot \bm^{(j)}; \theta) \right]
\eean
where $\by_j|\bx_{-j}$ is a random variable having a categorical distribution such that $\Pr(\by_j = l|\bx_{-j}) = \Pr(g(\bx; \hat{F})_j = l|\bx_{-j}) = \pi_{jl}^{*}(\bx_{-j})$ for all $l \in [L]$.
\end{prop}

In the definition of $Bias(\theta)$, the second term on the right-hand side corresponds to the classification loss with respect to the target distribution given by \eqref{eq:target_prob}. This implies that $Bias(\theta)$ can be minimized when the classification loss is minimized. And Proposition \ref{prop:main} demonstrates that the total variation distance between the ground-truth conditional density and our conditional density estimator $\hat{p}_{j}$ is upper bounded by $Bias(\theta)$. 

Therefore, minimizing the orderless multi-class classification loss leads to minimizing the total variation distance between conditional distributions, making Algorithm \ref{alg:syndata} capable of generating theoretically valid synthetic samples under certain assumptions. Note that Proposition \ref{prop:main} holds for any arbitrary permutation $\sigma$, and the CDF transformation is crucial for our proposed method, as it ensures the validity of Proposition \ref{prop:main}. 

\subsection{With Missing Data}
\label{sec:missing}

Suppose the pattern of missingness varies individually for each observation and is defined by a corresponding missing indicator, denoted as ${\bf r}$, where the indicators for all observations are represented as $\{\bbr^{(i)}\}_{i=1}^n$. Here, $\bbr_j = 0$ indicates that $\bx_j$ is missing, while $\bbr_j = 1$ denotes that $\bx_j$ is observed. In cases where the training data contains missing entries (e.g., not a number), it is not feasible to input these missing entries and minimize their log-likelihoods. Therefore, to handle missing value inputs using $g$, irrespective of the value of $\bx_j$, we further define $g$ for Definition \ref{def:preprocess} as follows: for any $F$, $g(\bx; F)_j \coloneqq 0$ if $\bbr_j = 0$.
This indicates the bin index `0' is assigned to a missing value input.

Therefore, the objective function with missing data is 
$\min_\theta \sum_{i=1}^n \mbE_{p(\bm)} \big[ \cL^*(\bx^{(i)}, \bbr^{(i)}, \bm;\theta) \big]$,
where
\bean
\cL^*(\bx, \bbr, \bm;\theta) &\coloneqq& - \sum_{j:\bm_j=0, \bbr_j=1} \sum_{l=1}^L \mathbb{I}(g(\bx;\hat{F})_j = l) \\
&& \times \log \pi_{jl}(g(\bx;\hat{F}) \odot \min(\bm, \bbr); \theta),
\eean
$\min(\bm, \bbr)$ is element-wise minimum operation, and $\hat{F}$ is estimated using the observed dataset \cite{Chenouri2009EmpiricalMF}. Note that, in $\cL^*(\bx, \bbr, \bm;\theta)$, we do not minimize the negative log-likelihood of missing entries. 

\begin{prop} \label{prop:mar}
Assuming $\bm \indep \bbr | \bx$ and $\bm \indep \bx$, if the data of $\bx$ is MAR, then the following holds for the missing data model $p(\bm, \bbr|\bx)$:
\bean
p(\bm, \bbr|\bx) = p(\bm, \bbr|\bx_{obs}),
\eean
where $\bx_{obs}$ represents the observed covariates from $\bx$, and the missingness pattern of $\min(\bm, \bbr)$ also follows the MAR mechanism.
\end{prop}


In $\cL^*(\bx, \bbr, \bm;\theta)$, the missingness pattern is described by $\min(\bm, \bbr)$, indicating that the missing data model is determined by the joint distribution $p(\bm, \bbr|\bx)$. Thus, as the masking vector defined in Definition \ref{def:mask} satisfies the conditions in Proposition \ref{prop:mar}, the missingness pattern of our proposed model also conforms to the MAR mechanism according to Proposition \ref{prop:mar} if the given data is MAR. It implies that our inference strategies based on the observed dataset can be justified \cite{Mattei2019MIWAEDG}. However, we empirically demonstrate in Section \ref{sec:exp} that our proposed model is also applicable to other missing data scenarios.


\section{Experiments}
\label{sec:exp}

\subsection{Overview}

We conduct experiments in which we can provide answers to the following three experimental questions:
\bed
    \item[\textbf{Q1}.] Does MaCoDE achieve state-of-the-art performance in synthetic data generation?
    \item[\textbf{Q2}.] Can MaCoDE generate high-quality synthetic data even when faced with missing data scenarios?
    \item[\textbf{Q3}.] Is MaCoDE capable of supporting multiple imputations for deriving statistically valid inferences from missing data?
\eed

\begin{table*}[t]
  \centering
  \resizebox{0.95\textwidth}{!}{
  \begin{tabular}{lrrrrrrrrrrrrrrrr}
    \toprule
    & \multicolumn{4}{c}{Statistical fidelity} & \multicolumn{4}{c}{Machine learning utility} \\
    \cmidrule(lr){2-5} \cmidrule(lr){6-9}
    Model & KL $\downarrow$ & GoF $\downarrow$ & MMD $\downarrow$ & WD $\downarrow$ & SMAPE $\downarrow$ & $F_1$ $\uparrow$ & Model $\uparrow$ & Feature $\uparrow$\\
    \midrule
Baseline & $.016_{\pm .002}$ & $.029_{\pm .002}$ & $.002_{\pm .000}$ & $1.019_{\pm .156}$ & $.107_{\pm .008}$ & $.686_{\pm .023}$ & $.887_{\pm .018}$ & $.956_{\pm .005}$\\
\midrule
CTGAN & $.221_{\pm .014}$ & $.561_{\pm .046}$ & $.094_{\pm .007}$ & $6.435_{\pm 1.011}$ & $.256_{\pm .016}$ & $.411_{\pm .027}$ & $.208_{\pm .048}$ & $.417_{\pm .043}$\\
TVAE & $.066_{\pm .003}$ & $.119_{\pm .005}$ & $.016_{\pm .001}$ & $\underline{1.631}_{\pm .173}$ & $.192_{\pm .011}$ & $.608_{\pm .021}$ & $.486_{\pm .041}$ & $.747_{\pm .027}$\\
CTAB-GAN & $.116_{\pm .008}$ & $.196_{\pm.025}$ & $.044_{\pm .004}$ & $3.327_{\pm.460}$ & $.218_{\pm.012}$ & $.524_{\pm.026}$ & $.263_{\pm.042}$ & $.568_{\pm .041}$\\
CTAB-GAN+ & $.136_{\pm .018}$ & $.144_{\pm.010}$ & $.054_{\pm .007}$ & $3.971_{\pm.772}$ & $.226_{\pm.017}$ & $.530_{\pm.020}$ & $.227_{\pm.048}$ & $.601_{\pm.041}$\\
DistVAE & $.059_{\pm .007}$ & $\underline{.070}_{\pm .004}$ & $.016_{\pm .001}$ & $2.272_{\pm .282}$ & $.226_{\pm .017}$ & $.588_{\pm .021}$ & $.194_{\pm .048}$ & $.695_{\pm .030}$\\
TabDDPM & $.696_{\pm .117}$ & $.374_{\pm .087}$ & $.057_{\pm .011}$ & $42.916_{\pm 8.127}$ & $\underline{.161}_{\pm .011}$ & $.576_{\pm .022}$ & $.507_{\pm .039}$ & $\underline{.770}_{\pm .027}$ \\
TabMT & $\mathbf{.011}_{\pm .001}$ & $\mathbf{.035}_{\pm .003}$ & $\underline{.012}_{\pm .001}$ & $2.299_{\pm .346}$ & $.188_{\pm .013}$ & $\underline{.622}_{\pm .024}$ & $\underline{.528}_{\pm .039}$ & $.761_{\pm .028}$ \\
\midrule
MaCoDE & $\underline{.034}_{\pm .004}$ & $.072_{\pm .004}$ & $\mathbf{.007}_{\pm .001}$ & $\mathbf{1.630}_{\pm .245}$ & $\mathbf{.158}_{\pm .010}$ & $\mathbf{.635}_{\pm .023}$ & $\mathbf{.599}_{\pm .035}$ & $\mathbf{.925}_{\pm .007}$\\
\midrule
MaCoDE(MCAR) & - & -& -& -& $.168_{\pm .011}$ & $.623_{\pm .023}$ & - & -\\
MaCoDE(MAR) & - & -& -& -&  $.167_{\pm .011}$ & $.626_{\pm .023}$& - & -\\
MaCoDE(MNARL) & - & -& -& -&  $.169_{\pm .011}$ & $.624_{\pm .023}$& - & -\\
MaCoDE(MNARQ) & - & -& -& -& $.164_{\pm .010}$ & $.630_{\pm .023}$& - & -\\
    \bottomrule
  \end{tabular}
  }
\caption{\textbf{Q1} and \textbf{Q2}. The means and the standard errors of the mean across 10 datasets and 10 repeated experiments are reported. Across all missingness patterns, a missingness rate of 0.3 is employed. $\uparrow$ ($\downarrow$) denotes higher (lower) is better. The best value is bolded, and the second best is underlined.}
\label{tab:performance}
\end{table*}

\textbf{Datasets.}
Similar to several recent studies \cite{gulati2023tabmt, kotelnikov2023tabddpm}, we utilize 10 publicly available real tabular UCI and Kaggle\footnote{\url{https://archive.ics.uci.edu/}, \url{https://www.kaggle.com/datasets/}} datasets of varying sizes and the number of columns. Detailed statistics of these datasets are provided in the Appendix. Note that we include \texttt{covtype} dataset, which comprises approximately 580K rows, to demonstrate the scalability of our proposed model. 

\textbf{Baseline models.}
For MaCoDE, we set $L=50$ and $\tau=1$ for all datasets. Detailed hyperparameter settings are provided in the Appendix.\footnote{We run experiments using NVIDIA A10 GPU, and our experimental codes are available with \texttt{pytorch}.}
For Q1 and Q2, we compare MaCoDE with CTGAN \cite{xu2019ctgan}, TVAE \cite{xu2019ctgan}, CTAB-GAN \cite{zhao2021ctabgan}, CTAB-GAN+ \cite{Zhao2023CTABGANET}, DistVAE \cite{an2023distributional}, TabDDPM \cite{kotelnikov2023tabddpm}, and TabMT \cite{gulati2023tabmt}. 
For Q3, we selected the following multiple imputation models that can handle mixed-type tabular datasets: MICE \cite{Buuren2011MICEMI}, GAIN \cite{YOON2018GAIN}, missMDA \cite{missMDA}, VAEAC \cite{ivanov2018variational}, MIWAE \cite{Mattei2019MIWAEDG}, not-MIWAE \cite{ipsen2021notmiwae}, and EGC \cite{zhao2022probabilistic}. Detailed experimental settings for these baseline models are provided in the Appendix.

To assist in interpreting the metrics, we include a baseline synthetic dataset where the synthetic data comprises half of the real training dataset. This dataset, referred to as `Baseline,' serves as a soft upper bound for evaluating the quality of the synthetic data.

\textbf{Additional evaluations.} Due to space constraints, (1) the results for controlling data privacy levels with varying temperature parameter $\tau$, and (2) the results for \textbf{Q3} (including related works and sensitivity analysis regarding the missingness rate) are provided in the Appendix.

\subsection{Evaluation Metrics}
\label{sec:4.2}

For all metrics, we report the mean and standard error of the mean (error bars) across 10 different random seeds and 10 datasets. For each random seed, the dataset is randomly split into training and testing sets with an 80\% training and 20\% testing ratio in the evaluation of Q1 and Q2. During evaluation, the synthetic dataset is generated to have the same number of samples as the real training dataset.

\textbf{Q1.} 
To evaluate the quality of generated synthetic data, we employ two metrics: statistical fidelity \cite{qian2023synthcity} and machine learning utility \cite{hansen2023reimagining}. For statistical fidelity, we utilize the Kullback–Leibler divergence (KL) and the Goodness-of-Fit (GoF) test (continuous: the two-sample Kolmogorov-Smirnov test statistic, categorical: the Chi-Squared test statistic) to assess marginal distributional similarity. Additionally, we employ the Maximum Mean Discrepancy (MMD) and 1-Wasserstein distance (WD) to measure joint distributional similarity. These metrics measure how well the synthetic data maintains statistical fidelity to the real training dataset.

Regarding machine learning utility, we use four metrics outlined in \cite{hansen2023reimagining}: regression performance (SMAPE, symmetric mean absolute percentage error), classification performance ($F_1$), model selection performance (Model), and feature selection performance (Feature). These metrics are assessed by fitting the machine learning model on the real training and synthetic datasets individually and then comparing the performance of the two models on the test dataset. 
For a detailed evaluation procedure of Q1, refer to the Appendix.

\textbf{Q2.} 
To assess our proposed model's capability to generate high-quality synthetic datasets despite missing values in the training data, we evaluate the model trained on incomplete training datasets. 
Since other metrics require a complete training dataset for measurement, we only assess synthetic data quality using downstream regression and classification tasks on the test dataset, as these metrics can be evaluated using the test dataset alone, as in Q1.
See the Appendix for a detailed evaluation procedure.

\textbf{Q3.}\footnote{Since some off-the-shelf packages (\texttt{missMDA}, \texttt{gcimpute}) fail to deliver useful results, we include only 5 datasets that provide meaningful results: \texttt{abalone}, \texttt{banknote}, \texttt{breast}, \texttt{redwine}, and \texttt{whitewine}.}
We assess the effectiveness of multiple imputations by employing interval inference for the population mean, which was proposed by Rubin \cite{Rubin1986MultipleIF}. We report the bias, coverage, and confidence interval length. 
The detailed evaluation procedure for multiple imputations and the missing value generation mechanisms (MCAR, MAR, MNARL, MNARQ) is provided in the Appendix.

\subsection{Results}

\textbf{Q1.}
As shown in Table \ref{tab:performance}, MaCoDE consistently achieves the highest metric scores in both joint distributional similarities and machine learning utility while also achieving competitive performance in marginal distributional similarity. This underscores the effectiveness of the synthetic data generation method based on estimating conditional distribution in preserving the joint statistical fidelity of the original data and enhancing the utility of synthetic data for downstream machine learning tasks. Notably, MaCoDE demonstrates remarkable performance in the feature selection downstream task.

\begin{remark}[How does MaCoDE achieve the remarkable performance in feature selection?]
We attribute the effectiveness of the feature selection downstream task to our emphasis on estimating `conditional' distributions. In Random Forest  \cite{breiman2001random}, each node in a decision tree represents a conditional distribution of a variable conditioned on the splits made by the tree up to that node, and the feature importance is determined by the purity of nodes. Therefore, accurately estimating the conditional distribution can lead to higher performance in preserving the feature importance ranking.
\end{remark}

\textbf{Q2.}
The missing data mechanism within the parentheses refers to the mechanism applied to the training dataset on which MaCoDE was trained.
Despite encountering missing data scenarios such as MAR, MCAR, MNARL, and MNARQ, Table \ref{tab:performance} illustrates MaCoDE's ability to generate high-quality synthetic data regarding machine learning utility while handling incomplete training datasets without significant performance degradation. Even in the presence of missing entries, MaCoDE achieves better metric scores than most baseline models in terms of SMAPE, except for TabDDPM. Additionally, concerning the $F_1$ score, MaCoDE either competes competitively or achieves a higher score than other baseline models.


\begin{table}[t]
    \centering
      \resizebox{0.95\linewidth}{!}{
      \begin{tabular}{lrrrrrrrrrrrrrrrr}
        \toprule
        Model & Bias $\downarrow$ & Coverage & Width $\downarrow$ \\
        \midrule
    MICE & $.010_{\pm .001}$ & $.845_{\pm .019}$ & $\mathbf{.040}_{\pm .002}$\\
    GAIN & $.019_{\pm .002}$ & $.633_{\pm .033}$ & $\mathbf{.040}_{\pm .002}$  \\
    missMDA & $.015_{\pm .001}$ & $.700_{\pm .022}$ & $.043_{\pm .002}$\\
    VAEAC & $\underline{.008}_{\pm .001}$ & $.905_{\pm .016}$ & $\mathbf{.040}_{\pm .002}$\\
    MIWAE & $\mathbf{.006}_{\pm .000}$ & $\underline{.952}_{\pm .012}$ & $.043_{\pm .002}$\\
    not-MIWAE & $\mathbf{.006}_{\pm .000}$ & $\mathbf{.949}_{\pm .012}$ & $\underline{.042}_{\pm .002}$\\
    EGC & $\mathbf{.006}_{\pm .000}$ & ${.996}_{\pm .004}$ & $.058_{\pm .002}$\\
        \midrule
    MaCoDE(MAR) & $\mathbf{.006}_{\pm .000}$ & $.963_{\pm .009}$ & $.051_{\pm .003}$\\
        \bottomrule
      \end{tabular}
      }
    \caption{\textbf{Q3} under MAR at 0.3 missingness. The means and standard errors of the mean across 5 datasets and 10 repeated experiments are reported. $\downarrow$ denotes lower is better. Coverage close to 0.95 indicates better performance. The best value is bolded, and the second best is underlined.}
    \label{tab:impute}
\end{table}

\textbf{Q3.}
The missing data mechanism within the parentheses refers to the mechanism applied to the dataset on which MaCoDE was trained.
Table \ref{tab:impute} indicates that MaCoDE consistently exhibits competitive performance against all baseline models across metrics assessing multiple imputation performances, including bias, coverage, and confidence interval length. This suggests our proposed approach can support multiple imputations for deriving statistically valid inferences from missing data with the MAR mechanism.

\begin{figure}[t]
    \centering
    \includegraphics[width=0.9\linewidth]{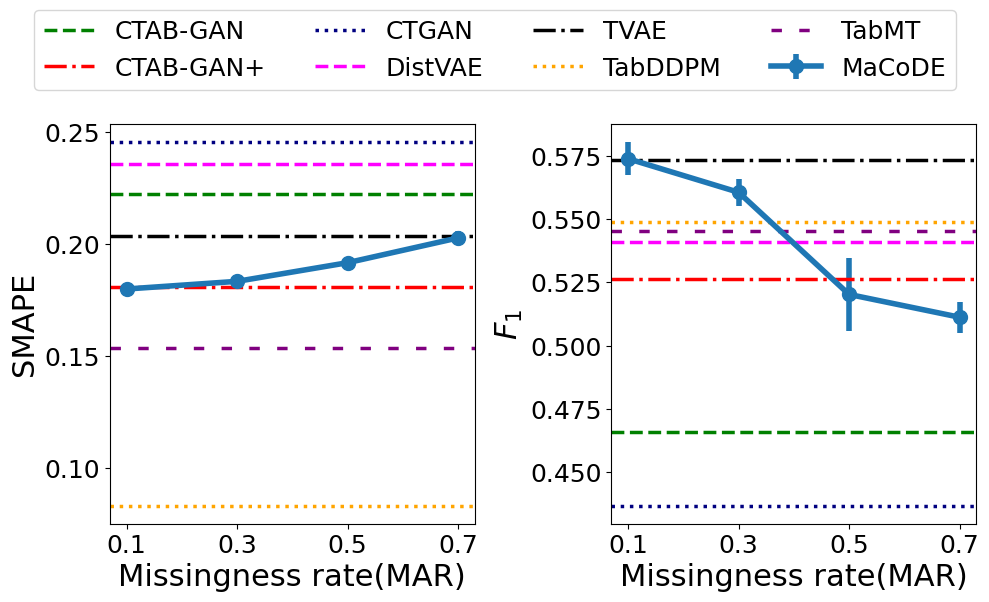}
    \caption{
    Sensitivity analysis with respect to missingness rate using \texttt{kings} dataset is performed for \textbf{Q2} under MAR. The analysis focuses on machine learning utility. Results are reported as means and standard errors of the mean from 10 repeated experiments, with error bars representing the standard errors.
    }
    \label{fig:sensitivity}
\end{figure}

\textbf{Sensitivity analysis.} 
We also conducted a sensitivity analysis by varying the missingness rate of \texttt{kings} dataset. Figure \ref{fig:sensitivity} illustrates that, in terms of SMAPE, MaCoDE maintains competitive performance even as the missingness rate increases. Concerning the $F_1$ score, MaCoDE outperforms other models at missingness rates of 0.1 and 0.3, but its performance declines beyond a missingness rate of 0.5. 

\section{Conclusions and Limitations}

This paper introduces an approach to generating synthetic data for mixed-type tabular datasets. Our proposed method integrates histogram-based non-parametric conditional density estimation and the MLM-based approach while bridging the theoretical gap between distributional learning and the consecutive multi-class classification task of MLM. Although our primary goal is to generate synthetic data with high MLu, we empirically demonstrate that we achieve high joint statistical fidelity and MLu simultaneously. Furthermore, empirical experiments validate that our proposed model can generate high-quality synthetic tabular datasets in terms of MLu even when incomplete training datasets are given.

Although MaCoDE demonstrates the ability to perform `arbitrary' conditional density estimation by accommodating various combinations of conditioning sets and target variables, and despite empirical results showing its effectiveness in handling diverse distributions of continuous columns and generating high-quality synthetic data, the model has theoretical limitations. Specifically, it is valid under Lipschitz continuity (Assumption \ref{assump:data}). Addressing the limitation of accommodating a broader range of continuous distributions is an important direction for future work.





\bibliography{ref}

\onecolumn
\appendix

\section{Appendix / supplemental material}
\label{sec:app}

\subsection{Proof of Proposition 1}
\label{app:proof}

\begin{proof}
Without loss of generality, let $\sigma$ be an identity permutation. We equally space grid points on the interval $[0, 1]$ such that bins are represented as $[b_{l-1}, b_l), l=1,\cdots,L$, where $h$ denotes the bin width such that $L h = 1$. We consider arbitrary $j \in I_C$ and $\bx \in \mbR^p$, where $\bx$ is given and fixed. For notational simplicity, let $\bx_{-j} \coloneqq (\bx_1, \cdots, \bx_{j-1})$. 

Suppose that $\hat{F}_j(\bx_j) \in [b_{l-1}, b_l)$. By the mean value theorem, there exists $u^* \in (b_{l-1}, b_l)$ such that 
\bean
h \cdot c^*_j(u^* | \bx_{-j}) = \int_{b_{l-1}}^{b_l} c^*_j(v | \bx_{-j}) dv = \pi_{jl}^{*}(\bx_{-j}).
\eean

And we have
\bean
&& \Big\vert c^*_j(\hat{F}(\bx_{j}) | \bx_{-j}) - \hat{c}_j(\hat{F}(\bx_{j}) | \bx_{-j}; \theta) \Big\vert \\
&=& \left\vert c^*_j(\hat{F}(\bx_{j}) | \bx_{-j}) - \frac{\pi_{jl}(g(\bx; F) \odot \bm^{(j)}; \theta)}{h} \right\vert \\
&\leq& \left\vert c^*_j(\hat{F}(\bx_{j}) | \bx_{-j}) - c^*_j(u^* | \bx_{-j}) \right\vert + \left\vert c^*_j(u^* | \bx_{-j}) - \frac{\pi_{jl}(g(\bx; F) \odot \bm^{(j)}; \theta)}{h} \right\vert \\
&\leq& |\hat{F}(\bx_{j}) - u^*| \left\vert \frac{c^*_j(\hat{F}(\bx_{j}) | \bx_{-j}) - c^*_j(u^* | \bx_{-j})}{\hat{F}(\bx_{j}) - u^*} \right\vert + \left\vert \frac{\pi_{jl}^{*}(\bx_{-j})}{h} - \frac{\pi_{jl}(g(\bx; F) \odot \bm^{(j)}; \theta)}{h} \right\vert \\
&\leq& h K_j + \frac{1}{h} \left\vert \pi_{jl}^{*}(\bx_{-j}) - \pi_{jl}(g(\bx; F) \odot \bm^{(j)}; \theta) \right\vert \text{\quad (by Assumption 2)}\\
&\leq& h K_j + \frac{1}{h} \sum_{l=1}^L \left\vert \pi_{jl}^{*}(\bx_{-j}) - \pi_{jl}(g(\bx; F) \odot \bm^{(j)}; \theta) \right\vert.
\eean
where the first equality follows from the definition of $\hat{c}_j$. 

By the Pinsker's inequality, 
\bean
&& \frac{1}{2} \sum_{l=1}^L \left\vert \pi_{jl}^{*}(\bx_{-j}) - \pi_{jl}(g(\bx; F) \odot \bm^{(j)}; \theta) \right\vert \\
&\leq& \frac{1}{\sqrt{2}} \left( \sum_{l=1}^L \pi_{jl}^{*}(\bx_{-j}) \log \pi_{jl}^{*}(\bx_{-j}) - \sum_{l=1}^L \pi_{jl}^{*}(\bx_{-j}) \log \pi_{jl}(g(\bx; F) \odot \bm^{(j)}; \theta) \right)^{1/2} \\
&\leq& \frac{1}{\sqrt{2}} \left( \sum_{l=1}^L \pi_{jl}^{*}(\bx_{-j}) \log \pi_{jl}^{*}(\bx_{-j}) - \mbE_{\by_j|\bx_{-j}} \left[ \sum_{l=1}^L \mathbb{I}(\by_j = l) \log \pi_{jl}(g(\bx; F) \odot \bm^{(j)}; \theta) \right] \right)^{1/2},
\eean
where $\by_j|\bx_{-j}$ is a random variable having a categorical distribution such that $\Pr(\by_j = l|\bx_{-j}) = \Pr(g(\bx; \hat{F})_j = l|\bx_{-j}) = \pi_{jl}^{*}(\bx_{-j})$ for all $l \in [L]$.

Then, 
\bea \label{eq:proof1}
&& \Big\vert c^*_j(\hat{F}(\bx_{j}) | \bx_{-j}) - \hat{c}_j(\hat{F}(\bx_{j}) | \bx_{-j}; \theta) \Big\vert \nonumber\\
&\leq& h K_j + \frac{\sqrt{2}}{h} \left( \sum_{l=1}^L \pi_{jl}^{*}(\bx_{-j}) \log \pi_{jl}^{*}(\bx_{-j}) - \mbE_{\by_j|\bx_{-j}} \left[ \sum_{l=1}^L \mathbb{I}(\by_j = l) \log \pi_{jl}(g(\bx; F) \odot \bm^{(j)}; \theta) \right] \right)^{1/2}.
\eea

The total variation distance between $p^*_j(\cdot | \bx_{-j})$ and $\hat{p}_j(\cdot | \bx_{-j}; \theta)$ is written as
\bean
&& \text{TV}\Big( p^*_j(\cdot | \bx_{-j}), \hat{p}_j(\cdot | \bx_{-j}; \theta) \Big) \\
&=& \frac{1}{2} \int_\mbR \Big\vert p^*_j(\bx_{j} | \bx_{-j}) - \hat{p}_j(\bx_{j} | \bx_{-j}; \theta) \Big\vert d\bx_j \\
&=& \frac{1}{2} \int_\mbR \Big\vert c^*_j(\hat{F}(\bx_{j}) | \bx_{-j}) - \hat{c}_j(\hat{F}(\bx_{j}) | \bx_{-j}; \theta) \Big\vert \cdot \hat{p}_j(\bx_j) d\bx_j \text{\quad (by Assumption 1)}\\
&\leq& \frac{h K_j}{2} + \frac{1}{\sqrt{2}h} \left( \sum_{l=1}^L \pi_{jl}^{*}(\bx_{-j}) \log \pi_{jl}^{*}(\bx_{-j}) - \mbE_{\by_j|\bx_{-j}} \left[ \sum_{l=1}^L \mathbb{I}(\by_j = l) \log \pi_{jl}(g(\bx; F) \odot \bm^{(j)}; \theta) \right] \right)^{1/2} \\
&=& \frac{h K_j}{2} + \frac{\sqrt{Bias(\theta)}}{\sqrt{2}h} \\
&=& \frac{K_j}{2L} + \frac{\sqrt{Bias(\theta)}}{\sqrt{2}/L},
\eean
where the inequality holds for \eqref{eq:proof1}.

The proof is complete.
\end{proof}

\subsection{Proof of Proposition 2}

\begin{proof}
Since the data is MAR, $p(\bbr|\bx) = p(\bbr|\bx_{obs})$ and
\bean
p(\bm, \bbr|\bx) &=& p(\bbr|\bx) \cdot p(\bm|\bx) = p(\bbr|\bx) \cdot p(\bm) \\
&=& p(\bbr|\bx_{obs}) \cdot p(\bm) = p(\bm, \bbr|\bx_{obs}).
\eean
The proof is complete.
\end{proof}

\clearpage
\subsection{Dataset Descriptions} \label{app:data}

\textbf{Download links.}
\bed
    \small
    \item \texttt{abalone} \cite{abalone}: \url{https://archive.ics.uci.edu/dataset/1/abalone}
    \item \texttt{banknote} \cite{banknote}: \url{https://archive.ics.uci.edu/dataset/267/banknote+authentication}
    \item \texttt{breast} \cite{breast}: \\ \url{https://archive.ics.uci.edu/dataset/17/breast+cancer+wisconsin+diagnostic}
    \item \texttt{concrete} \cite{concrete}: \\ \url{https://archive.ics.uci.edu/dataset/165/concrete+compressive+strength}
    \item \texttt{covertype} \cite{covertype}: \url{https://www.kaggle.com/datasets/uciml/forest-cover-type-dataset}
    \item \texttt{kings} (CC0: Public Domain): \url{https://www.kaggle.com/datasets/harlfoxem/housesalesprediction}
    \item \texttt{letter} \cite{letter}: \url{https://archive.ics.uci.edu/dataset/59/letter+recognition}
    \item \texttt{loan} (CC0: Public Domain): \url{https://www.kaggle.com/datasets/teertha/personal-loan-modeling}
    \item \texttt{redwine} \cite{wine}: \url{https://archive.ics.uci.edu/dataset/186/wine+quality}
    \item \texttt{whitewine} \cite{wine}: \url{https://archive.ics.uci.edu/dataset/186/wine+quality}
\eed

\begin{table}[ht]
  \centering
  \begin{tabular}{lrrrrrr}
    \toprule
    Dataset & Train/Test Split & \#continuous & \#categorical & Classification Target  \\
    \midrule
\texttt{abalone} & 3.3K/0.8K & 7 & 2 & \texttt{Rings}\\
\texttt{banknote} &1.1K/0.3K & 4 & 1 &  \texttt{class}\\
\texttt{breast} & 0.5K/0.1K & 30 & 1 & \texttt{Diagnosis}\\
\texttt{concrete} & 0.8K/0.2K & 8 & 1 & \texttt{Age}\\
\texttt{covtype} & 0.46M/0.12M & 10 & 1 & \texttt{Cover\_Type}\\
\texttt{kings} & 17.3K/4.3K & 11 & 7 & \texttt{grade}\\
\texttt{letter} & 16K/4K & 16 & 1 & \texttt{lettr}\\
\texttt{loan} & 4K/1K & 5 & 6 & \texttt{Personal Loan}\\
\texttt{redwine} & 1.3K/0.3K & 11& 1 & \texttt{quality}\\
\texttt{whitewine} & 3.9K/1K & 11 & 1 & \texttt{quality}\\
    \bottomrule
  \end{tabular}
\caption{\textbf{Description of datasets.} \#continuous represents the number of continuous and ordinal variables. \#categorical denotes the number of categorical variables. The `Classification Target' refers to the variable used as the response variable in a classification task to evaluate machine learning utility.}
\label{tab:data_description}
\end{table}

\subsection{Experimental Settings for Reproduction}
\label{app:settings}

\bed
    \item We run experiments using NVIDIA A10 GPU, and our experimental codes are available with \texttt{pytorch}.
    \item In practice, for all $j \in I_C$, we estimate $\hat{F}_j$ using empirical measure. In the presence of missing data, $\hat{F}_j$ is estimated solely using the observed dataset (refer to \cite{Chenouri2009EmpiricalMF}).
\eed

\textbf{Hyper-parameters of MaCoDE:} 
As shown in Section \ref{app:results}, MaCoDE consistently generates high-quality synthetic data without requiring an extensive hyperparameter tuning process, unlike methods such as \cite{kotelnikov2023tabddpm, lee2023CoDi, kim2023stasy}. This demonstrates the generalizability of our proposed model to various tabular datasets. \textbf{Our implementation codes for the proposed model, MaCoDE, are provided in the supplementary material.} \textbf{For all tabular datasets, we applied the following hyperparameters uniformly without any additional tuning}:
\bed
    \item epochs: 500
    \item batch size: 1024
    \item learning rate: 0.001 (with AdamW optimizer \cite{loshchilov2017decoupled} with 0.001 weight decay parameter)
    \item the number of bins: $L = 50$
    \item Transformer encoder dimension: 128
    \item Transformer encoder \#heads: 4
    \item Transformer encoder \#layer: 2
    \item Transformer encoder dropout ratio: 0.0
\eed

\subsubsection{Details of Implementing Baseline Models}

To compare our proposed method with baseline models, we performed experiments in a manner similar to \cite{zhang2024mixedtype}. We adjusted the hidden or latent dimensions across various methods, ensuring that the number of trainable parameters is comparable. Since each model employs distinct neural network architectures, it would be misleading to compare their performances using the same network configuration. It's worth noting that model size typically reflects the performance of the baseline models. Under these conditions, we reproduced the baseline methods using their official codes and default configurations (except for the hidden or latent dimensions). \textbf{Our reproduced codes for the baseline models are provided in the supplementary material.} Below are the detailed implementations of the baseline methods:

\bed 
    \item CTGAN and TVAE \cite{xu2019ctgan}: 
    We follow the implementations provided in the official repository\footnote{\url{https://github.com/sdv-dev/CTGAN}} for CTGAN and TVAE. We adopt the default hyperparameters specified in the module. To ensure fairness in comparison by aligning model sizes, we adjust the latent dimension of CTGAN and TVAE to 100.
    
    \item CTAB-GAN and CTAB-GAN+ \cite{zhao2021ctabgan, Zhao2023CTABGANET}: 
    We follow the implementations provided in the official repository\footnote{\url{https://github.com/Team-TUD/CTAB-GAN}, \url{https://github.com/Team-TUD/CTAB-GAN-Plus}}. The latent dimension is set to 100, and the maximum number of clusters is set to 10. For the specification of feature types, we use three categories: continuous, integer, and categorical variables.
    
    \item DistVAE \cite{an2023distributional}: We follow the implementations provided in the official repository\footnote{\url{https://github.com/an-seunghwan/DistVAE}}. We changed the latent dimension to 100 to increase the model size. 

    \item TabDDPM \cite{kotelnikov2023tabddpm}: 
    We utilized the TabDDPM module in \texttt{synthcity.Plugins} \footnote{\url{https://github.com/vanderschaarlab/synthcity}} for synthetic data generation. This module follows the implementations provided in the official repository. 
    
    \item TabMT \cite{gulati2023tabmt}: Rather than using K-means clustering, we utilize the Gaussian Mixture Model (GMM) to discretize continuous columns to preserve the original continuous domain. We determine the optimal number of clusters for the GMM, ranging from 2 to 10, based on the Bayesian Information Criterion (BIC). During the generation of continuous columns, we initially predict the component label and then sample from the selected Gaussian component.

    \item MICE \cite{Buuren2011MICEMI}: We employed the \texttt{IterativeImputer} package from \texttt{Scikit-learn} for multiple imputation using chained equations. Following the authors' experiments, a \texttt{max\_iter} range of 10 to 20 was considered sufficient for convergence, and we adopted this setting for our experiments. Additionally, to introduce randomness, we set \texttt{imputation\_order} to \texttt{random}, which randomly selects a variable for imputation in each iteration. The remaining parameters were left at their default values to maintain the integrity of the MICE implementation.

    \item GAIN \cite{YOON2018GAIN}: We follow the implementations provided in the official repository\footnote{\url{https://github.com/jsyoon0823/GAIN/tree/master}}. As the paper does not explicitly discuss the separate handling of categorical and continuous variables, the code treats them simultaneously. Consequently, a rounding process is employed to handle categorical variables afterward.

    \item missMDA \cite{missMDA}: We utilized the \texttt{missMDA} package\footnote{\url{https://cran.r-project.org/web/packages/missMDA/index.html}} in \texttt{R} for multiple imputation.   

    \item VAEAC \cite{ivanov2018variational}: We follow the implementations provided in the official repository\footnote{\url{https://github.com/tigvarts/vaeac}}. The authors provided hyperparameters that adequately address both continuous and categorical variables, so we used these without further modification during model fitting.
    
    \item MIWAE \cite{Mattei2019MIWAEDG}: The implemented MIWAE code in the official repository\footnote{\url{https://github.com/pamattei/miwae}} was designed for continuous variables only. To accommodate heterogeneous tabular datasets, we treated the conditional distribution of categorical columns as categorical distributions and employed cross-entropy loss for reconstruction. For comparison with not-MIWAE, we set the latent dimension to $p-1$.

    \item not-MIWAE \cite{ipsen2021notmiwae}: The implemented not-MIWAE code in the official repository\footnote{\url{https://github.com/nbip/notMIWAE}} also focused on continuous variables exclusively. To handle categorical variables, we made the same modifications as in MIWAE. For comparison with MIWAE, we set the latent dimension to $p-1$. Training was conducted for 100K steps, consistent with the official implementations.

    \item EGC \cite{zhao2022probabilistic}: We utilized the \texttt{gcimpute} package\footnote{\url{https://github.com/udellgroup/gcimpute/tree/master}} to implement EGC. The model is free of hyperparameters.
\eed

\begin{table}[h]
  \centering
  \begin{tabular}{ccc}
    \toprule
    Tasks & Model & Description \\
    \midrule
    \texttt{Regression}
    & Random Forest & \makecell{Package: \texttt{sklearn.ensemble.RandomForestRegressor}, \\ setting: random\_state=0, defaulted values} \\
    \midrule
    \multirow{9}{*}{\texttt{Classification}}
    & Logistic Regression & \makecell{Package: \texttt{sklearn.linear\_model.LogisticRegression}, \\ setting: random\_state=0, max\_iter=1000, defaulted values} \\
    & Gaussian Navie Bayes & \makecell{Package: \texttt{sklearn.naive\_bayes.GaussianNB}, \\ setting: defaulted values} \\
    & K-Nearest Neighbors & \makecell{Package: \texttt{sklearn.neighbors.KNeighborsClassifier}, \\ setting: defaulted values} \\
    & Decision Tree & \makecell{Package: \texttt{sklearn.tree.DecisionTreeClassifier}, \\ setting: random\_state=0, defaulted values} \\
    & Random Forest & \makecell{Package: \texttt{sklearn.ensemble.RandomForestClassifier}, \\ setting: random\_state=0, defaulted values} \\
    \bottomrule
  \end{tabular}
  \caption{\textbf{Regressor and classifier used to evaluate synthetic data quality in machine learning utility.} The names of all parameters used in the description are consistent with those defined in corresponding packages.}
\label{tab:mlu_setting}
\end{table}

\subsection{Evaluation Procedure of Q1}

\textbf{Regression performance (SMAPE).}
\begin{enumerate}
    \item Train a synthesizer using the real training dataset.
    \item Generate a synthetic dataset with the same size as the real training dataset.
    \item Train a machine learning model (Random Forest regressor) using the synthetic dataset, where each continuous column serves as the regression target variable.
    \item Assess regression prediction performance by averaging the SMAPE values from the test dataset for each Random Forest regressor trained on the continuous columns.
\end{enumerate}

\textbf{Classification performance ($F_1$).}
\begin{enumerate}
    \item Train a synthesizer using the real training dataset.
    \item Generate a synthetic dataset with the same size as the real training dataset.
    \item Train machine learning models (Logistic Regression, Gaussian Naive Bayes, K-Nearest Neighbors classifier, Decision Tree classifier, and Random Forest classifier) using the synthetic dataset (see Table \ref{tab:data_description} for the classification target variable and refer to Table \ref{tab:mlu_setting} for detailed configuration).
    \item Assess classification prediction performance by averaging the $F_1$ values from the test dataset from five different classifiers.
\end{enumerate}

\textbf{Model selection performance (Model).}
\begin{enumerate}
    \item Train a synthesizer using the real training dataset.
    \item Generate a synthetic dataset with the same size as the real training dataset.
    \item Train machine learning models (Logistic Regression, Gaussian Naive Bayes, K-Nearest Neighbors classifier, Decision Tree classifier, and Random Forest classifier) using both the real training dataset and the synthetic dataset (see Table \ref{tab:data_description} for the classification target variable and refer to Table \ref{tab:mlu_setting} for detailed configuration).
    \item Evaluate the classification performance (AUROC) of all trained classifiers on the test dataset.
    \item Assess model selection performance by comparing the AUROC rank orderings of classifiers trained on the real training dataset and those trained on the synthetic dataset using Spearman’s Rank Correlation.
\end{enumerate}

\textbf{Feature selection performance (Feature).}
\begin{enumerate}
    \item Train a synthesizer using the real training dataset.
    \item Generate a synthetic dataset with the same size as the real training dataset.
    \item Train a Random Forest classifier using both the real training dataset and the synthetic dataset (see Table \ref{tab:data_description} for the classification target variable and refer to Table \ref{tab:mlu_setting} for detailed configuration).
    \item Determine the rank-ordering of important features for both classifiers.
    \item Assess feature selection performance by comparing the feature importance rank orderings of classifiers trained on the real training dataset and those trained on the synthetic dataset using Spearman’s Rank Correlation.
\end{enumerate}

\subsection{Evaluation Procedure of Q2}

\textbf{Regression performance (SMAPE).}
\begin{enumerate}
    \item For each random seed, we generate the mask and train MaCoDE using a masked training dataset (i.e., incomplete dataset).
    \item Generate a synthetic dataset with the same size as the real training dataset.
    \item Train a machine learning model (Random Forest regressor) using the synthetic dataset, where each continuous column serves as the regression target variable.
    \item Assess regression prediction performance by averaging the SMAPE values from the test dataset for each Random Forest regressor trained on the continuous columns.
\end{enumerate}

\textbf{Classification performance ($F_1$).}
\begin{enumerate}
    \item For each random seed, we generate the mask and train MaCoDE using a masked training dataset (i.e., incomplete dataset).
    \item Generate a synthetic dataset with the same size as the real training dataset.
    \item Train machine learning models (Logistic Regression, Gaussian Naive Bayes, K-Nearest Neighbors classifier, Decision Tree classifier, and Random Forest classifier) using the synthetic dataset (see Table \ref{tab:data_description} for the classification target variable and refer to Table \ref{tab:mlu_setting} for detailed configuration).
    \item Assess classification prediction performance by averaging the $F_1$ values from the test dataset from five different classifiers.
\end{enumerate}
    
\subsection{Evaluation of Q3}


\subsubsection{Related Works}

Data missingness is a common challenge in research and practical analysis, categorized into three primary missing mechanisms: (1) Missing Completely at Random (MCAR), (2) Missing at Random (MAR), and (3) Missing Not at Random (MNAR).

Under the \textbf{MCAR} mechanism, the reason for missingness has no relationship with any data, neither observed nor unobserved. In other words, the likelihood of data being missing is equal across all observations. The primary advantage of MCAR is that it does not introduce bias into the data analysis. However, despite this advantage, data missingness can still reduce the statistical power of the study because of the reduced sample size.

\textbf{MAR} occurs when the probability of missingness is related to the observed data but not the unobserved missing data. Essentially, even though the data is missing, the mechanism assumes that the missingness is explainable by other variables in the dataset. That is, the missingness can be modeled and imputed using the information available in the data, allowing for more accurate analyses despite the missingness. 

Lastly, if the missingness is not specified by either MCAR or MAR, it becomes \textbf{MNAR}. The MNAR is the most challenging mechanism, as it implies that the missingness is related to the unobserved data itself. In this case, the missing data is systematically different from the observed data, which introduces bias if not properly accounted for. For example, patients with severe symptoms may be less likely to report their health status, making their data missing. MNAR requires sophisticated statistical methods to address, as ignoring or improperly handling it can lead to biased and unreliable results.

Recent methods employ the deep generative model to estimate a joint distribution and generate samples for imputations. GAN-based imputers such as GAIN \cite{YOON2018GAIN} and MisGAN \cite{li2018learning} adopt an adversarial learning approach to generate both missing entries and masking vectors. Other approaches like VAEAC \cite{ivanov2018variational}, HI-VAE \cite{nazabal2020hivae}, and ReMasker \cite{du2024remasker} employ strategies that allow them to learn conditional distributions on arbitrary conditioning sets using the uniform masking strategy. MIWAE \cite{Mattei2019MIWAEDG}, based on the Importance Weighted Autoencoder \cite{Burda2015ImportanceWA}, demonstrates that under mild conditions and the MAR assumption, the target likelihood can be approximated regardless of the imputation function. Additionally, not-MIWAE \cite{ipsen2021notmiwae} extends MIWAE to handle cases where the data is under MNAR assumption by modeling missing entries as a latent variable. In parallel, the Optimal Transport (OT)-based method employs distributional matching, utilizing the 2-Wasserstein distance to compare distributions in both data and latent spaces \cite{Muzellec2020MissingDI, zhao2023TDM}. 
To handle mixed-type tabular datasets, \cite{zhao2022probabilistic} introduced EGC (extended Gaussian copula), which relies on a latent Gaussian distribution to support single and multiple imputations.

\subsubsection{Evaluation Metrics}

\begin{algorithm*}[h]
\caption{Evaluation procedure for multiple imputation \cite{Buuren2012FlexibleIO, Rubin1986MultipleIF}} 
\textbf{Input}: Complete dataset $D = \{x_i\}_{i=1}^n$ \\
\textbf{Output}: Bias, Coverage, and Confidence interval length
\begin{algorithmic}[1]
    \STATE Target estimand: $Q^* = 1/n \sum_{i=1}^n \mathbb{I}(x_i > \Bar{x})$, where $\Bar{x} = 1/n \sum_{i=1}^n x_i$ 
    \FOR {(random seeds) $s=1,\cdots,S$}
        \STATE Generate missing dataset $D^{(s)}$
        \STATE Training the imputation model using the dataset $D^{(s)}$
        \STATE Perform multiple imputation: $\hat{D}_m^{(s)}, m=1,\cdots,M$
        \FOR {$m = 1,2,\cdots,M$}
            \STATE $\hat{Q}_m^{(s)} = \frac{1}{n} \sum_{i=1}^n \mathbb{I}(x_i > \Bar{x})$, where $\hat{D}_m^{(s)} = \{x_i\}_{i=1}^n$ and $\Bar{x} = \frac{1}{n} \sum_{i=1}^n x_i$ 
            \STATE $\hat{U}_m^{(s)} = \hat{Q}_m^{(s)}(1 - \hat{Q}_m^{(s)}) / n$
        \ENDFOR 
        \STATE $\hat{Q}^{(s)} = \frac{1}{M} \sum_{m=1}^M \hat{Q}_m^{(s)}$
        \STATE $\hat{U}^{(s)} = \frac{1}{M} \sum_{m=1}^M \hat{U}_m^{(s)} + (M + 1)/M \cdot \frac{1}{M-1} \sum_{m=1}^M (\hat{U}_m^{(s)} - \Bar{U}^{(s)})^2$, where $\Bar{U}^{(s)} = \frac{1}{M} \sum_{m=1}^M \hat{U}_m^{(s)}$
    \ENDFOR 
    \STATE Bias: $\frac{1}{S} \sum_{s=1}^S |\hat{Q}^{(s)} - Q^*|$
    \STATE Coverage: $\frac{1}{S} \sum_{s=1}^S \mathbb{I}(Q^* \in (\hat{Q}^{(s)} \pm 1.96 \cdot \sqrt{\hat{U}^{(s)}}))$
    \STATE Confidence interval length: $\frac{1}{S} \sum_{s=1}^S 2 \cdot 1.96 \cdot \sqrt{\hat{U}^{(s)}}$
\end{algorithmic} \label{alg:multiple}
\end{algorithm*}


For Q3, we selected the following multiple imputation models that can handle heterogeneous tabular datasets: MICE \cite{Buuren2011MICEMI}, GAIN \cite{YOON2018GAIN}, missMDA \cite{missMDA}, VAEAC \cite{ivanov2018variational}, MIWAE \cite{Mattei2019MIWAEDG}, not-MIWAE \cite{ipsen2021notmiwae}, and EGC \cite{zhao2022probabilistic}. 

We assess the effectiveness of multiple imputations by employing interval inference for the population mean, which was proposed by Rubin \cite{Rubin1986MultipleIF}. However, since obtaining the population mean is not feasible, we utilize the sample column-wise mean of the complete dataset as a parameter of interest \cite{Lee2021EvaluationOM, Zhang2023UsingMI}. The evaluation procedure for multiple imputations is outlined in Algorithm \ref{alg:multiple}, and we utilize \texttt{torch.random.manual\_seed()} to set seeds in performing multiple imputations. We report the mean and standard error of bias, coverage, and confidence interval length across 10 different random seeds, all continuous columns, and datasets \cite{Buuren2012FlexibleIO}. Note that we do not split the data into training and test sets \cite{ipsen2021notmiwae}.

\begin{remark}
We acknowledge that existing imputation methods have been evaluated using RMSE (root mean square error), a metric for assessing single imputation methods \cite{ipsen2021notmiwae, Mattei2019MIWAEDG, nazabal2020hivae, Muzellec2020MissingDI, zhao2023TDM, jarrett22hyperimpute}. However, since our objective focuses on distributional learning rather than recovering missing values, we adopt the evaluation procedure proposed in \cite{Rubin1986MultipleIF, Buuren2012FlexibleIO}, better suited for assessing multiple imputation methods.
\end{remark}
    
Following \cite{Muzellec2020MissingDI, jarrett22hyperimpute, zhao2023TDM}, we generate the missing value mask for each dataset with three mechanisms in four settings. (MCAR) In the MCAR setting, each value is masked according to the realization of a Bernoulli random variable with a fixed parameter. (MAR) In the MAR setting, for each experiment, a fixed subset of variables that cannot have missing values is sampled. Then, the remaining variables have missing values according to a logistic model with random weights, which takes the non-missing variables as inputs. A bias term is fitted using line search to attain the desired proportion of missing values. (MNAR) Finally, two different mechanisms are implemented in the MNAR setting. The first, MNARL, is identical to the previously described MAR mechanism, but the inputs of the logistic model are then masked by an MCAR mechanism. Hence, the logistic model’s outcome depends on missing values. The second mechanism, MNARQ, samples a subset of variables whose values in the lower and upper $p$th percentiles are masked according to a Bernoulli random variable, and the values in-between are left not missing.

\begin{table}[h]
    \centering
    \begin{tabular}{lrrrrrrrrrrrrrrrr}
    \toprule
    Model & Bias $\downarrow$ & Coverage & Width $\downarrow$ \\
    \midrule
    MICE & $.010_{\pm .001}$ & $.845_{\pm .019}$ & $\mathbf{.040}_{\pm .002}$\\
    GAIN & $.019_{\pm .002}$ & $.633_{\pm .033}$ & $\mathbf{.040}_{\pm .002}$  \\
    missMDA & $.015_{\pm .001}$ & $.700_{\pm .022}$ & $.043_{\pm .002}$\\
    VAEAC & $\underline{.008}_{\pm .001}$ & $.905_{\pm .016}$ & $\mathbf{.040}_{\pm .002}$\\
    MIWAE & $\mathbf{.006}_{\pm .000}$ & $\underline{.952}_{\pm .012}$ & $.043_{\pm .002}$\\
    not-MIWAE & $\mathbf{.006}_{\pm .000}$ & $\mathbf{.949}_{\pm .012}$ & $\underline{.042}_{\pm .002}$\\
    EGC & $\mathbf{.006}_{\pm .000}$ & ${.996}_{\pm .004}$ & $.058_{\pm .002}$\\
        \midrule
    MaCoDE(MAR) & $\mathbf{.006}_{\pm .000}$ & $.963_{\pm .009}$ & $.051_{\pm .003}$\\
        \bottomrule
      \end{tabular}
      \caption{\textbf{Q3}: Multiple imputation under MAR at 0.3 missingness. The means and standard errors of the mean across 5 datasets and 10 repeated experiments are reported. $\downarrow$ denotes lower is better. Coverage close to 0.95 indicates better performance. The best value is bolded, and the second best is underlined.}
    \label{tab:impute}
\end{table}

\subsubsection{Results}

The missing data mechanism within the parentheses refers to the mechanism applied to the dataset on which MaCoDE was trained.
Table \ref{tab:impute} indicates that MaCoDE consistently exhibits competitive performance against all baseline models across metrics assessing multiple imputation performances, including bias, coverage, and confidence interval length. This suggests our proposed approach can support multiple imputations for deriving statistically valid inferences from missing data with the MAR mechanism.




\subsection{Additional Experiments}

\subsubsection{Privacy Preservability}
\label{app:privacy_results}


\bed
    \item \textbf{Evaluation metrics:} \textit{The $k$-anonymity property} \cite{Sweeney2002kAnonymityAM} is a measure used to assess the level of privacy protection in synthetic data. It ensures that each individual's information in the dataset cannot be distinguished from that of at least $k-1$ other individuals. In other words, each record in the dataset has at least $k-1$ similar records in terms of quasi-identifiers, attributes that could potentially identify a subject. A higher $k$ value implies a higher level of anonymity and better privacy preservation.
    
    \textit{DCR (Distance to Closest Record)} \cite{Park2018DataSB, zhao2021ctabgan} is defined as the distances between all real training and synthetic samples. A higher DCR value indicates more effective privacy preservation, indicating a lack of overlap between the real training data and the synthetic samples. Conversely, an excessively large DCR score suggests a lower quality of the generated synthetic dataset. Therefore, the DCR metric provides insights into both the privacy-preserving capability and the quality of the synthetic dataset.

    \textit{Attribute disclosure} \cite{choi2017generating, Matwin2015ARO} refers to the situation where attackers can uncover additional covariates of a record by leveraging a subset of covariates they already possess, along with similar records from the synthetic dataset. To quantify the extent to which attackers can accurately identify these additional covariates, we employ classification metrics. Higher attribute disclosure metrics indicate an increased risk of privacy leakage, implying that attackers can precisely infer unknown variables. In terms of privacy concerns, attribute disclosure can be considered a more significant issue than membership inference attacks, as attackers are assumed to have access to only a subset of covariates for a given record.
    
    \item \textbf{Evaluation procedure:} We evaluate the $k$-anonymity following the approach described in \cite{qian2023synthcity} \footnote{\url{https://github.com/vanderschaarlab/synthcity}}. Additionally, similar to \cite{zhao2021ctabgan}, we define DCR as the $5^{th}$ percentile of the $L_2$ distances between all real training samples and synthetic samples. Since DCR relies on $L_2$ distance and continuous variables, it is computed solely using continuous variables. We assess attribute disclosure using the methodology outlined in \cite{choi2017generating}.

    \item \textbf{Result:} The right panel of Figure \ref{fig:privacy_control} demonstrates MaCoDE's capability to regulate the privacy level by adjusting the temperature parameter $\tau$. Simultaneously, the left panel illustrates that the quality of synthetic data, measured by feature selection performance, remains notable even with increasing privacy levels from $\tau=1$ to $\tau=3$. However, increasing $\tau$ beyond 3 leads to declining feature selection performance compared to other models despite DCR remaining competitive. For additional results on other metrics related to the trade-off between privacy level and synthetic data quality as $\tau$ varies, please refer to Table \ref{tab:performance_withprivacy} and Table \ref{tab:privacy}.
\eed

\begin{figure}[ht]
    \centering
    \includegraphics[width=0.6\textwidth]{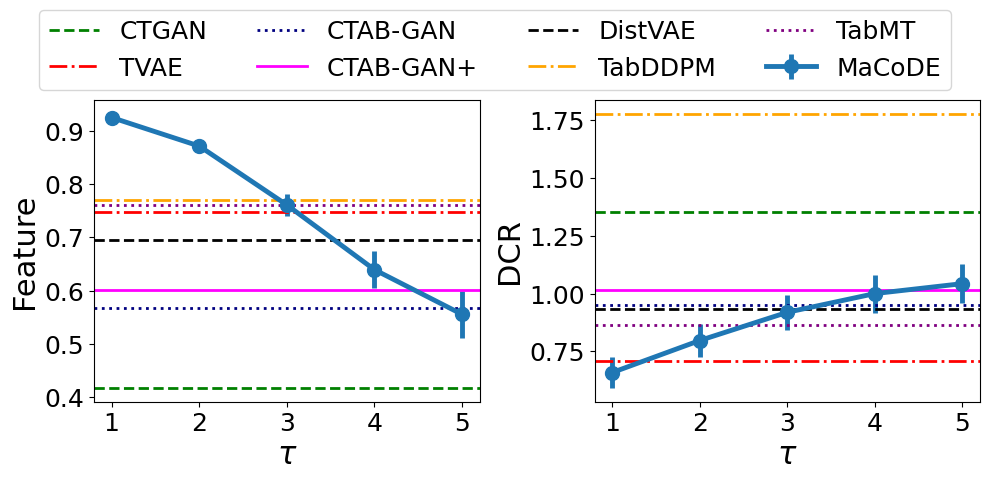}
    \caption{\textbf{Trade-off between privacy and quality.} Left: feature selection performance (synthetic data quality). Right: DCR (privacy preservability). The means and standard errors of the mean across 10 datasets and 10 repeated experiments are reported. Error bars represent the standard errors of the mean.}
    \label{fig:privacy_control}
\end{figure}

\begin{table}[ht]
  \centering
  \resizebox{0.99\textwidth}{!}{
  \begin{tabular}{lrrrrrrrrrrrrrrrr}
    \toprule
    & \multicolumn{4}{c}{Statistical fidelity} & \multicolumn{4}{c}{Machine learning utility} \\
    \cmidrule(lr){2-5} \cmidrule(lr){6-9}
    Model & KL $\downarrow$ & GoF $\downarrow$ & MMD $\downarrow$ & WD $\downarrow$ & SMAPE $\downarrow$ & $F_1$ $\uparrow$ & Model $\uparrow$ & Feature $\uparrow$\\
    \midrule
Baseline & $.016_{\pm .002}$ & $.029_{\pm .002}$ & $.002_{\pm .000}$ & $1.019_{\pm .156}$ & $.107_{\pm .008}$ & $.686_{\pm .023}$ & $.887_{\pm .018}$ & $.956_{\pm .005}$\\
\midrule
CTGAN & $.221_{\pm .014}$ & $.561_{\pm .046}$ & $.094_{\pm .007}$ & $6.435_{\pm 1.011}$ & $.256_{\pm .016}$ & $.411_{\pm .027}$ & $.208_{\pm .048}$ & $.417_{\pm .043}$\\
TVAE & $.066_{\pm .003}$ & $.119_{\pm .005}$ & $.016_{\pm .001}$ & $1.631_{\pm .173}$ & $.192_{\pm .011}$ & $.608_{\pm .021}$ & $.486_{\pm .041}$ & $.747_{\pm .027}$\\
CTAB-GAN & $.116_{\pm .008}$ & $.196_{\pm.025}$ & $.044_{\pm .004}$ & $3.327_{\pm.460}$ & $.218_{\pm.012}$ & $.524_{\pm.026}$ & $.263_{\pm.042}$ & $.568_{\pm .041}$\\
CTAB-GAN+ & $.136_{\pm .018}$ & $.144_{\pm.010}$ & $.054_{\pm .007}$ & $3.971_{\pm.772}$ & $.226_{\pm.017}$ & $.530_{\pm.020}$ & $.227_{\pm.048}$ & $.601_{\pm.041}$\\
DistVAE & $.059_{\pm .007}$ & $.070_{\pm .004}$ & $.016_{\pm .001}$ & $2.272_{\pm .282}$ & $.226_{\pm .017}$ & $.588_{\pm .021}$ & $.194_{\pm .048}$ & $.695_{\pm .030}$\\
TabDDPM & $.696_{\pm .117}$ & $.374_{\pm .087}$ & $.057_{\pm .011}$ & $42.916_{\pm 8.127}$ & $.161_{\pm .011}$ & $.576_{\pm .022}$ & $.507_{\pm .039}$ & $.770_{\pm .027}$ \\
TabMT & $.011_{\pm .001}$ & $.035_{\pm .003}$ & $.012_{\pm .001}$ & $2.299_{\pm .346}$ & $.188_{\pm .013}$ & $.622_{\pm .024}$ & $.528_{\pm .039}$ & $.761_{\pm .028}$ \\
\midrule
MaCoDE($\tau=1$) & $.034_{\pm .004}$ & $.072_{\pm .004}$ & $.007_{\pm .001}$ & $1.630_{\pm .245}$ & $.158_{\pm .010}$ & $.635_{\pm .023}$ & $.599_{\pm .035}$ & $.925_{\pm .007}$\\
MaCoDE($\tau=2$) & $.047_{\pm .004}$ & $.117_{\pm .009}$ & $.021_{\pm .002}$ & $2.450_{\pm .302}$ & $.177_{\pm .012}$ & $.601_{\pm .024}$ & $.432_{\pm .038}$ & $.871_{\pm .011}$\\
MaCoDE($\tau=3$) & $.071_{\pm .006}$ & $.290_{\pm .046}$ & $.035_{\pm .003}$ & $3.191_{\pm .354}$ & $.196_{\pm .013}$ & $.562_{\pm .024}$ & $.305_{\pm .041}$ & $.761_{\pm .020}$\\
MaCoDE($\tau=4$) & $.088_{\pm .008}$ & $.555_{\pm .109}$ & $.043_{\pm .003}$ & $3.601_{\pm .383}$ & $.210_{\pm .014}$ & $.528_{\pm .024}$ & $.128_{\pm .041}$ & $.639_{\pm .035}$\\
MaCoDE($\tau=5$) & $.099_{\pm .010}$ & $.847_{\pm .181}$ & $.045_{\pm .003}$ & $3.822_{\pm .402}$ & $.219_{\pm .014}$ & $.501_{\pm .025}$ & $.065_{\pm .047}$ & $.555_{\pm .044}$\\
    \bottomrule
  \end{tabular}
  }
  \caption{\textbf{Trade-off between privacy and quality} (statistical fidelity and machine learning utility). The means and standard errors of the mean across 10 datasets and 10 repeated experiments are reported. `Baseline' refers to the result obtained using half of the real training dataset. $\uparrow$ ($\downarrow$) denotes higher (lower) is better.}
\label{tab:performance_withprivacy}
\end{table}

\begin{table}[ht]
  \centering
  \resizebox{0.65\linewidth}{!}{
  \begin{tabular}{lrrrrrrrrrrrrrrrr}
    \toprule
    Model & $k$-anonimity(\%) $\uparrow$ & DCR $\uparrow$ & AD $\downarrow$ \\
    \midrule
Baseline & $1.151_{\pm 0.089}$ & $0.501_{\pm 0.051}$ & $0.631_{\pm 0.023}$\\
\midrule
CTGAN & $2.413_{\pm 0.139}$ & $1.355_{\pm 0.152}$ & $0.398_{\pm 0.024}$\\ 
TVAE & $1.648_{\pm 0.116}$ & $ 0.709_{\pm 0.056}$ & $0.583_{\pm 0.023}$\\ 
CTAB-GAN & $2.048_{\pm 0.126}$ & $0.951_{\pm 0.076}$ & $0.514_{\pm 0.025}$\\ 
CTAB-GAN+ & $2.215_{\pm 0.121}$ & $1.016_{\pm 0.113}$ & $0.471_{\pm 0.018}$\\
DistVAE & $2.345_{\pm 0.136}$ & $0.932_{\pm 0.071}$ & $0.540_{\pm 0.019}$\\ 
TabDDPM & $1.249_{\pm 0.108}$ & $1.779_{\pm 0.391}$ & $0.567_{\pm 0.024}$\\ 
TabMT & $1.656_{\pm 0.135}$ & $ 0.862_{\pm 0.072}$ & $0.566_{\pm 0.021}$\\ 
\midrule
MaCoDE($\tau=1$) & $1.405_{\pm 0.104}$ & $0.658_{\pm 0.067}$ & $0.589_{\pm 0.022}$\\
MaCoDE($\tau=2$) & $1.656_{\pm 0.109}$ & $0.797_{\pm 0.070}$ & $0.547_{\pm 0.022}$\\
MaCoDE($\tau=3$) & $1.959_{\pm 0.147}$ & $0.919_{\pm 0.077}$ & $0.508_{\pm 0.022}$\\
MaCoDE($\tau=4$) & $2.075_{\pm 0.159}$ & $1.000_{\pm 0.082}$ & $0.476_{\pm 0.022}$\\
MaCoDE($\tau=5$) & $2.262_{\pm 0.166}$ & $1.043_{\pm 0.085}$ & $0.457_{\pm 0.022}$\\
    \bottomrule
  \end{tabular}
  }
  \caption{\textbf{Privacy preservability}. The means and standard errors of the mean across 10 datasets and 10 repeated experiments are reported. `Baseline' refers to the result obtained using half of the real training dataset. $\uparrow$ ($\downarrow$) denotes higher (lower) is better.}
\label{tab:privacy}
\end{table}

\clearpage
\begin{table}[ht]
\begin{minipage}[t]{0.50\textwidth}
  \vspace{0pt}
  \resizebox{0.99\columnwidth}{!}{
  \begin{tabular}{lrrrrrrrrrrrrrrrr}
    \toprule
Dataset & & \texttt{abalone} &\\
    \midrule
Model & $k$-anonimity(\%) $\uparrow$ & DCR $\uparrow$ & AD $\downarrow$ \\
    \midrule	
CTGAN& $3.634_{\pm 0.318}$ & $0.423_{\pm 0.012}$ & $0.346_{\pm 0.009}$  \\
TVAE& $2.029_{\pm 0.234}$ & $0.333_{\pm 0.010}$ & $0.385_{\pm 0.018}$  \\
CTAB-GAN& $2.769_{\pm 0.315}$ & $0.449_{\pm 0.027}$ & $0.312_{\pm 0.023}$  \\
CTAB-GAN+& $2.718_{\pm 0.285}$ & $0.311_{\pm 0.010}$ & $0.323_{\pm 0.010}$  \\
DistVAE& $2.074_{\pm 0.149}$ & $0.316_{\pm 0.006}$ & $0.349_{\pm 0.009}$  \\
TabDDPM& $0.233_{\pm 0.047}$ & $0.145_{\pm 0.003}$ & $0.371_{\pm 0.016}$ \\
TabMT& $1.595_{\pm 0.617}$ & $0.423_{\pm 0.007}$ & $0.337_{\pm 0.010}$  \\
    \midrule
MaCoDE& $1.182_{\pm 0.292}$ & $0.148_{\pm 0.002}$ & $0.368_{\pm 0.014}$\\
    \bottomrule
  \end{tabular}
  }
\end{minipage}
\begin{minipage}[t]{0.50\textwidth} 
  \vspace{0pt}
  \resizebox{0.99\columnwidth}{!}{
  \begin{tabular}{lrrrrrrrrrrrrrrrr}
    \toprule
Dataset & & \texttt{banknote} &\\
    \midrule
Model & $k$-anonimity(\%) $\uparrow$ & DCR $\uparrow$ & AD $\downarrow$ \\
    \midrule	
CTGAN& $2.862_{\pm 0.146}$ & $0.341_{\pm 0.020}$ & $0.763_{\pm 0.033}$  \\
TVAE& $2.607_{\pm 0.305}$ & $0.220_{\pm 0.007}$ & $0.920_{\pm 0.021}$  \\
CTAB-GAN& $3.118_{\pm 0.202}$ & $0.275_{\pm 0.008}$ & $0.847_{\pm 0.025}$  \\
CTAB-GAN+& $2.799_{\pm 0.151}$ & $0.279_{\pm 0.012}$ & $0.560_{\pm 0.038}$  \\
DistVAE& $3.081_{\pm 0.267}$ & $0.290_{\pm 0.010}$ & $0.757_{\pm 0.027}$  \\
TabDDPM& $0.985_{\pm 0.180}$ & $0.168_{\pm 0.007}$ & $0.997_{\pm 0.003}$\\
TabMT& $2.662_{\pm 0.262}$ & $0.257_{\pm 0.009}$ & $0.910_{\pm 0.015}$  \\
    \midrule
MaCoDE& $2.552_{\pm 0.255}$ & $0.076_{\pm 0.005}$ & $0.930_{\pm 0.016}$\\
    \bottomrule
    \end{tabular}
    }
\end{minipage}

\vspace{3mm}

\begin{minipage}[t]{0.50\textwidth}
  \vspace{0pt}
  \resizebox{0.99\columnwidth}{!}{
  \begin{tabular}{lrrrrrrrrrrrrrrrr}
    \toprule
Dataset & & \texttt{breast} &\\
    \midrule
Model & $k$-anonimity(\%) $\uparrow$ & DCR $\uparrow$ & AD $\downarrow$ \\
    \midrule	
CTGAN& $1.582_{\pm 0.383}$ & $5.536_{\pm 0.167}$ & $0.446_{\pm 0.050}$  \\
TVAE& $0.813_{\pm 0.239}$ & $2.148_{\pm 0.026}$ & $0.900_{\pm 0.042}$  \\
CTAB-GAN& $0.857_{\pm 0.198}$ & $2.807_{\pm 0.078}$ & $0.850_{\pm 0.037}$  \\
CTAB-GAN+& $1.473_{\pm 0.289}$ & $4.025_{\pm 0.315}$ & $0.596_{\pm 0.078}$  \\
DistVAE& $2.264_{\pm 0.390}$ & $2.666_{\pm 0.060}$ & $0.810_{\pm 0.036}$  \\
TabDDPM& $2.835_{\pm 0.301}$ & $12.733_{\pm 0.896}$ & $0.740_{\pm 0.089}$\\
TabMT& $0.967_{\pm 0.164}$ & $2.718_{\pm 0.054}$ & $0.825_{\pm 0.041}$  \\
    \midrule
MaCoDE& $0.527_{\pm 0.140}$ & $2.264_{\pm 0.058}$ & $0.906_{\pm 0.032}$\\
    \bottomrule
  \end{tabular}
  }
\end{minipage}
\begin{minipage}[t]{0.50\textwidth} 
  \vspace{0pt}
  \resizebox{0.99\columnwidth}{!}{
  \begin{tabular}{lrrrrrrrrrrrrrrrr}
    \toprule
Dataset & & \texttt{concrete} &\\
    \midrule
Model & $k$-anonimity(\%) $\uparrow$ & DCR $\uparrow$ & AD $\downarrow$ \\
    \midrule	
CTGAN& $3.750_{\pm 0.305}$ & $1.236_{\pm 0.076}$ & $0.156_{\pm 0.040}$  \\
TVAE& $2.464_{\pm 0.386}$ & $0.681_{\pm 0.018}$ & $0.340_{\pm 0.054}$  \\
CTAB-GAN& $3.519_{\pm 0.216}$ & $0.973_{\pm 0.044}$ & $0.310_{\pm 0.054}$  \\
CTAB-GAN+& $4.066_{\pm 0.201}$ & $0.972_{\pm 0.034}$ & $0.310_{\pm 0.033}$  \\
DistVAE& $4.235_{\pm 0.231}$ & $1.012_{\pm 0.037}$ & $0.358_{\pm 0.051}$  \\
TabDDPM& $1.663_{\pm 0.253}$ & $0.425_{\pm 0.011}$ & $0.388_{\pm 0.044}$\\
TabMT& $4.029_{\pm 0.228}$ & $0.949_{\pm 0.038}$ & $0.308_{\pm 0.053}$  \\
    \midrule
MaCoDE & $2.864_{\pm 0.293}$ & $0.239_{\pm 0.019}$ & $0.348_{\pm 0.043}$\\
    \bottomrule
    \end{tabular}
    }
\end{minipage}

\vspace{3mm}

\begin{minipage}[t]{0.50\textwidth}
  \vspace{0pt}
  \resizebox{0.99\columnwidth}{!}{
  \begin{tabular}{lrrrrrrrrrrrrrrrr}
    \toprule
Dataset & & \texttt{covtype} &\\
    \midrule
Model & $k$-anonimity(\%) $\uparrow$ & DCR $\uparrow$ & AD $\downarrow$ \\
    \midrule	
CTGAN& $0.215_{\pm 0.000}$ & $0.451_{\pm 0.003}$ & $0.490_{\pm 0.023}$  \\
TVAE& $0.215_{\pm 0.000}$ & $0.415_{\pm 0.002}$ & $0.639_{\pm 0.009}$  \\
CTAB-GAN& $0.215_{\pm 0.000}$ & $0.418_{\pm 0.004}$ & $0.579_{\pm 0.004}$  \\
CTAB-GAN+& $0.215_{\pm 0.000}$ & $0.353_{\pm 0.001}$ & $0.645_{\pm 0.001}$  \\
DistVAE& $0.215_{\pm 0.000}$ & $0.566_{\pm 0.002}$ & $0.612_{\pm 0.002}$  \\
TabDDPM& $0.211_{\pm 0.002}$ & $0.313_{\pm 0.001}$ & $0.653_{\pm 0.002}$\\
TabMT& $0.215_{\pm 0.000}$ & $0.409_{\pm 0.001}$ & $0.671_{\pm 0.003}$  \\
    \midrule
MaCoDE& $0.215_{\pm 0.000}$ & $0.266_{\pm 0.000}$ & $0.696_{\pm 0.001}$\\
    \bottomrule
  \end{tabular}
  }
\end{minipage}
\begin{minipage}[t]{0.50\textwidth} 
  \vspace{0pt}
  \resizebox{0.99\columnwidth}{!}{
  \begin{tabular}{lrrrrrrrrrrrrrrrr}
    \toprule
Dataset & & \texttt{kings} &\\
    \midrule
Model & $k$-anonimity(\%) $\uparrow$ & DCR $\uparrow$ & AD $\downarrow$ \\
    \midrule	
CTGAN& $1.243_{\pm 0.168}$ & $0.502_{\pm 0.006}$ & $0.601_{\pm 0.004}$ \\
TVAE& $0.848_{\pm 0.066}$ & $0.361_{\pm 0.003}$ & $0.685_{\pm 0.003}$  \\
CTAB-GAN& $1.158_{\pm 0.103}$ & $0.461_{\pm 0.004}$ & $0.631_{\pm 0.003}$  \\
CTAB-GAN+& $1.028_{\pm 0.126}$ & $0.411_{\pm 0.006}$ & $0.655_{\pm 0.003}$  \\
DistVAE& $0.699_{\pm 0.113}$ & $0.488_{\pm 0.002}$ & $0.658_{\pm 0.004}$  \\
TabDDPM& $0.236_{\pm 0.032}$ & $0.325_{\pm 0.003}$ & $0.668_{\pm 0.005}$\\
TabMT& $0.396_{\pm 0.103}$ & $0.427_{\pm 0.003}$ & $0.652_{\pm 0.003}$  \\
    \midrule
MaCoDE& $0.445_{\pm 0.151}$ & $0.301_{\pm 0.005}$ & $0.677_{\pm 0.004}$\\
    \bottomrule
    \end{tabular}
    }
\end{minipage}

\vspace{3mm}

\begin{minipage}[t]{0.50\textwidth}
  \vspace{0pt}
  \resizebox{0.99\columnwidth}{!}{
  \begin{tabular}{lrrrrrrrrrrrrrrrr}
    \toprule
Dataset & & \texttt{letter} &\\
    \midrule
Model & $k$-anonimity(\%) $\uparrow$ & DCR $\uparrow$ & AD $\downarrow$ \\
    \midrule	
CTGAN& $3.520_{\pm 0.180}$ & $1.583_{\pm 0.042}$ & $0.089_{\pm 0.009}$  \\
TVAE& $2.657_{\pm 0.232}$ & $0.779_{\pm 0.009}$ & $0.274_{\pm 0.008}$  \\
CTAB-GAN& $3.343_{\pm 0.267}$ & $1.438_{\pm 0.016}$ & $0.100_{\pm 0.006}$  \\
CTAB-GAN+& $3.112_{\pm 0.253}$ & $1.218_{\pm 0.008}$ & $0.207_{\pm 0.004}$  \\
DistVAE& $1.959_{\pm 0.265}$ & $1.378_{\pm 0.006}$ & $0.273_{\pm 0.005}$  \\
TabDDPM& $1.435_{\pm 0.173}$ & $0.744_{\pm 0.005}$ & $0.260_{\pm 0.008}$\\
TabMT& $2.112_{\pm 0.236}$ & $0.822_{\pm 0.008}$ & $0.402_{\pm 0.005}$  \\
    \midrule
MaCoDE & $1.819_{\pm 0.089}$ & $1.174_{\pm 0.011}$ & $0.380_{\pm 0.007}$\\
    \bottomrule
  \end{tabular}
  }
\end{minipage}
\begin{minipage}[t]{0.50\textwidth} 
  \vspace{0pt}
  \resizebox{0.99\columnwidth}{!}{
  \begin{tabular}{lrrrrrrrrrrrrrrrr}
    \toprule
Dataset & & \texttt{loan} &\\
    \midrule
Model & $k$-anonimity(\%) $\uparrow$ & DCR $\uparrow$ & AD $\downarrow$ \\
    \midrule	
CTGAN& $2.450_{\pm 0.244}$ & $0.241_{\pm 0.007}$ & $0.660_{\pm 0.012}$  \\
TVAE& $2.442_{\pm 0.338}$ & $0.160_{\pm 0.004}$ & $0.716_{\pm 0.008}$  \\
CTAB-GAN& $2.407_{\pm 0.156}$ & $0.208_{\pm 0.004}$ & $0.687_{\pm 0.010}$  \\
CTAB-GAN+& $2.570_{\pm 0.159}$ & $0.216_{\pm 0.015}$ & $0.577_{\pm 0.005}$  \\
DistVAE& $2.348_{\pm 0.092}$ & $0.238_{\pm 0.004}$ & $0.683_{\pm 0.002}$  \\
TabDDPM& $0.992_{\pm 0.107}$ & $0.123_{\pm 0.002}$ & $0.699_{\pm 0.006}$\\
TabMT& $1.895_{\pm 0.126}$ & $0.190_{\pm 0.003}$ & $0.687_{\pm 0.003}$  \\
    \midrule
MaCoDE & $1.942_{\pm 0.212}$ & $0.127_{\pm 0.003}$ & $0.697_{\pm 0.007}$\\
    \bottomrule
    \end{tabular}
    }
\end{minipage}

\vspace{3mm}

\begin{minipage}[t]{0.50\textwidth}
  \vspace{0pt}
  \resizebox{0.99\columnwidth}{!}{
  \begin{tabular}{lrrrrrrrrrrrrrrrr}
    \toprule
Dataset & & \texttt{redwine} &\\
    \midrule
Model & $k$-anonimity(\%) $\uparrow$ & DCR $\uparrow$ & AD $\downarrow$ \\
    \midrule	
CTGAN& $2.291_{\pm 0.411}$ & $1.794_{\pm 0.032}$ & $0.192_{\pm 0.034}$  \\
TVAE& $1.009_{\pm 0.325}$ & $0.976_{\pm 0.016}$ & $0.500_{\pm 0.041}$  \\
CTAB-GAN& $1.345_{\pm 0.140}$ & $1.191_{\pm 0.027}$ & $0.454_{\pm 0.027}$  \\
CTAB-GAN+& $2.056_{\pm 0.208}$ & $1.227_{\pm 0.027}$ & $0.448_{\pm 0.024}$  \\
DistVAE& $3.815_{\pm 0.220}$ & $1.171_{\pm 0.016}$ & $0.436_{\pm 0.026}$  \\
TabDDPM& $2.697_{\pm 0.289}$ & $1.330_{\pm 0.046}$ & $0.480_{\pm 0.025}$ \\
TabMT& $1.071_{\pm 0.205}$ & $1.236_{\pm 0.024}$ & $0.420_{\pm 0.021}$  \\
    \midrule
MaCoDE & $1.016_{\pm 0.086}$ & $0.940_{\pm 0.018}$ & $0.454_{\pm 0.026}$\\
    \bottomrule
  \end{tabular}
  }
\end{minipage}
\begin{minipage}[t]{0.50\textwidth} 
  \vspace{0pt}
  \resizebox{0.99\columnwidth}{!}{
  \begin{tabular}{lrrrrrrrrrrrrrrrr}
    \toprule
Dataset & & \texttt{whitewine} &\\
    \midrule
Model & $k$-anonimity(\%) $\uparrow$ & DCR $\uparrow$ & AD $\downarrow$ \\
    \midrule	
CTGAN& $2.580_{\pm 0.425}$ & $1.445_{\pm 0.018}$ & $0.239_{\pm 0.020}$  \\
TVAE& $1.399_{\pm 0.222}$ & $1.016_{\pm 0.005}$ & $0.475_{\pm 0.032}$  \\
CTAB-GAN& $1.746_{\pm 0.287}$ & $1.292_{\pm 0.027}$ & $0.373_{\pm 0.022}$  \\
CTAB-GAN+& $2.116_{\pm 0.171}$ & $1.145_{\pm 0.010}$ & $0.385_{\pm 0.018}$  \\
DistVAE& $2.762_{\pm 0.184}$ & $1.189_{\pm 0.007}$ & $0.461_{\pm 0.016}$  \\
TabDDPM& $0.896_{\pm 0.064}$ & $1.043_{\pm 0.009}$ & $0.441_{\pm 0.009}$\\
TabMT& $1.618_{\pm 0.303}$ & $1.192_{\pm 0.013}$ & $0.446_{\pm 0.014}$  \\
    \midrule
MaCoDE & $1.539_{\pm 0.183}$ & $0.988_{\pm 0.013}$ & $0.446_{\pm 0.016}$\\
    \bottomrule
    \end{tabular}
    }
\end{minipage}
\caption{\textbf{Privacy preservability} for each dataset. The means and standard errors of the mean across 10 repeated experiments are reported. $\uparrow$ ($\downarrow$) denotes higher (lower) is better.}
\end{table}

\clearpage
\subsubsection{Sensitivity Analysis}
\label{app:sensitivity_results}

\bed
    \item \textbf{Evaluation procedure:} We generated missing values into the \texttt{kings} dataset at rates of 0.1, 0.3, 0.5, and 0.7 for each missing mechanism and proceeded to train the model. Subsequently, we evaluated the machine learning utility using metrics such as SMAPE and F1-score, along with assessing the multiple imputation performance of the model using the same methodology described earlier.
    
\begin{figure}[ht]
    \centering
    \includegraphics[width=0.90\linewidth]{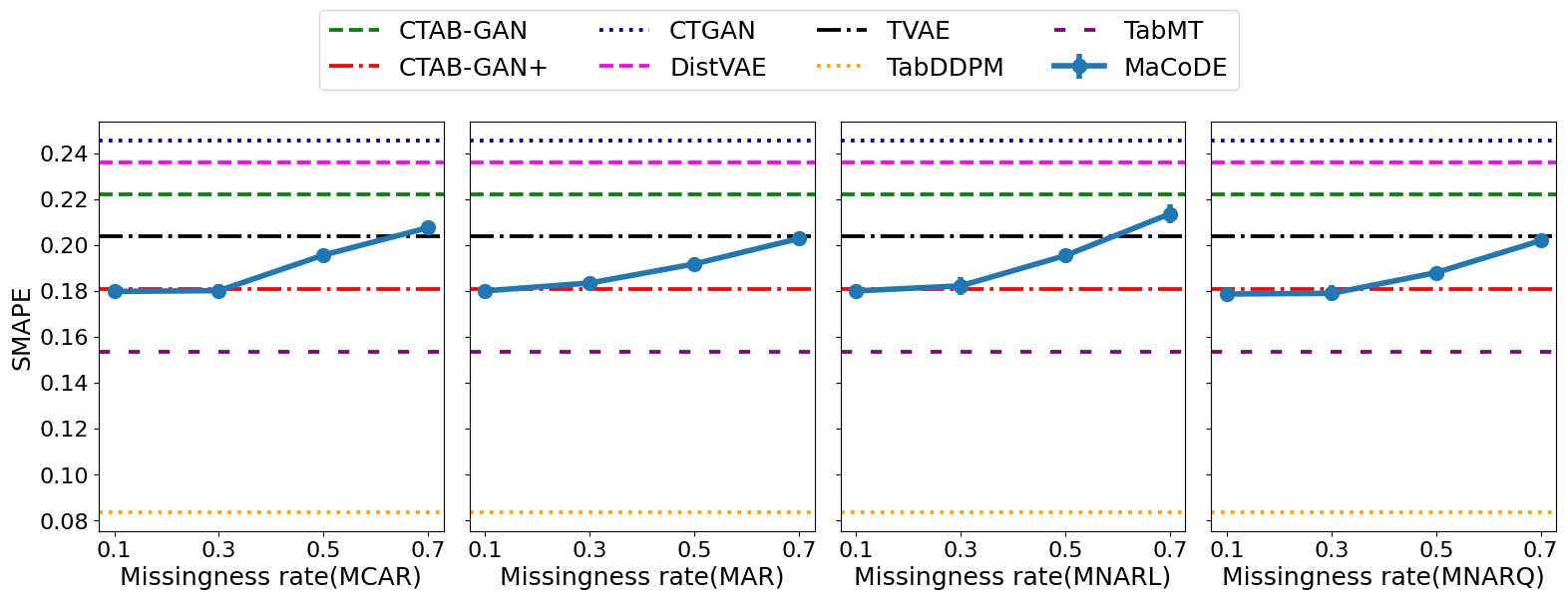}
    \includegraphics[width=0.90\linewidth]{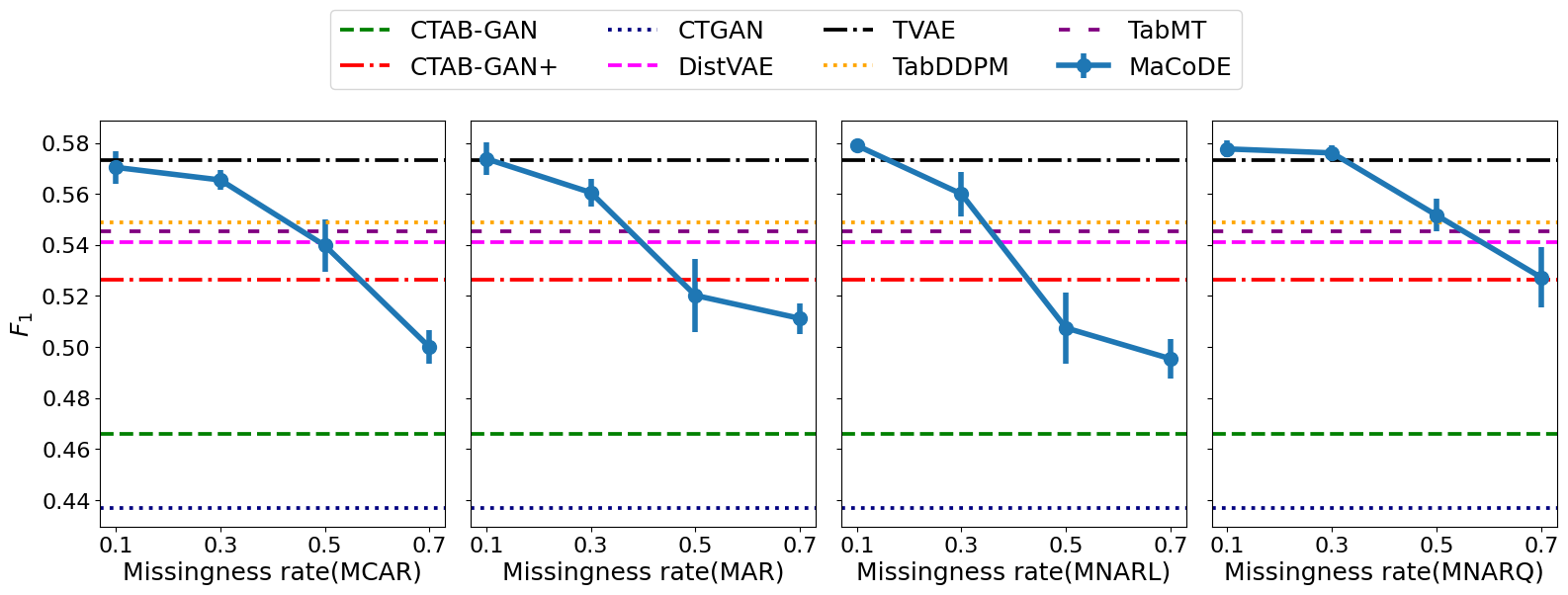}
    \caption{\textbf{Q2.} Sensitivity analysis of machine learning utility according to missingness rate. Machine learning utility is evaluated using \texttt{kings} dataset under four missing mechanisms. The means and standard errors of the mean across 10 repeated experiments are reported. Error bars represent standard errors.}
    \label{fig:sensitivity_Q2}
\end{figure}

\begin{figure}[ht]
    \centering
    \includegraphics[width=0.90\linewidth]{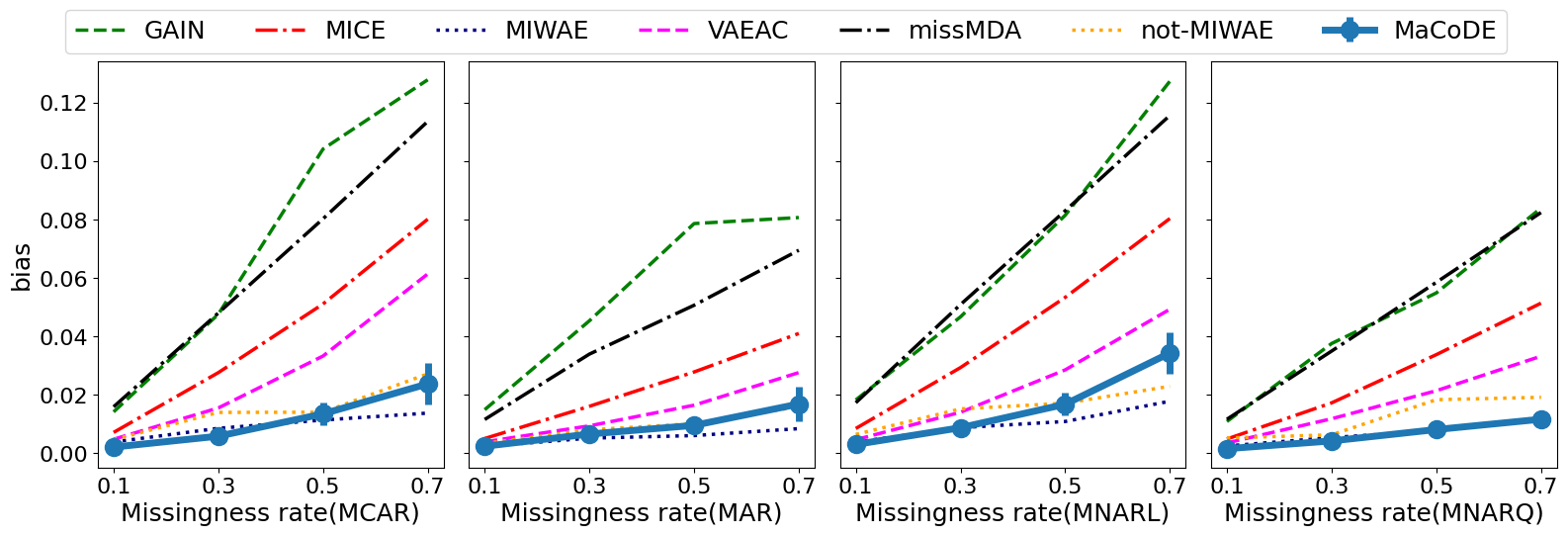}
    \includegraphics[width=0.90\linewidth]{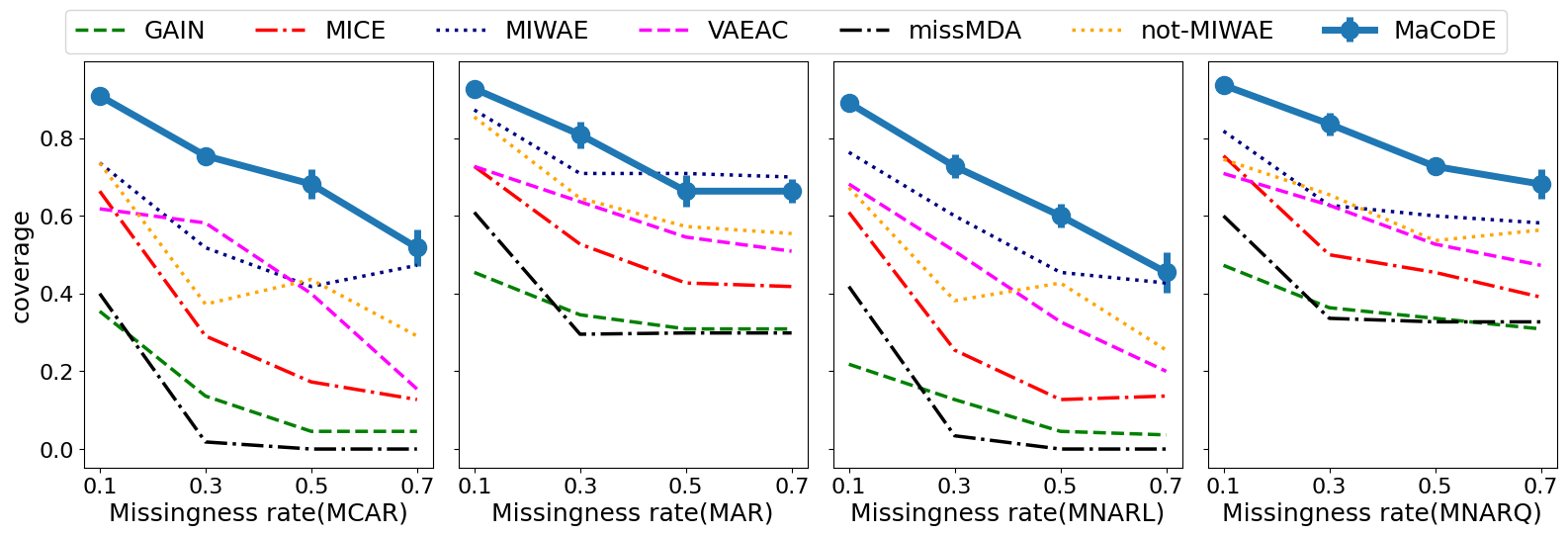}
    \includegraphics[width=0.90\linewidth]{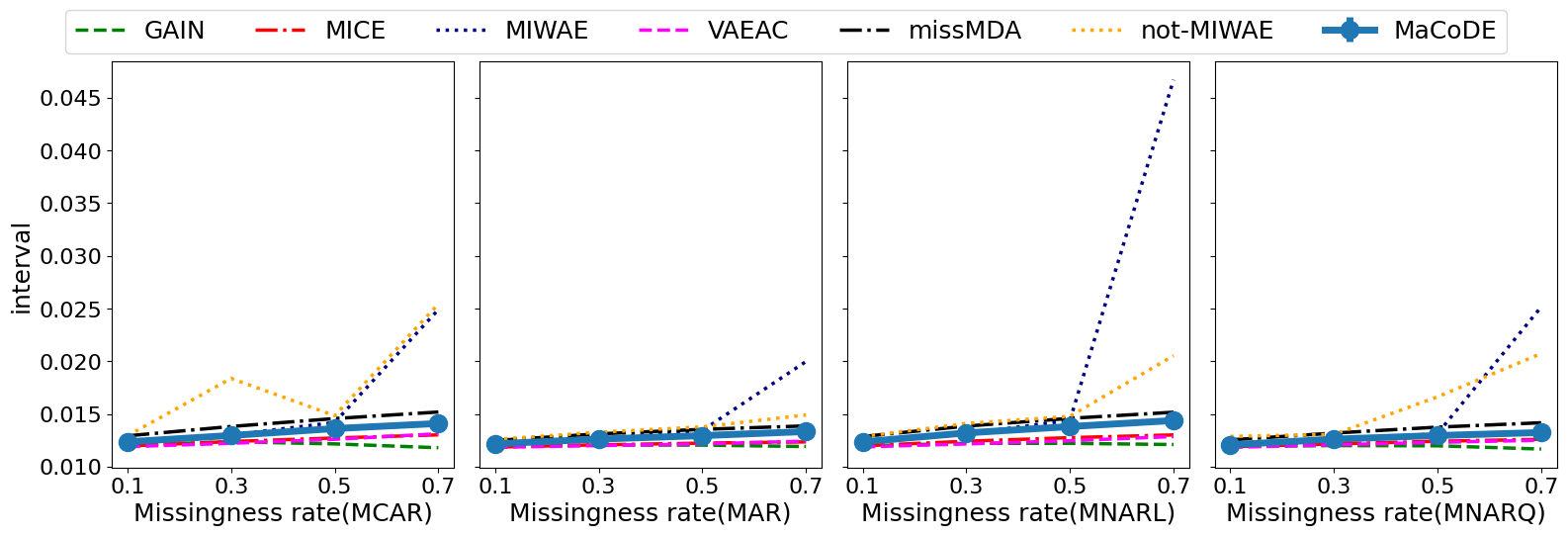}
    \caption{\textbf{Q3.} Sensitivity analysis of multiple imputations according to missingness rate. Multiple imputation is evaluated using \texttt{kings} dataset under four missing mechanisms. The means and standard errors of the mean across 10 repeated experiments are reported. Error bars represent standard errors.}
    \label{fig:sensitivity_Q3}
\end{figure}

    \item \textbf{Result:} Figure \ref{fig:sensitivity_Q2} and \ref{fig:sensitivity_Q3} show the sensitivity analysis conducted by varying the missingness rate of \texttt{kings} dataset across four missing data mechanisms (MCAR, MAR, MNARL, MNARQ). In Figure \ref{fig:sensitivity_Q2}, it's evident that MaCoDE maintains competitive performance in terms of SMAPE, even with increasing missingness rates across all missing data mechanisms. Regarding the $F_1$ score, MaCoDE outperforms other models at missingness rates of 0.1 and 0.3, but its performance diminishes beyond a missingness rate of 0.5 (except for the MNARQ missing data mechanism). Additionally, as shown in Figure \ref{fig:sensitivity_Q3}, MaCoDE consistently exhibits competitive performance in the multiple imputation, regardless of the increasing missingness rate, without significant performance degradation compared to other imputation models.

    Hence, Figures \ref{fig:sensitivity_Q2} and \ref{fig:sensitivity_Q3} demonstrate that MaCoDE maintains comparable performance to other models even when trained on a dataset with missing values (i.e., incomplete dataset) without compromising the quality of the synthetic data it generates. This is a notable advantage of our proposed model over other baseline models, which struggle with training on datasets containing missing values.
    
\eed


\clearpage
\subsection{Detailed Experimental Results}\label{app:results}
\subsubsection{Q1. Synthetic Data Quality} \textbf{}

\begin{table}[!ht]
  \centering
  \resizebox{0.65\textwidth}{!}{
  \begin{tabular}{lrrrrrrrrrrrrrrrrr}
    \toprule
    & \multicolumn{8}{c}{\texttt{abalone}}\\
    \midrule
    & \multicolumn{4}{c}{Statistical fidelity} & \multicolumn{4}{c}{Machine learning utility} \\
    \cmidrule(lr){2-5} \cmidrule(lr){6-9}
    Model & KL $\downarrow$ & GoF $\downarrow$ & MMD $\downarrow$ & WD $\downarrow$ & SMAPE $\downarrow$ & $F_1$ $\uparrow$ & Model $\uparrow$ & Feature $\uparrow$\\
    \midrule 
Baseline & $.006_{\pm .000}$ & $.023_{\pm .001}$ & $.001_{\pm .000}$ & $.152_{\pm .010}$ & $.031_{\pm .000}$ & $.229_{\pm .003}$ & $.790_{\pm .057}$ & $.981_{\pm .006}$\\
    \midrule
CTGAN& $.114_{\pm .011}$ & $.289_{\pm .058}$ & $.109_{\pm .010}$ & $1.174_{\pm .069}$ & $.101_{\pm .003}$ & $.166_{\pm .004}$ & $.625_{\pm .087}$ & $.283_{\pm .095}$ \\
TVAE& $.064_{\pm .003}$ & $.121_{\pm .004}$ & $.016_{\pm .001}$ & $.387_{\pm .007}$ & $.120_{\pm .005}$ & $.247_{\pm .003}$ & $.152_{\pm .146}$ & $.417_{\pm .105}$  \\
CTAB-GAN& $.208_{\pm .012}$ & $.403_{\pm .054}$ & $.096_{\pm .013}$ & $1.161_{\pm .120}$ & $.109_{\pm .005}$ & $.162_{\pm .005}$ & $.274_{\pm .161}$ & $.395_{\pm .158}$  \\
CTAB-GAN+& $.035_{\pm .008}$ & $.091_{\pm .020}$ & $.022_{\pm .006}$ & $.485_{\pm .094}$ & $.068_{\pm .006}$ & $.189_{\pm .013}$ & $.805_{\pm .147}$ & $.771_{\pm .100}$  \\
DistVAE& $.010_{\pm .001}$ & $.043_{\pm .003}$ & $.005_{\pm .000}$ & $.263_{\pm .007}$ & $.061_{\pm .001}$ & $.216_{\pm .002}$ & $.825_{\pm .050}$ & $.288_{\pm .034}$  \\
TabDDPM& $.119_{\pm .030}$ & $.905_{\pm .577}$ & $.005_{\pm .002}$ & $4.592_{\pm 1.055}$ & $.033_{\pm .000}$ & $.222_{\pm .002}$ & $.747_{\pm .059}$ & $.957_{\pm .008}$  \\
TabMT& $.003_{\pm .000}$ & $.022_{\pm .001}$ & $.033_{\pm .001}$ & $.949_{\pm .033}$ & $.081_{\pm .001}$ & $.198_{\pm .003}$ & $.744_{\pm .050}$ & $.852_{\pm .022}$  \\
    \midrule
MaCoDE & $.017_{\pm .002}$ & $.064_{\pm .006}$ & $.005_{\pm .001}$ & $.150_{\pm .009}$ & $.039_{\pm .000}$ & $.217_{\pm .003}$ & $.744_{\pm .049}$ & $.952_{\pm .010}$\\
    \bottomrule
  \end{tabular}
  }
  
  
  \resizebox{0.65\textwidth}{!}{
  \begin{tabular}{lrrrrrrrrrrrrrrrrr}
    \toprule
    & \multicolumn{8}{c}{\texttt{banknote}}\\
    \midrule
    & \multicolumn{4}{c}{Statistical fidelity} & \multicolumn{4}{c}{Machine learning utility} \\
    \cmidrule(lr){2-5} \cmidrule(lr){6-9}
    Model & KL $\downarrow$ & GoF $\downarrow$ & MMD $\downarrow$ & WD $\downarrow$ & SMAPE $\downarrow$ & $F_1$ $\uparrow$ & Model $\uparrow$ & Feature $\uparrow$\\
    \midrule
Baseline & $.016_{\pm .001}$ & $.040_{\pm .003}$ & $.002_{\pm .000}$ & $.059_{\pm .004}$ & $.232_{\pm .002}$ & $.957_{\pm .002}$ & $.903_{\pm .027}$ & $1.000_{\pm .000}$\\
    \midrule
CTGAN& $.146_{\pm .013}$ & $.187_{\pm .013}$ & $.095_{\pm .009}$ & $1.260_{\pm .119}$ & $.646_{\pm .009}$ & $.745_{\pm .020}$ & $-.321_{\pm .143}$ & $.720_{\pm .085}$  \\
TVAE& $.047_{\pm .009}$ & $.084_{\pm .008}$ & $.019_{\pm .004}$ & $.260_{\pm .023}$ & $.436_{\pm .006}$ & $.879_{\pm .006}$ & $.616_{\pm .079}$ & $.940_{\pm .031}$  \\
CTAB-GAN& $.030_{\pm .003}$ & $.066_{\pm .003}$ & $.021_{\pm .001}$ & $.415_{\pm .018}$ & $.519_{\pm .004}$ & $.826_{\pm .011}$ & $.207_{\pm .166}$ & $.880_{\pm .033}$  \\
CTAB-GAN+& $.176_{\pm .076}$ & $.137_{\pm .042}$ & $.072_{\pm .038}$ & $.847_{\pm .227}$ & $.670_{\pm .031}$ & $.634_{\pm .059}$ & $-.388_{\pm .274}$ & $.280_{\pm .605}$  \\
DistVAE& $.075_{\pm .002}$ & $.068_{\pm .001}$ & $.031_{\pm .001}$ & $.563_{\pm .011}$ & $.651_{\pm .005}$ & $.763_{\pm .009}$ & $-.513_{\pm .081}$ & $1.000_{\pm .000}$  \\
TabDDPM& $.052_{\pm .009}$ & $.054_{\pm .004}$ & $.006_{\pm .001}$ & $.364_{\pm .059}$ & $.315_{\pm .004}$ & $.942_{\pm .003}$ & $.839_{\pm .033}$ & $1.000_{\pm .000}$  \\
TabMT& $.012_{\pm .001}$ & $.033_{\pm .001}$ & $.012_{\pm .001}$ & $.301_{\pm .008}$ & $.495_{\pm .005}$ & $.883_{\pm .007}$ & $.468_{\pm .126}$ & $.960_{\pm .027}$  \\
    \midrule
MaCoDE & $.031_{\pm .003}$ & $.074_{\pm .005}$ & $.009_{\pm .001}$ & $.148_{\pm .011}$ & $.374_{\pm .005}$ & $.918_{\pm .005}$ & $.652_{\pm .071}$ & $1.000_{\pm .000}$\\
    \bottomrule
  \end{tabular}
  }

  \resizebox{0.65\textwidth}{!}{
  \begin{tabular}{lrrrrrrrrrrrrrrrrr}
    \toprule
    & \multicolumn{8}{c}{\texttt{breast}}\\
    \midrule
    & \multicolumn{4}{c}{Statistical fidelity} & \multicolumn{4}{c}{Machine learning utility} \\
    \cmidrule(lr){2-5} \cmidrule(lr){6-9}
    Model & KL $\downarrow$ & GoF $\downarrow$ & MMD $\downarrow$ & WD $\downarrow$ & SMAPE $\downarrow$ & $F_1$ $\uparrow$ & Model $\uparrow$ & Feature $\uparrow$\\
    \midrule
Baseline & $.081_{\pm .005}$ & $.080_{\pm .003}$ & $.007_{\pm .001}$ & $5.511_{\pm .116}$ & $.056_{\pm .001}$ & $.946_{\pm .002}$ & $.662_{\pm .106}$ & $.865_{\pm .015}$\\
    \midrule
CTGAN& $.419_{\pm .029}$ & $.345_{\pm .013}$ & $.242_{\pm .008}$ & $35.684_{\pm 1.061}$ & $.237_{\pm .006}$ & $.645_{\pm .046}$ & $.057_{\pm .167}$ & $-.163_{\pm .045}$ \\
TVAE& $.049_{\pm .002}$ & $.087_{\pm .002}$ & $.010_{\pm .001}$ & $6.519_{\pm .079}$ & $.090_{\pm .001}$ & $.938_{\pm .004}$ & $.687_{\pm .071}$ & $.895_{\pm .012}$  \\
CTAB-GAN& $.218_{\pm .010}$ & $.186_{\pm .005}$ & $.092_{\pm .005}$ & $16.515_{\pm .534}$ & $.151_{\pm .005}$ & $.877_{\pm .010}$ & $.350_{\pm .129}$ & $.558_{\pm .034}$  \\
CTAB-GAN+& $.564_{\pm .255}$ & $.356_{\pm .092}$ & $.229_{\pm .090}$ & $25.009_{\pm 9.867}$ & $.242_{\pm .046}$ & $.696_{\pm .138}$ & $.265_{\pm .525}$ & $.187_{\pm .275}$  \\
DistVAE& $.071_{\pm .002}$ & $.090_{\pm .001}$ & $.033_{\pm .001}$ & $1.407_{\pm .107}$ & $.123_{\pm .002}$ & $.889_{\pm .006}$ & $.274_{\pm .071}$ & $.750_{\pm .022}$  \\
TabDDPM& $3.110_{\pm .025}$ & $.587_{\pm .006}$ & $.306_{\pm .001}$ & $262.226_{\pm 6.849}$ & $.345_{\pm .008}$ & $.638_{\pm .016}$ & $-.260_{\pm .151}$ & $.364_{\pm .041}$  \\
TabMT& $.032_{\pm .002}$ & $.060_{\pm .002}$ & $.028_{\pm .001}$ & $12.405_{\pm .123}$ & $.128_{\pm .002}$ & $.911_{\pm .003}$ & $.503_{\pm .131}$ & $.756_{\pm .017}$  \\
    \midrule
MaCoDE & $.036_{\pm .003}$ & $.079_{\pm .006}$ & $.012_{\pm .001}$ & $8.634_{\pm .122}$ & $.098_{\pm .002}$ & $.921_{\pm .004}$ & $.332_{\pm .167}$ & $.807_{\pm .010}$\\
    \bottomrule
  \end{tabular}
  }

  \resizebox{0.65\textwidth}{!}{
  \begin{tabular}{lrrrrrrrrrrrrrrrrr}
    \toprule
    & \multicolumn{8}{c}{\texttt{concrete}}\\
    \midrule
    & \multicolumn{4}{c}{Statistical fidelity} & \multicolumn{4}{c}{Machine learning utility} \\
    \cmidrule(lr){2-5} \cmidrule(lr){6-9}
    Model & KL $\downarrow$ & GoF $\downarrow$ & MMD $\downarrow$ & WD $\downarrow$ & SMAPE $\downarrow$ & $F_1$ $\uparrow$ & Model $\uparrow$ & Feature $\uparrow$\\
    \midrule
Baseline & $.026_{\pm .002}$ & $.055_{\pm .002}$ & $.003_{\pm .000}$ & $.433_{\pm .006}$ & $.121_{\pm .002}$ & $.466_{\pm .009}$ & $.912_{\pm .017}$ & $.950_{\pm .009}$\\
    \midrule
CTGAN& $.384_{\pm .027}$ & $.830_{\pm .129}$ & $.134_{\pm .012}$ & $4.443_{\pm .289}$ & $.334_{\pm .004}$ & $.091_{\pm .020}$ & $.070_{\pm .151}$ & $.114_{\pm .138}$  \\
TVAE& $.125_{\pm .007}$ & $.219_{\pm .007}$ & $.017_{\pm .002}$ & $1.057_{\pm .031}$ & $.276_{\pm .003}$ & $.398_{\pm .010}$ & $.030_{\pm .161}$ & $.290_{\pm .045}$  \\
CTAB-GAN& $.204_{\pm .015}$ & $.226_{\pm .010}$ & $.051_{\pm .006}$ & $2.383_{\pm .098}$ & $.290_{\pm .004}$ & $.325_{\pm .011}$ & $.035_{\pm .061}$ & $-.260_{\pm .097}$  \\
CTAB-GAN+& $.263_{\pm .054}$ & $.278_{\pm .027}$ & $.066_{\pm .023}$ & $2.472_{\pm .380}$ & $.285_{\pm .012}$ & $.331_{\pm .045}$ & $-.009_{\pm .175}$ & $.017_{\pm .344}$  \\
DistVAE& $.186_{\pm .026}$ & $.159_{\pm .002}$ & $.024_{\pm .001}$ & $1.808_{\pm .011}$ & $.289_{\pm .003}$ & $.374_{\pm .007}$ & $-.160_{\pm .092}$ & $.107_{\pm .104}$  \\
TabDDPM& $.036_{\pm .001}$ & $.087_{\pm .007}$ & $.001_{\pm .000}$ & $.532_{\pm .031}$ & $.156_{\pm .004}$ & $.446_{\pm .007}$ & $.537_{\pm .049}$ & $.938_{\pm .019}$  \\
TabMT& $.031_{\pm .001}$ & $.115_{\pm .002}$ & $.019_{\pm .001}$ & $1.810_{\pm .034}$ & $.284_{\pm .003}$ & $.355_{\pm .010}$ & $.392_{\pm .060}$ & $.121_{\pm .086}$  \\
    \midrule
MaCoDE & $.089_{\pm .019}$ & $.118_{\pm .008}$ & $.006_{\pm .001}$ & $.684_{\pm .025}$ & $.229_{\pm .002}$ & $.405_{\pm .009}$ & $.439_{\pm .081}$ & $.919_{\pm .021}$\\
    \bottomrule
  \end{tabular}
  }
  
  
  \resizebox{0.65\textwidth}{!}{
  \begin{tabular}{lrrrrrrrrrrrrrrrrr}
    \toprule
    & \multicolumn{8}{c}{\texttt{covtype}}\\
    \midrule
    & \multicolumn{4}{c}{Statistical fidelity} & \multicolumn{4}{c}{Machine learning utility} \\
    \cmidrule(lr){2-5} \cmidrule(lr){6-9}
    Model & KL $\downarrow$ & GoF $\downarrow$ & MMD $\downarrow$ & WD $\downarrow$ & SMAPE $\downarrow$ & $F_1$ $\uparrow$ & Model $\uparrow$ & Feature $\uparrow$\\
    \midrule
Baseline & $.000_{\pm .000}$ & $.002_{\pm .000}$ & $.000_{\pm .000}$ & $.454_{\pm .002}$ & $.067_{\pm .000}$ & $.804_{\pm .000}$ & $1.000_{\pm .000}$ & $1.000_{\pm .000}$\\
    \midrule
CTGAN& $.105_{\pm .007}$ & $.493_{\pm .063}$ & $.017_{\pm .002}$ & $1.338_{\pm .053}$ & $.154_{\pm .001}$ & $.495_{\pm .023}$ & $.320_{\pm .098}$ & $.878_{\pm .011}$  \\
TVAE& $.048_{\pm .004}$ & $.078_{\pm .004}$ & $.013_{\pm .002}$ & $.950_{\pm .021}$ & $.151_{\pm .001}$ & $.634_{\pm .008}$ & $.420_{\pm .055}$ & $.930_{\pm .019}$  \\
CTAB-GAN& $.020_{\pm .002}$ & $.064_{\pm .005}$ & $.005_{\pm .000}$ & $.892_{\pm .019}$ & $.147_{\pm .001}$ & $.574_{\pm .005}$ & $.250_{\pm .067}$ & $.922_{\pm .010}$  \\
CTAB-GAN+& $.002_{\pm .000}$ & $.020_{\pm .003}$ & $.001_{\pm .000}$ & $.601_{\pm .007}$ & $.122_{\pm .001}$ & $.652_{\pm .003}$ & $.300_{\pm .000}$ & $.958_{\pm .006}$  \\
DistVAE& $.007_{\pm .000}$ & $.028_{\pm .001}$ & $.005_{\pm .000}$ & $1.228_{\pm .010}$ & $.182_{\pm .001}$ & $.626_{\pm .002}$ & $-.080_{\pm .033}$ & $.895_{\pm .016}$  \\
TabDDPM& $.003_{\pm .000}$ & $.011_{\pm .000}$ & $.001_{\pm .000}$ & $.556_{\pm .006}$ & $.115_{\pm .000}$ & $.665_{\pm .001}$ & $.300_{\pm .000}$ & $.964_{\pm .000}$  \\
TabMT& $.001_{\pm .000}$ & $.009_{\pm .000}$ & $.001_{\pm .000}$ & $.621_{\pm .005}$ & $.119_{\pm .000}$ & $.701_{\pm .002}$ & $.700_{\pm .000}$ & $.960_{\pm .002}$  \\
    \midrule
MaCoDE & $.001_{\pm .000}$ & $.026_{\pm .001}$ & $.001_{\pm .000}$ & $.509_{\pm .002}$ & $.098_{\pm .000}$ & $.742_{\pm .001}$ & $.900_{\pm .000}$ & $.956_{\pm .002}$\\
    \bottomrule
  \end{tabular}
  }
  \caption{\textbf{Q1}: Statistical fidelity and machine learning utility for each dataset. The means and the standard errors of the mean across 10 repeated experiments are reported. `Baseline' refers to the result obtained using half of the real training dataset. $\uparrow$ ($\downarrow$) denotes higher (lower) is better.}
\end{table}

\begin{table}[ht]
  \centering
  \resizebox{0.65\textwidth}{!}{
  \begin{tabular}{lrrrrrrrrrrrrrrrrr}
    \toprule
    & \multicolumn{8}{c}{\texttt{kings}}\\
    \midrule
    & \multicolumn{4}{c}{Statistical fidelity} & \multicolumn{4}{c}{Machine learning utility} \\
    \cmidrule(lr){2-5} \cmidrule(lr){6-9}
    Model & KL $\downarrow$ & GoF $\downarrow$ & MMD $\downarrow$ & WD $\downarrow$ & SMAPE $\downarrow$ & $F_1$ $\uparrow$ & Model $\uparrow$ & Feature $\uparrow$\\
    \midrule
Baseline & $.002_{\pm .000}$ & $.008_{\pm .000}$ & $.000_{\pm .000}$ & $.414_{\pm .014}$ & $.088_{\pm .001}$ & $.605_{\pm .003}$ & $1.000_{\pm .000}$ & $.994_{\pm .001}$\\
    \midrule
CTGAN& $.141_{\pm .009}$ & $.753_{\pm .083}$ & $.036_{\pm .004}$ & $2.702_{\pm .347}$ & $.245_{\pm .003}$ & $.437_{\pm .010}$ & $.880_{\pm .025}$ & $.881_{\pm .008}$  \\
TVAE& $.032_{\pm .001}$ & $.097_{\pm .002}$ & $.004_{\pm .001}$ & $.942_{\pm .038}$ & $.204_{\pm .000}$ & $.573_{\pm .003}$ & $.940_{\pm .016}$ & $.894_{\pm .008}$  \\
CTAB-GAN& $.056_{\pm .004}$ & $.421_{\pm .223}$ & $.017_{\pm .001}$ & $1.388_{\pm .114}$ & $.222_{\pm .002}$ & $.466_{\pm .010}$ & $.947_{\pm .016}$ & $.898_{\pm .008}$  \\
CTAB-GAN+& $.037_{\pm .030}$ & $.100_{\pm .048}$ & $.039_{\pm .037}$ & $2.107_{\pm 1.594}$ & $.181_{\pm .007}$ & $.526_{\pm .022}$ & $.937_{\pm .055}$ & $.956_{\pm .022}$  \\
DistVAE& $.159_{\pm .017}$ & $.085_{\pm .002}$ & $.005_{\pm .000}$ & $1.039_{\pm .035}$ & $.236_{\pm .001}$ & $.541_{\pm .004}$ & $.930_{\pm .030}$ & $.907_{\pm .003}$  \\
TabDDPM& $.096_{\pm .017}$ & $1.621_{\pm .684}$ & $.001_{\pm .000}$ & $27.567_{\pm 6.882}$ & $.083_{\pm .001}$ & $.549_{\pm .007}$ & $.900_{\pm .038}$ & $.978_{\pm .005}$  \\
TabMT& $.001_{\pm .000}$ & $.010_{\pm .000}$ & $.001_{\pm .000}$ & $.741_{\pm .011}$ & $.154_{\pm .001}$ & $.545_{\pm .005}$ & $.940_{\pm .016}$ & $.964_{\pm .002}$  \\
    \midrule
MaCoDE & $.011_{\pm .001}$ & $.046_{\pm .002}$ & $.009_{\pm .002}$ & $.998_{\pm .089}$ & $.179_{\pm .001}$ & $.580_{\pm .004}$ & $.920_{\pm .025}$ & $.975_{\pm .002}$\\
    \bottomrule
  \end{tabular}
  }
  
  
  \resizebox{0.65\textwidth}{!}{
  \begin{tabular}{lrrrrrrrrrrrrrrrrr}
    \toprule
    & \multicolumn{8}{c}{\texttt{letter}}\\
    \midrule
    & \multicolumn{4}{c}{Statistical fidelity} & \multicolumn{4}{c}{Machine learning utility} \\
    \cmidrule(lr){2-5} \cmidrule(lr){6-9}
    Model & KL $\downarrow$ & GoF $\downarrow$ & MMD $\downarrow$ & WD $\downarrow$ & SMAPE $\downarrow$ & $F_1$ $\uparrow$ & Model $\uparrow$ & Feature $\uparrow$\\
    \midrule
Baseline & $.001_{\pm .000}$ & $.010_{\pm .000}$ & $.000_{\pm .000}$ & $.921_{\pm .002}$ & $.059_{\pm .000}$ & $.821_{\pm .002}$ & $1.000_{\pm .000}$ & $.990_{\pm .001}$\\
    \midrule
CTGAN& $.171_{\pm .004}$ & $.294_{\pm .004}$ & $.022_{\pm .002}$ & $4.743_{\pm .182}$ & $.169_{\pm .003}$ & $.169_{\pm .010}$ & $.120_{\pm .135}$ & $.555_{\pm .055}$  \\
TVAE& $.040_{\pm .002}$ & $.064_{\pm .002}$ & $.005_{\pm .000}$ & $1.604_{\pm .013}$ & $.106_{\pm .001}$ & $.496_{\pm .011}$ & $.550_{\pm .056}$ & $.944_{\pm .005}$  \\
CTAB-GAN& $.113_{\pm .005}$ & $.121_{\pm .008}$ & $.013_{\pm .001}$ & $3.921_{\pm .059}$ & $.153_{\pm .002}$ & $.197_{\pm .010}$ & $.130_{\pm .087}$ & $.729_{\pm .030}$  \\
CTAB-GAN+& $.051_{\pm .012}$ & $.073_{\pm .015}$ & $.005_{\pm .001}$ & $2.782_{\pm .112}$ & $.123_{\pm .001}$ & $.471_{\pm .021}$ & $.440_{\pm .126}$ & $.899_{\pm .023}$  \\
DistVAE& $.021_{\pm .001}$ & $.044_{\pm .001}$ & $.004_{\pm .000}$ & $2.526_{\pm .015}$ & $.120_{\pm .000}$ & $.577_{\pm .003}$ & $.110_{\pm .046}$ & $.854_{\pm .009}$  \\
TabDDPM& $.009_{\pm .000}$ & $.031_{\pm .001}$ & $.002_{\pm .000}$ & $1.371_{\pm .023}$ & $.075_{\pm .000}$ & $.448_{\pm .005}$ & $.500_{\pm .001}$ & $.531_{\pm .008}$  \\
TabMT& $.002_{\pm .000}$ & $.010_{\pm .000}$ & $.001_{\pm .000}$ & $1.579_{\pm .009}$ & $.083_{\pm .000}$ & $.745_{\pm .003}$ & $.900_{\pm .001}$ & $.976_{\pm .003}$  \\
    \midrule
MaCoDE & $.095_{\pm .002}$ & $.144_{\pm .003}$ & $.017_{\pm .001}$ & $2.135_{\pm .035}$ & $.106_{\pm .001}$ & $.689_{\pm .003}$ & $.850_{\pm .017}$ & $.965_{\pm .003}$\\
    \bottomrule
  \end{tabular}
  }
  
  
  \resizebox{0.65\textwidth}{!}{
  \begin{tabular}{lrrrrrrrrrrrrrrrrr}
    \toprule
    & \multicolumn{8}{c}{\texttt{loan}}\\
    \midrule
    & \multicolumn{4}{c}{Statistical fidelity} & \multicolumn{4}{c}{Machine learning utility} \\
    \cmidrule(lr){2-5} \cmidrule(lr){6-9}
    Model & KL $\downarrow$ & GoF $\downarrow$ & MMD $\downarrow$ & WD $\downarrow$ & SMAPE $\downarrow$ & $F_1$ $\uparrow$ & Model $\uparrow$ & Feature $\uparrow$\\
    \midrule
Baseline & $.002_{\pm .000}$ & $.011_{\pm .000}$ & $.001_{\pm .000}$ & $.071_{\pm .001}$ & $.267_{\pm .001}$ & $.930_{\pm .001}$ & $.907_{\pm .037}$ & $.965_{\pm .008}$\\
    \midrule
CTGAN& $.079_{\pm .009}$ & $.177_{\pm .014}$ & $.056_{\pm .006}$ & $.963_{\pm .070}$ & $.368_{\pm .003}$ & $.886_{\pm .004}$ & $.272_{\pm .079}$ & $.682_{\pm .045}$  \\
TVAE& $.121_{\pm .004}$ & $.170_{\pm .003}$ & $.023_{\pm .000}$ & $.751_{\pm .028}$ & $.339_{\pm .002}$ & $.850_{\pm .007}$ & $.685_{\pm .102}$ & $.916_{\pm .017}$  \\
CTAB-GAN& $.037_{\pm .004}$ & $.087_{\pm .004}$ & $.023_{\pm .003}$ & $.502_{\pm .032}$ & $.333_{\pm .004}$ & $.904_{\pm .003}$ & $.440_{\pm .097}$ & $.802_{\pm .014}$  \\
CTAB-GAN+& $.050_{\pm .028}$ & $.084_{\pm .009}$ & $.039_{\pm .024}$ & $.505_{\pm .251}$ & $.304_{\pm .020}$ & $.892_{\pm .011}$ & $.144_{\pm .313}$ & $.810_{\pm .043}$  \\
DistVAE& $.011_{\pm .001}$ & $.058_{\pm .002}$ & $.030_{\pm .000}$ & $.673_{\pm .007}$ & $.361_{\pm .001}$ & $.903_{\pm .003}$ & $.517_{\pm .069}$ & $.781_{\pm .016}$  \\
TabDDPM& $.004_{\pm .000}$ & $.010_{\pm .000}$ & $.001_{\pm .000}$ & $.075_{\pm .007}$ & $.281_{\pm .001}$ & $.925_{\pm .001}$ & $.677_{\pm .039}$ & $.978_{\pm .006}$  \\
TabMT& $.002_{\pm .000}$ & $.013_{\pm .001}$ & $.003_{\pm .000}$ & $.192_{\pm .008}$ & $.307_{\pm .001}$ & $.913_{\pm .002}$ & $.623_{\pm .043}$ & $.891_{\pm .012}$  \\
    \midrule
MaCoDE & $.031_{\pm .004}$ & $.047_{\pm .003}$ & $.007_{\pm .001}$ & $.126_{\pm .008}$ & $.273_{\pm .001}$ & $.919_{\pm .002}$ & $.726_{\pm .033}$ & $.896_{\pm .021}$\\
    \bottomrule
  \end{tabular}
  }
  
  
  \resizebox{0.65\textwidth}{!}{
  \begin{tabular}{lrrrrrrrrrrrrrrrrr}
    \toprule
    & \multicolumn{8}{c}{\texttt{redwine}}\\
    \midrule
    & \multicolumn{4}{c}{Statistical fidelity} & \multicolumn{4}{c}{Machine learning utility} \\
    \cmidrule(lr){2-5} \cmidrule(lr){6-9}
    Model & KL $\downarrow$ & GoF $\downarrow$ & MMD $\downarrow$ & WD $\downarrow$ & SMAPE $\downarrow$ & $F_1$ $\uparrow$ & Model $\uparrow$ & Feature $\uparrow$\\
    \midrule
Baseline & $.025_{\pm .004}$ & $.042_{\pm .002}$ & $.003_{\pm .000}$ & $1.180_{\pm .027}$ & $.080_{\pm .001}$ & $.576_{\pm .005}$ & $.802_{\pm .072}$ & $.905_{\pm .014}$\\
    \midrule
CTGAN& $.463_{\pm .017}$ & $.914_{\pm .178}$ & $.164_{\pm .007}$ & $8.166_{\pm .528}$ & $.177_{\pm .005}$ & $.228_{\pm .031}$ & $.116_{\pm .144}$ & $.071_{\pm .119}$  \\
TVAE& $.057_{\pm .002}$ & $.115_{\pm .002}$ & $.012_{\pm .001}$ & $1.842_{\pm .042}$ & $.103_{\pm .001}$ & $.569_{\pm .005}$ & $.238_{\pm .117}$ & $.684_{\pm .043}$  \\
CAB-GAN& $.111_{\pm .012}$ & $.143_{\pm .008}$ & $.042_{\pm .006}$ & $3.075_{\pm .150}$ & $.139_{\pm .003}$ & $.506_{\pm .007}$ & $.010_{\pm .109}$ & $.445_{\pm .056}$  \\
CTAB-GAN+& $.118_{\pm .025}$ & $.157_{\pm .031}$ & $.044_{\pm .014}$ & $2.969_{\pm .312}$ & $.152_{\pm .012}$ & $.488_{\pm .049}$ & $-.145_{\pm .413}$ &$.510_{\pm .343}$  \\
DistVAE& $.035_{\pm .001}$ & $.064_{\pm .001}$ & $.015_{\pm .000}$ & $2.423_{\pm .018}$ & $.134_{\pm .001}$ & $.533_{\pm .005}$ & $-.108_{\pm .091}$ & $.776_{\pm .030}$  \\
TabDDPM& $2.678_{\pm .108}$ & $.652_{\pm .050}$ & $.214_{\pm .009}$ & $86.984_{\pm 4.091}$ & $.108_{\pm .002}$ & $.481_{\pm .010}$ & $.412_{\pm .122}$ & $.372_{\pm .066}$  \\
TabMT& $.023_{\pm .002}$ & $.052_{\pm .001}$ & $.015_{\pm .000}$ & $2.534_{\pm .024}$ & $.127_{\pm .002}$ & $.507_{\pm .009}$ & $.010_{\pm .112}$ & $.567_{\pm .057}$  \\
    \midrule
MaCoDE & $.022_{\pm .001}$ & $.071_{\pm .002}$ & $.005_{\pm .000}$ & $1.526_{\pm .016}$ & $.105_{\pm .001}$ & $.524_{\pm .007}$ & $.197_{\pm .092}$ & $.870_{\pm .032}$\\
    \bottomrule
  \end{tabular}
  }
  
  
  \resizebox{0.65\textwidth}{!}{
  \begin{tabular}{lrrrrrrrrrrrrrrrrr}
    \toprule
    & \multicolumn{8}{c}{\texttt{whitewine}}\\
    \midrule
    & \multicolumn{4}{c}{Statistical fidelity} & \multicolumn{4}{c}{Machine learning utility} \\
    \cmidrule(lr){2-5} \cmidrule(lr){6-9}
    Model & KL $\downarrow$ & GoF $\downarrow$ & MMD $\downarrow$ & WD $\downarrow$ & SMAPE $\downarrow$ & $F_1$ $\uparrow$ & Model $\uparrow$ & Feature $\uparrow$\\
    \midrule
Baseline & $.006_{\pm .001}$ & $.022_{\pm .001}$ & $.001_{\pm .000}$ & $.993_{\pm .008}$ & $.065_{\pm .000}$ & $.523_{\pm .004}$ & $.897_{\pm .027}$ & $.913_{\pm .012}$\\
    \midrule
CTGAN& $.186_{\pm .008}$ & $1.330_{\pm .159}$ & $.068_{\pm .005}$ & $3.881_{\pm .153}$ & $.123_{\pm .002}$ & $.248_{\pm .019}$ & $-.060_{\pm .064}$ & $.148_{\pm .083}$  \\
TVAE& $.078_{\pm .005}$ & $.152_{\pm .005}$ & $.040_{\pm .004}$ & $1.996_{\pm .064}$ & $.100_{\pm .001}$ & $.492_{\pm .003}$ & $.545_{\pm .127}$ & $.561_{\pm .066}$  \\
CTAB-GAN& $.166_{\pm .016}$ & $.241_{\pm .019}$ & $.080_{\pm .011}$ & $3.016_{\pm .147}$ & $.117_{\pm .002}$ & $.407_{\pm .010}$ & $-.010_{\pm .089}$ & $.311_{\pm .090}$  \\
CTAB-GAN+& $.064_{\pm .022}$ & $.139_{\pm .022}$ & $.029_{\pm .012}$ & $1.929_{\pm .265}$ & $.109_{\pm .004}$ & $.424_{\pm .021}$ & $-.080_{\pm .326}$ & $.620_{\pm .189}$  \\
DistVAE& $.018_{\pm .001}$ & $.059_{\pm .001}$ & $.009_{\pm .001}$ & $1.787_{\pm .017}$ & $.106_{\pm .001}$ & $.459_{\pm .003}$ & $.150_{\pm .095}$ & $.595_{\pm .026}$  \\
TabDDPM& $.675_{\pm .041}$ & $.161_{\pm .011}$ & $.022_{\pm .002}$ & $4.292_{\pm 2.951}$ & $.079_{\pm .001}$ & $.439_{\pm .008}$ & $.535_{\pm .070}$ & $.685_{\pm .056}$  \\
TabMT& $.005_{\pm .001}$ & $.031_{\pm .000}$ & $.007_{\pm .000}$ & $1.861_{\pm .010}$ & $.102_{\pm .001}$ & $.462_{\pm .004}$ & $-.001_{\pm .068}$ & $.562_{\pm .035}$  \\
    \midrule
MaCoDE & $.010_{\pm .001}$ & $.049_{\pm .002}$ & $.002_{\pm .000}$ & $1.238_{\pm .008}$ & $.087_{\pm .000}$ & $.465_{\pm .003}$ & $.240_{\pm .070}$ & $.903_{\pm .013}$\\
    \bottomrule
  \end{tabular}
  }
  \caption{\textbf{Q1}: Statistical fidelity and machine learning utility for each dataset. The means and the standard errors of the mean across 10 repeated experiments are reported. `Baseline' refers to the result obtained using half of the real training dataset. $\uparrow$ ($\downarrow$) denotes higher (lower) is better.}
\end{table}

\clearpage

\subsubsection{Q1. Visualization of Marginal Histogram} \textbf{}

\begin{figure}[!hb]
    \centering
    \subfigure[\texttt{abalone}]{
        \includegraphics[width=0.25\textwidth]{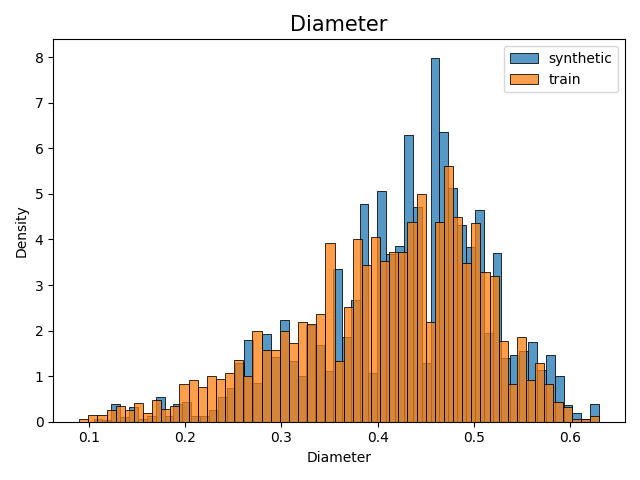}
        \includegraphics[width=0.25\textwidth]{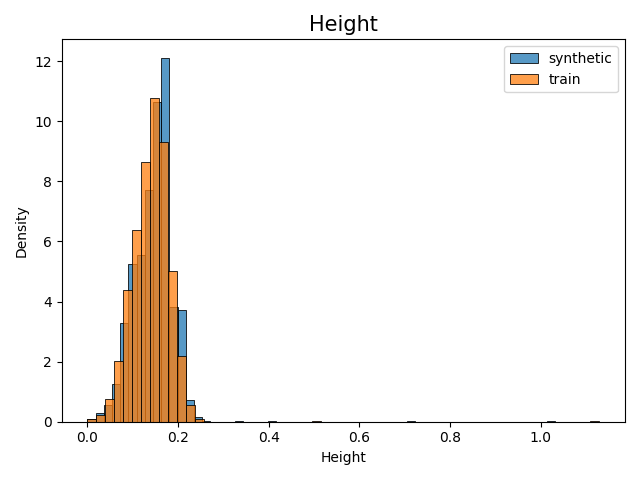}
        \includegraphics[width=0.25\textwidth]{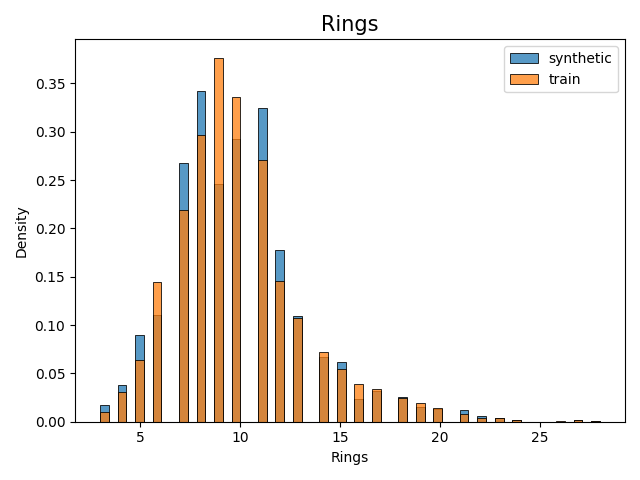}
        \includegraphics[width=0.25\textwidth]{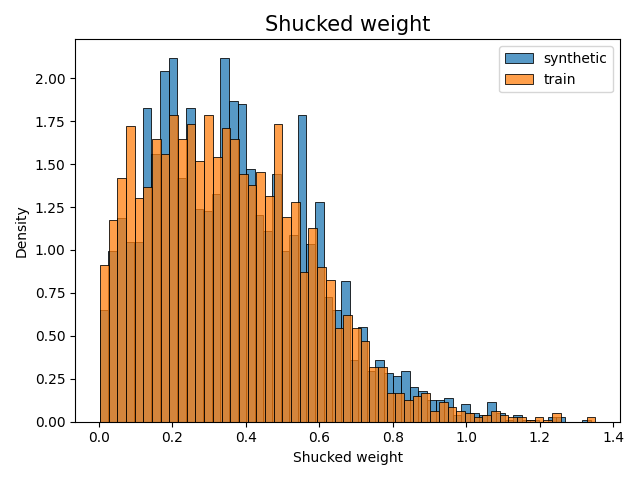}
    }
    \subfigure[\texttt{banknote}]{
        \includegraphics[width=0.25\textwidth]{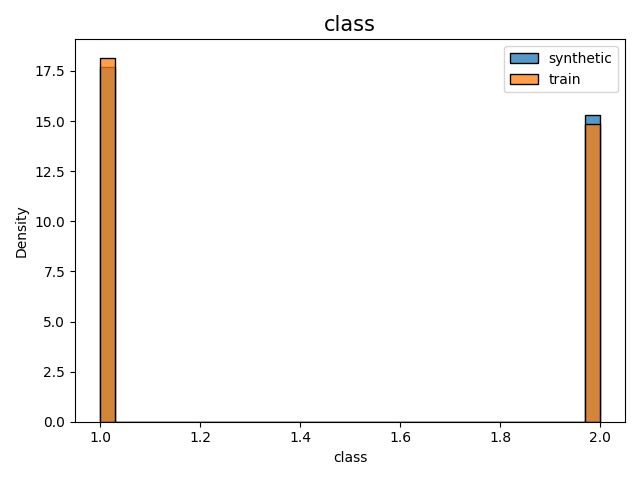}
        \includegraphics[width=0.25\textwidth]{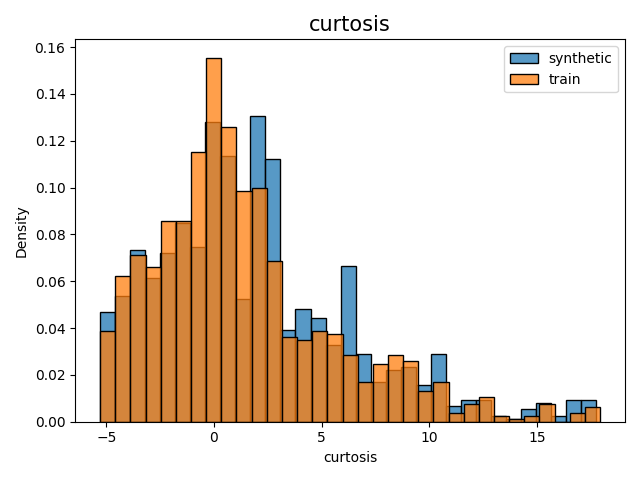}
        \includegraphics[width=0.25\textwidth]{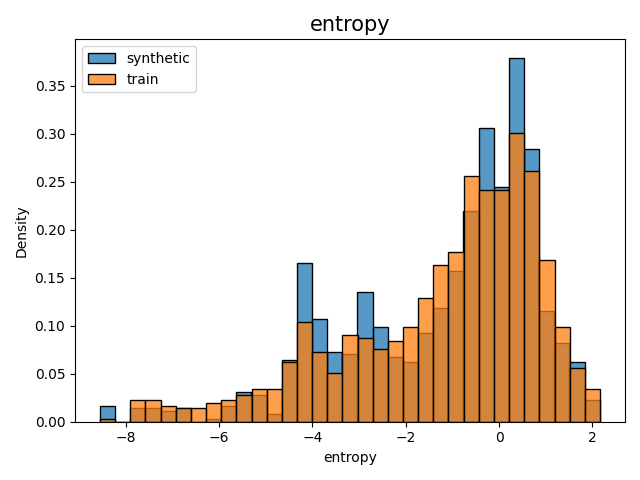}
        \includegraphics[width=0.25\textwidth]{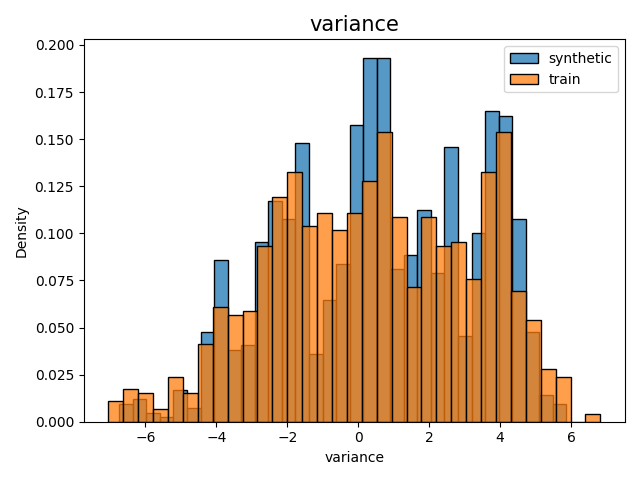}
    }
    \subfigure[\texttt{breast}]{
        \includegraphics[width=0.25\textwidth]{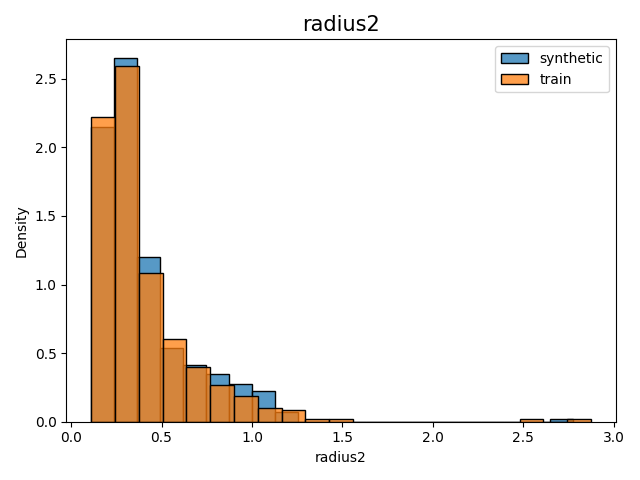}
        \includegraphics[width=0.25\textwidth]{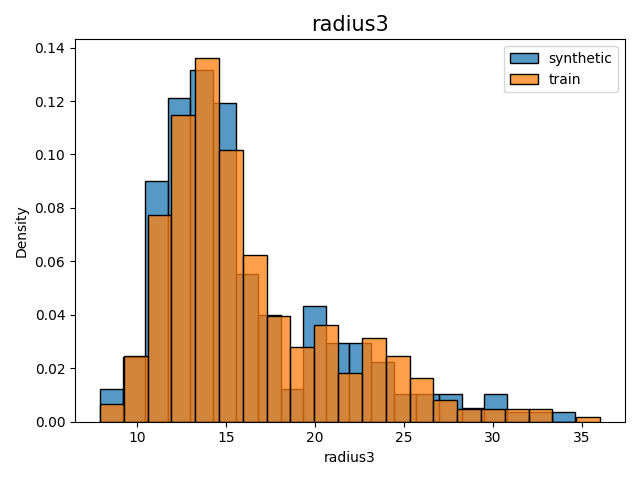}
        \includegraphics[width=0.25\textwidth]{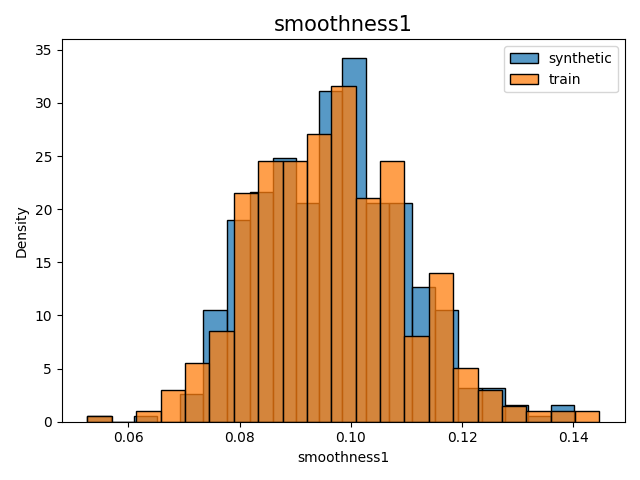}
        \includegraphics[width=0.25\textwidth]{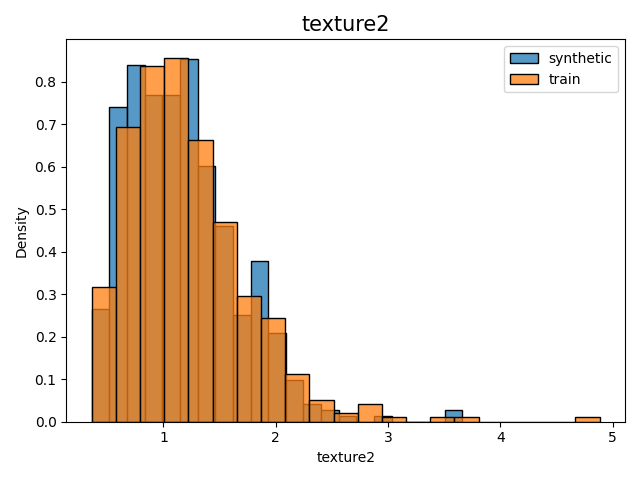}
    }
    \subfigure[\texttt{concreate}]{
        \includegraphics[width=0.25\textwidth]{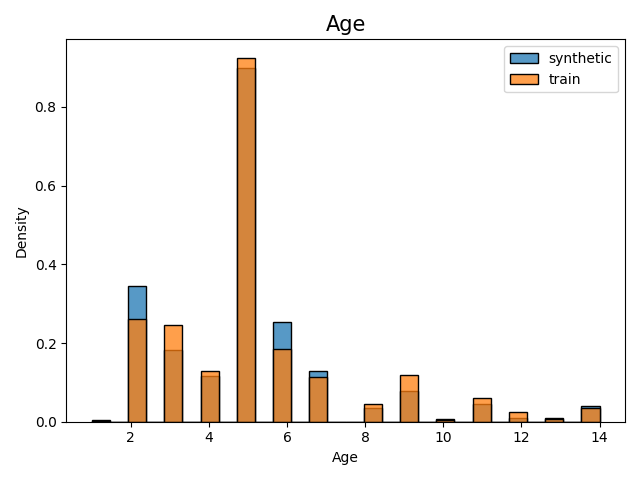}
        \includegraphics[width=0.25\textwidth]{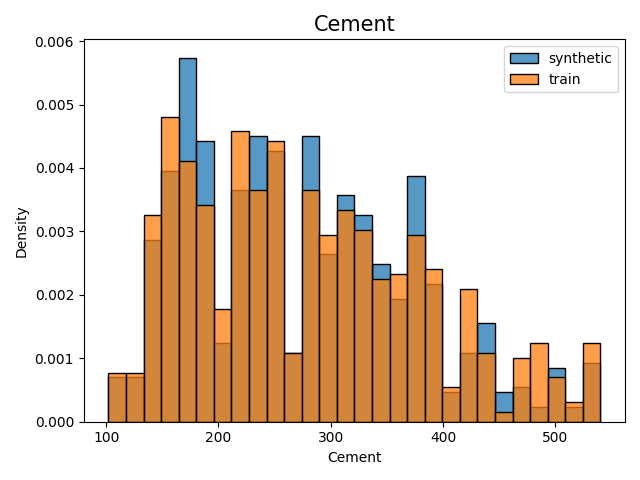}
        \includegraphics[width=0.25\textwidth]{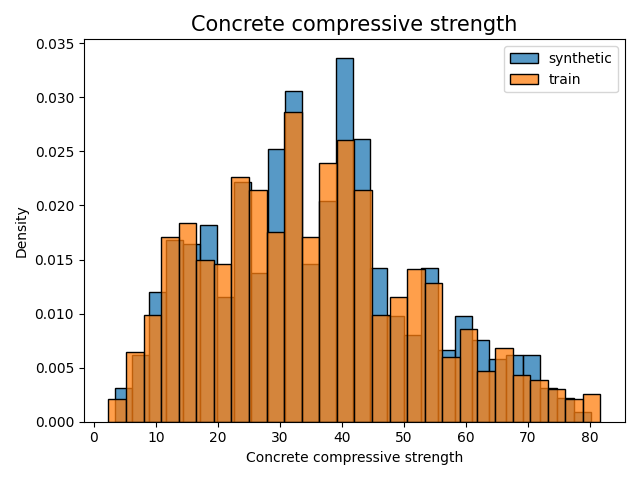}
        \includegraphics[width=0.25\textwidth]{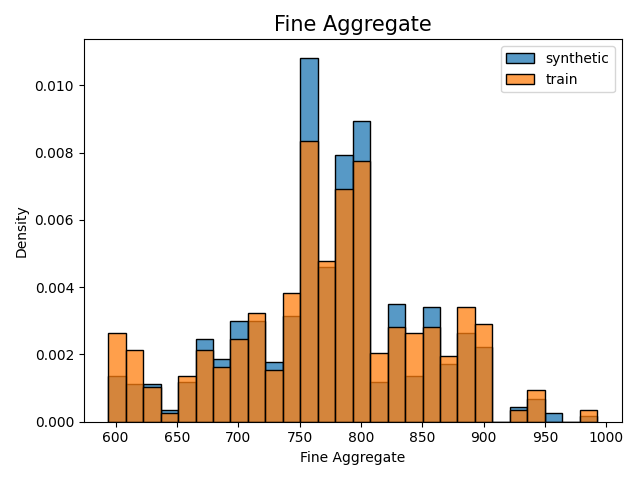}
    }
    \subfigure[\texttt{covtype}]{
        \includegraphics[width=0.25\textwidth]{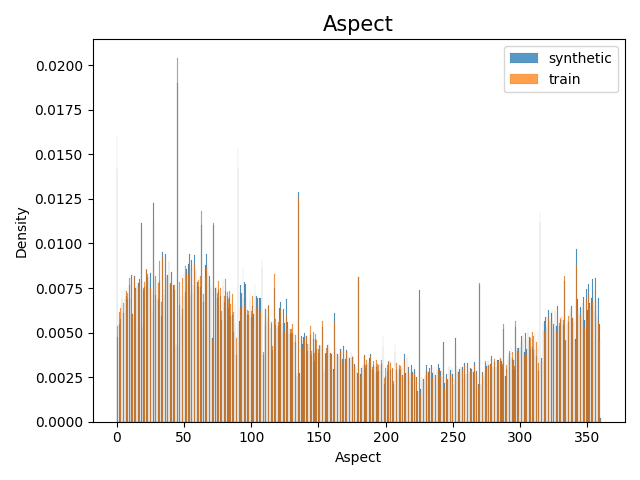}
        \includegraphics[width=0.25\textwidth]{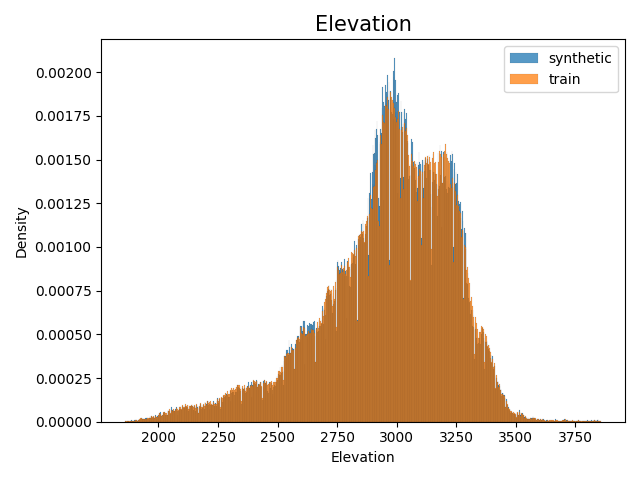}
        \includegraphics[width=0.25\textwidth]{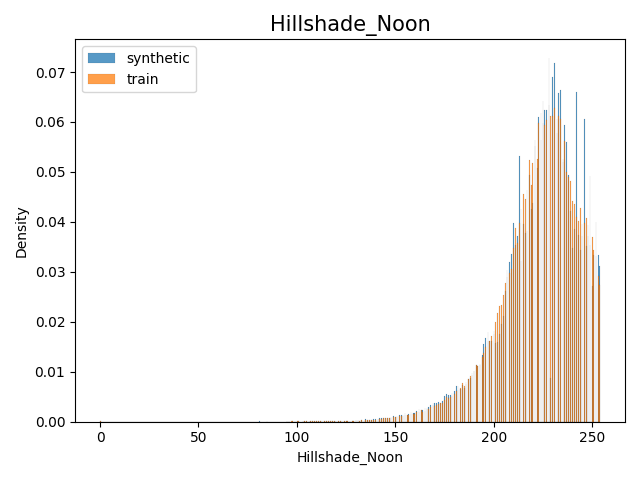}
        \includegraphics[width=0.25\textwidth]{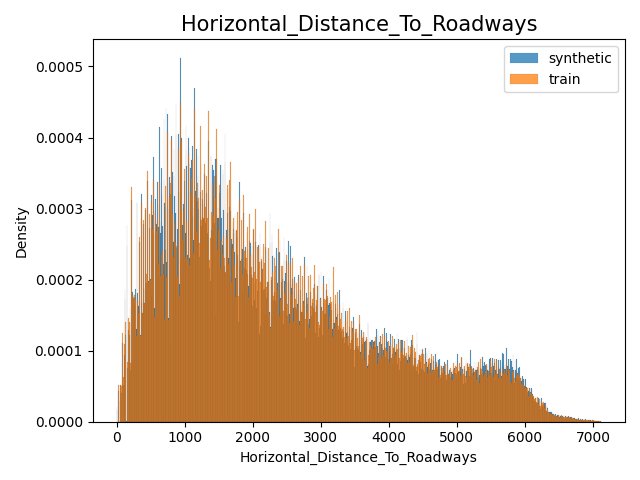}
    }
    \caption{Histograms of observed dataset and synthetic dataset, generated by MaCoDE.}
    \label{fig:marginal_hist1}
\end{figure}

\clearpage
\begin{figure}[ht]
    \centering
    \subfigure[\texttt{kings}]{
        \includegraphics[width=0.25\textwidth]{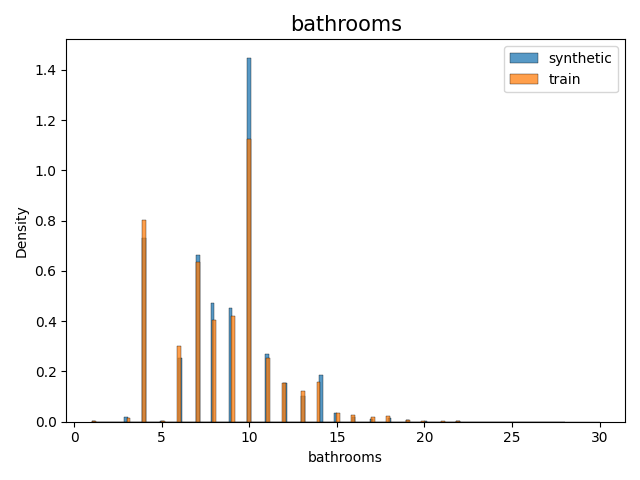}
        \includegraphics[width=0.25\textwidth]{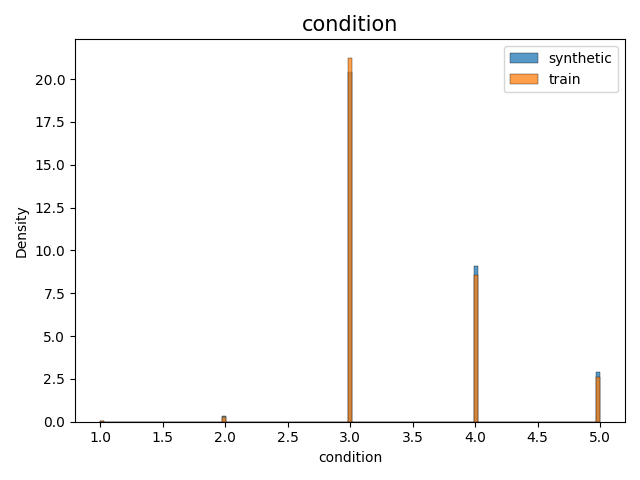}
        \includegraphics[width=0.25\textwidth]{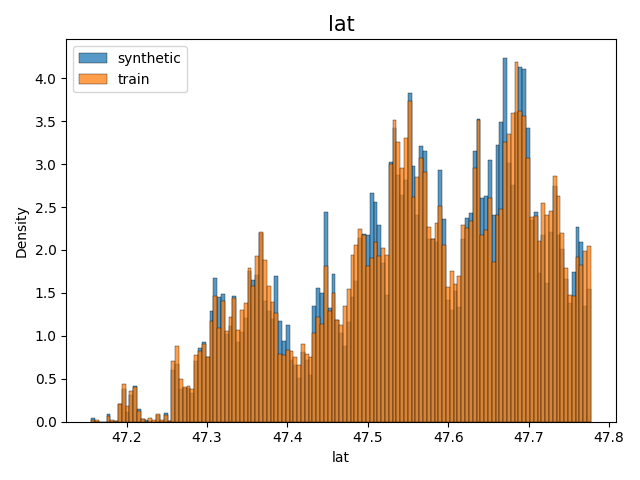}
        \includegraphics[width=0.25\textwidth]{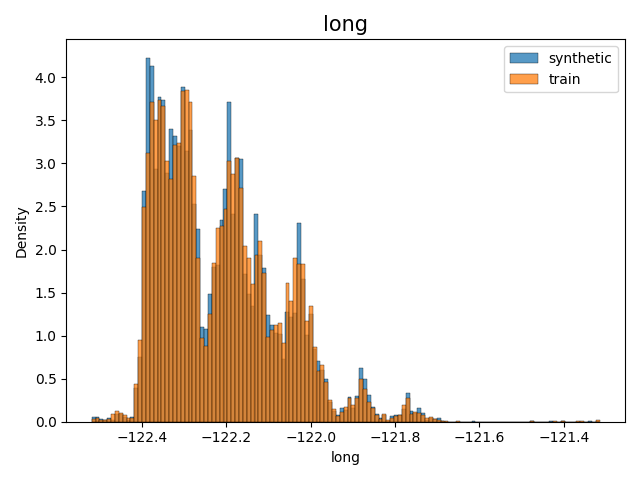}
    }
    \subfigure[\texttt{letter}]{
        \includegraphics[width=0.25\textwidth]{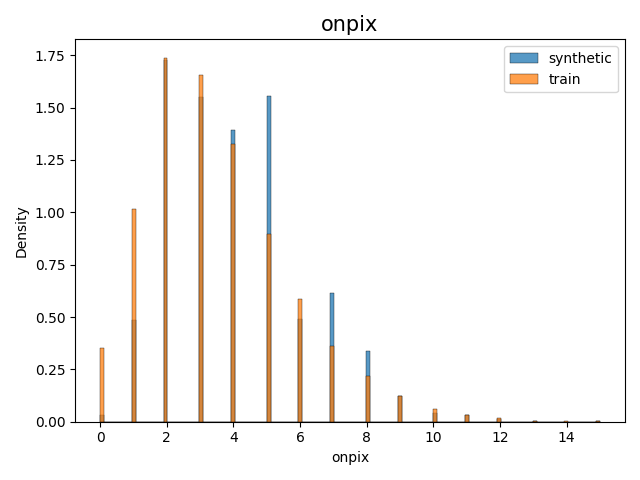}
        \includegraphics[width=0.25\textwidth]{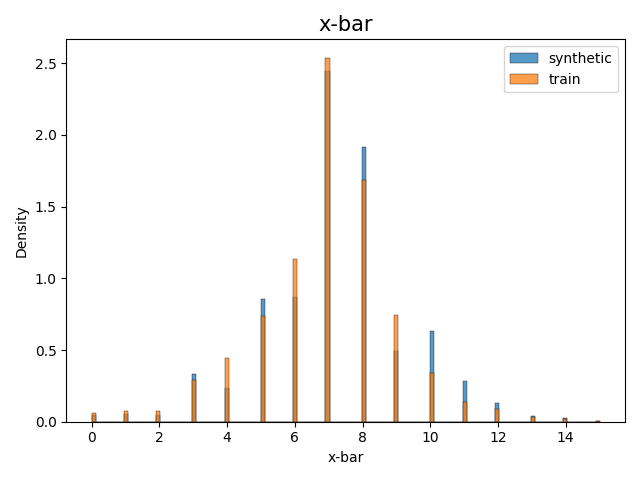}
        \includegraphics[width=0.25\textwidth]{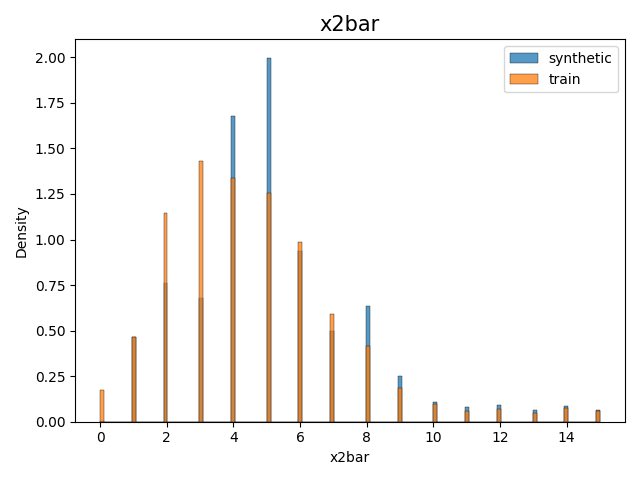}
        \includegraphics[width=0.25\textwidth]{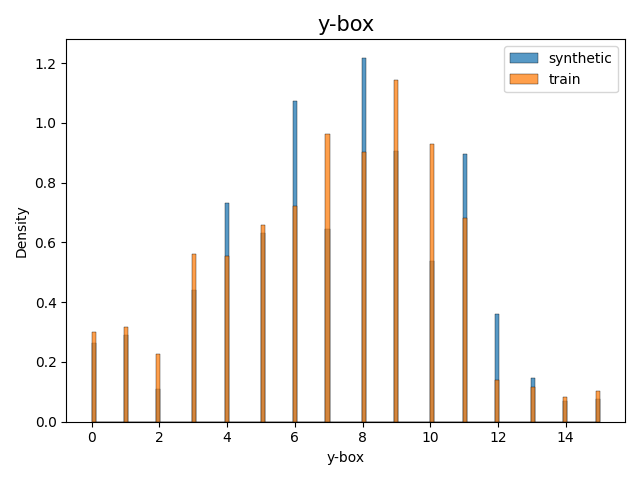}
    }
    \subfigure[\texttt{loan}]{
        \includegraphics[width=0.25\textwidth]{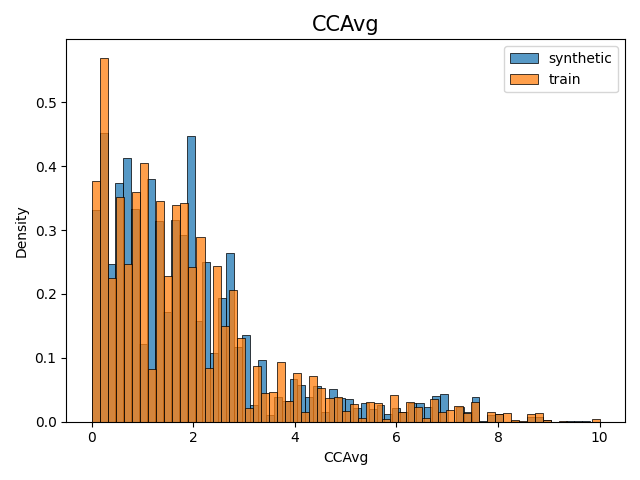}
        \includegraphics[width=0.25\textwidth]{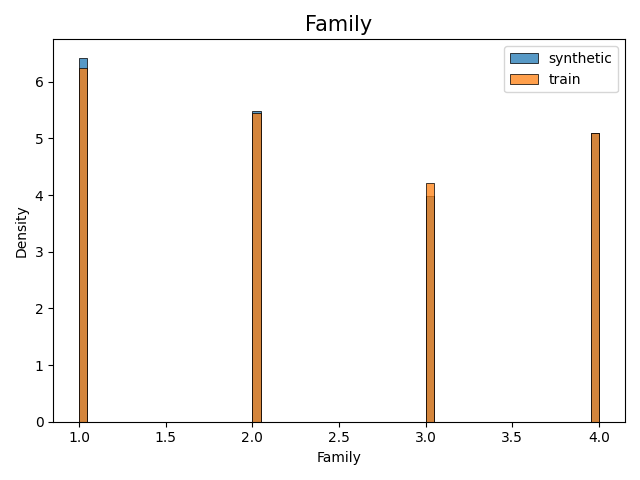}
        \includegraphics[width=0.25\textwidth]{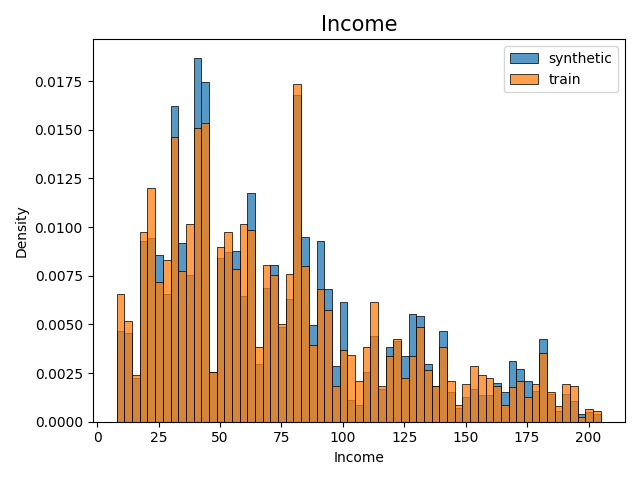}
        \includegraphics[width=0.25\textwidth]{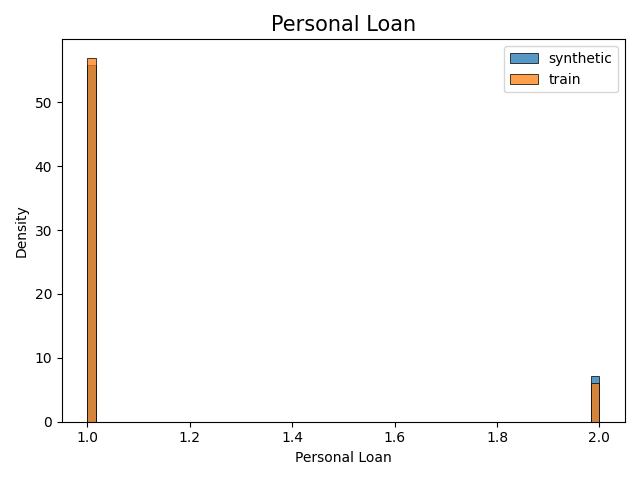}
    }
    \subfigure[\texttt{redwine}]{
        \includegraphics[width=0.25\textwidth]{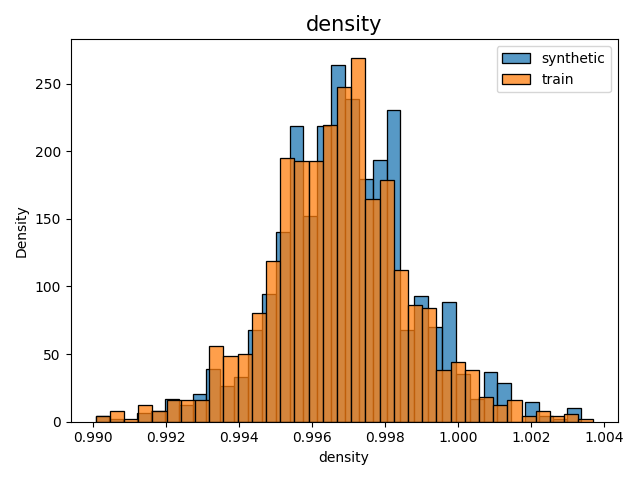}
        \includegraphics[width=0.25\textwidth]{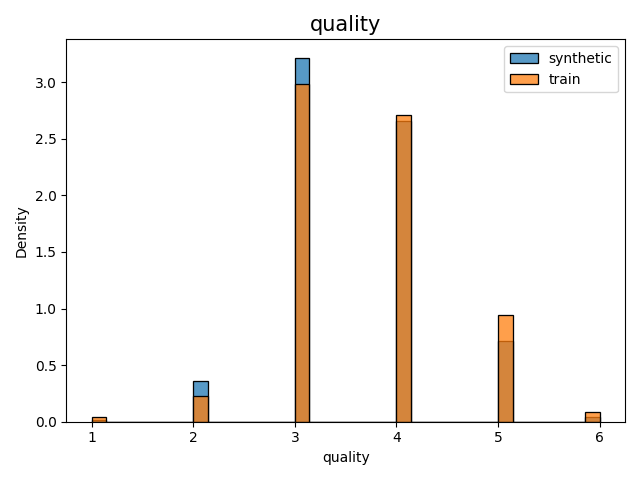}
        \includegraphics[width=0.25\textwidth]{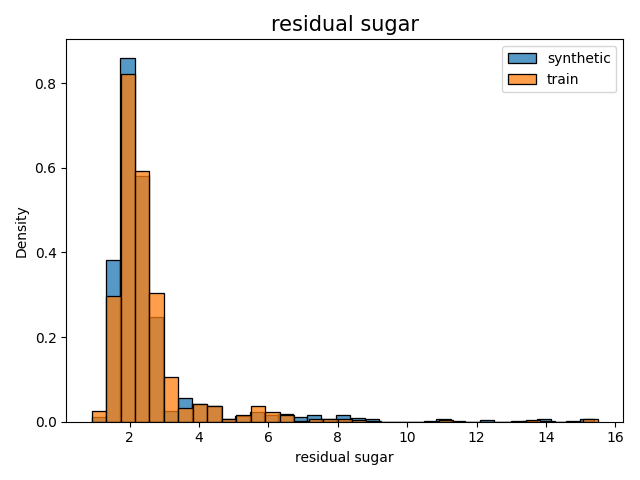}
        \includegraphics[width=0.25\textwidth]{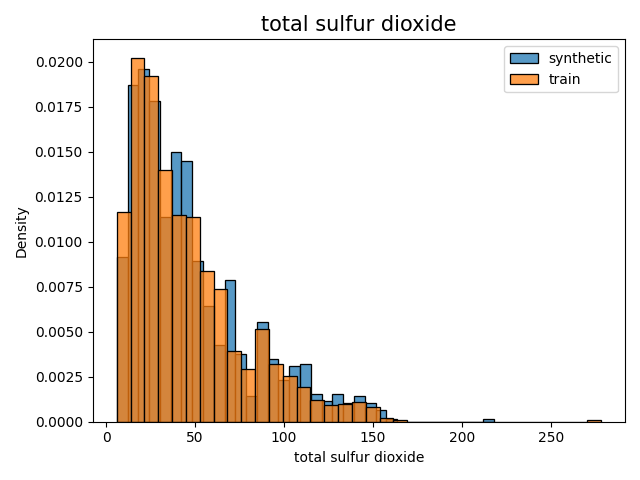}
    }
    \subfigure[\texttt{whitewine}]{
        \includegraphics[width=0.25\textwidth]{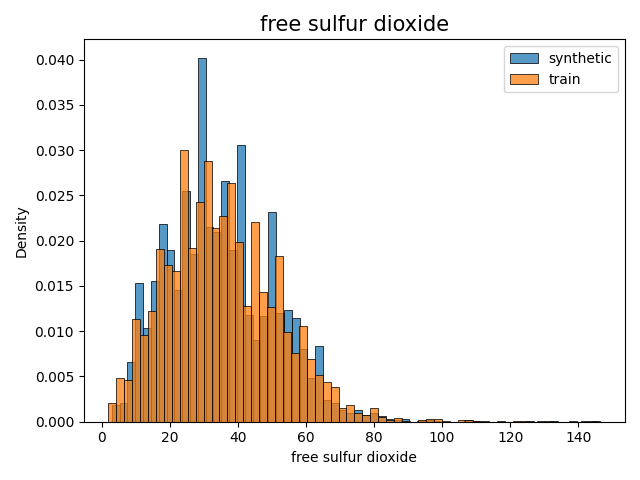}
        \includegraphics[width=0.25\textwidth]{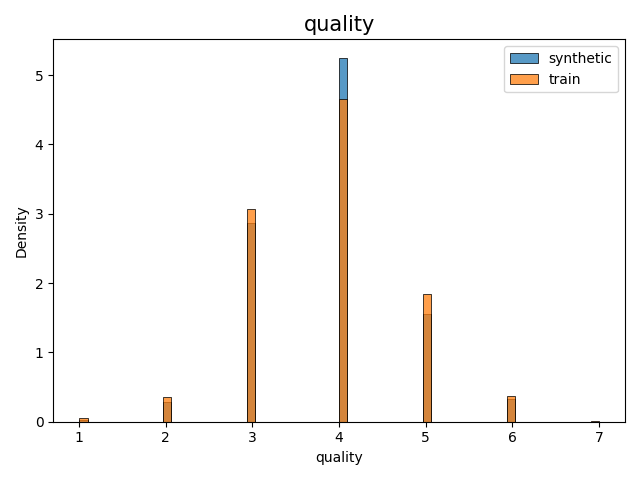}
        \includegraphics[width=0.25\textwidth]{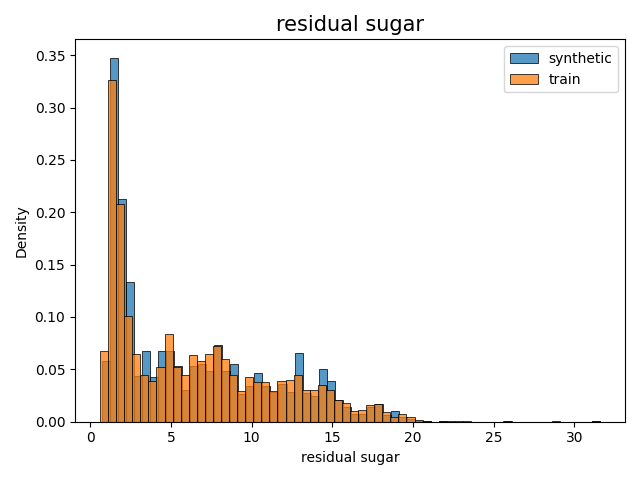}
        \includegraphics[width=0.25\textwidth]{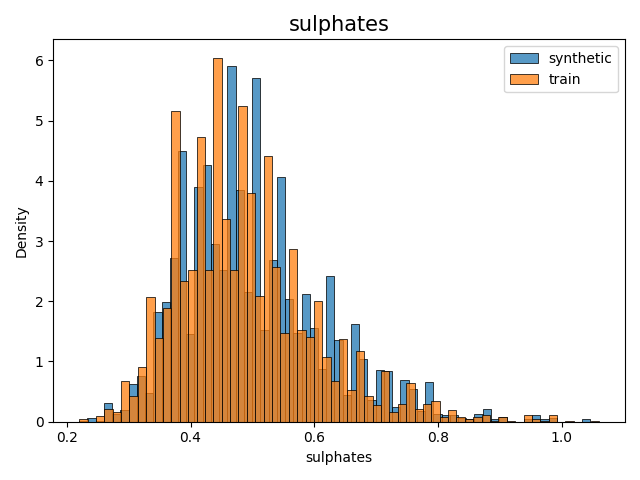}
    }
    \caption{Histograms of observed dataset and synthetic dataset, generated by MaCoDE.}
    \label{fig:marginal_hist2}
\end{figure}

\clearpage

\subsubsection{Q2: Synthetic Data Quality in Scenarios with Incomplete Training Dataset} \textbf{}

\begin{table}[!ht]
\centering
\begin{minipage}[t]{0.35\textwidth} 
  \centering
  \resizebox{0.99\linewidth}{!}{
  \begin{tabular}{lrrrrrrrrrrrrrrrr}
    \toprule
    & \multicolumn{2}{c}{MCAR}\\
    \midrule
    Dataset & SMAPE $\downarrow$ & $F_1$ $\uparrow$ \\
    \midrule
\texttt{abalone} & $0.042_{\pm 0.000}$ & $0.216_{\pm 0.003}$\\
\texttt{banknote} & $0.417_{\pm 0.004}$ & $0.879_{\pm 0.006}$\\
\texttt{breast} & $0.111_{\pm 0.001}$ & $0.909_{\pm 0.006}$\\
\texttt{concrete} & $0.240_{\pm 0.002}$ & $0.373_{\pm 0.010}$\\
\texttt{covtype} & $0.103_{\pm 0.000}$ & $0.721_{\pm 0.001}$\\
\texttt{kings} & $0.180_{\pm 0.003}$ & $0.565_{\pm 0.004}$\\
\texttt{letter} & $0.107_{\pm 0.000}$ & $0.674_{\pm 0.004}$\\
\texttt{loan} & $0.278_{\pm 0.003}$ & $0.915_{\pm 0.002}$\\
\texttt{redwine} & $0.112_{\pm 0.001}$ & $0.515_{\pm 0.009}$\\
\texttt{whitewine} & $0.091_{\pm 0.000}$ & $0.461_{\pm 0.005}$\\
    \bottomrule
  \end{tabular}
  }
  
\end{minipage}
\begin{minipage}[t]{0.35\textwidth}
    \centering
  \resizebox{0.99\linewidth}{!}{
  \begin{tabular}{lrrrrrrrrrrrrrrrr}
    \toprule
    & \multicolumn{2}{c}{MAR}\\
    \midrule
    Dataset & SMAPE $\downarrow$ & $F_1$ $\uparrow$ \\
    \midrule
\texttt{abalone} & $0.041_{\pm 0.000}$ & $0.211_{\pm 0.003}$\\
\texttt{banknote} & $0.410_{\pm 0.005}$ & $0.891_{\pm 0.005}$\\
\texttt{breast} & $0.107_{\pm 0.002}$ & $0.916_{\pm 0.005}$\\
\texttt{concrete} & $0.244_{\pm 0.003}$ & $0.383_{\pm 0.007}$\\
\texttt{covtype} & $0.102_{\pm 0.000}$ & $0.723_{\pm 0.001}$\\
\texttt{kings} & $0.183_{\pm 0.001}$ & $0.561_{\pm 0.005}$\\
\texttt{letter} & $0.107_{\pm 0.000}$ & $0.682_{\pm 0.004}$\\
\texttt{loan} & $0.278_{\pm 0.001}$ & $0.914_{\pm 0.002}$\\
\texttt{redwine} & $0.110_{\pm 0.001}$ & $0.518_{\pm 0.005}$\\
\texttt{whitewine} & $0.090_{\pm 0.001}$ & $0.465_{\pm 0.006}$\\
    \bottomrule
  \end{tabular}
  }
\end{minipage}

\vspace{3mm}

\begin{minipage}[t]{0.35\textwidth} 
    \centering
  \resizebox{0.99\linewidth}{!}{
  \begin{tabular}{lrrrrrrrrrrrrrrrr}
    \toprule
    & \multicolumn{2}{c}{MNARL}\\
    \midrule
    Dataset & SMAPE $\downarrow$ & $F_1$ $\uparrow$ \\
    \midrule
\texttt{abalone} & $0.042_{\pm 0.000}$ & $0.212_{\pm 0.004}$\\
\texttt{banknote} & $0.418_{\pm 0.007}$ & $0.883_{\pm 0.006}$\\
\texttt{breast} & $0.112_{\pm 0.002}$ & $0.908_{\pm 0.006}$\\
\texttt{concrete} & $0.243_{\pm 0.003}$ & $0.374_{\pm 0.011}$\\
\texttt{covtype} & $0.104_{\pm 0.000}$ & $0.718_{\pm 0.001}$\\
\texttt{kings} & $0.182_{\pm 0.004}$ & $0.560_{\pm 0.009}$\\
\texttt{letter} & $0.108_{\pm 0.001}$ & $0.676_{\pm 0.003}$\\
\texttt{loan} & $0.277_{\pm 0.001}$ & $0.915_{\pm 0.002}$\\
\texttt{redwine} & $0.112_{\pm 0.001}$ & $0.511_{\pm 0.009}$\\
\texttt{whitewine} & $0.091_{\pm 0.001}$ & $0.464_{\pm 0.006}$\\
    \bottomrule
  \end{tabular}
  }
\end{minipage}
\begin{minipage}[t]{0.35\textwidth}
    \centering
  \centering
  \resizebox{0.99\linewidth}{!}{
  \begin{tabular}{lrrrrrrrrrrrrrrrr}
    \toprule
    & \multicolumn{2}{c}{MNARQ}\\
    \midrule
    Dataset & SMAPE $\downarrow$ & $F_1$ $\uparrow$ \\
    \midrule
\texttt{abalone} & $0.041_{\pm 0.000}$ & $0.211_{\pm 0.003}$\\
\texttt{banknote} & $0.399_{\pm 0.003}$ & $0.899_{\pm 0.009}$\\
\texttt{breast} & $0.105_{\pm 0.001}$ & $0.918_{\pm 0.003}$\\
\texttt{concrete} & $0.239_{\pm 0.003}$ & $0.384_{\pm 0.009}$\\
\texttt{covtype} & $0.101_{\pm 0.000}$ & $0.730_{\pm 0.001}$\\
\texttt{kings} & $0.179_{\pm 0.004}$ & $0.576_{\pm 0.003}$\\
\texttt{letter} & $0.107_{\pm 0.001}$ & $0.676_{\pm 0.004}$\\
\texttt{loan} & $0.275_{\pm 0.001}$ & $0.916_{\pm 0.001}$\\
\texttt{redwine} & $0.108_{\pm 0.001}$ & $0.527_{\pm 0.006}$\\
\texttt{whitewine} & $0.089_{\pm 0.001}$ & $0.460_{\pm 0.004}$\\
    \bottomrule
  \end{tabular}
  }
  \end{minipage}
  \caption{\textbf{Q2}: Machine learning utility for each dataset under \textbf{MCAR}, \textbf{MAR}, \textbf{MNARL}, and \textbf{MNARQ} at 0.3 missingness. The means and standard errors of the mean 10 repeated experiments are reported. $\uparrow$ ($\downarrow$) denotes higher (lower) is better.}
\end{table}

\clearpage

\subsubsection{Q3: Multiple Imputation Performance} \textbf{}

\begin{table}[!ht]
\begin{minipage}[t]{0.99\textwidth}
  \centering
  \resizebox{0.55\columnwidth}{!}{
  \begin{tabular}{lrrrrrrrrrrrrrrrr}
    \toprule
    & \multicolumn{3}{c}{MCAR}\\
    \midrule
    Model & Bias $\downarrow$ & Coverage & Width $\downarrow$ \\
    \midrule
MICE & $ 0.014 _{\pm 0.001 }$ & $ 0.795 _{\pm 0.022 }$ & $ 0.040 _{\pm 0.002 }$ \\
GAIN & $ 0.024 _{\pm 0.002 }$ & $ 0.591 _{\pm 0.037 }$ & $ 0.040 _{\pm 0.002 }$ \\
missMDA & $ 0.018 _{\pm 0.001 }$ & $ 0.644 _{\pm 0.015 }$ & $ 0.045 _{\pm 0.002 }$ \\
VAEAC & $ 0.010 _{\pm 0.001 }$ & $ 0.874 _{\pm 0.022 }$ & $ 0.041 _{\pm 0.002 }$ \\
MIWAE & $ 0.008 _{\pm 0.000 }$ & $ 0.936 _{\pm 0.014 }$ & $ 0.045 _{\pm 0.002 }$ \\
notMIWAE & $ 0.008 _{\pm 0.000 }$ & $ 0.910 _{\pm 0.013 }$ & $ 0.044 _{\pm 0.002 }$ \\
EGC & $ 0.005 _{\pm 0.000 }$ & $ 1.000 _{\pm 0.000 }$ & $ 0.060 _{\pm 0.002 }$ \\
    \midrule
MaCoDE(MCAR) & $0.008_{\pm 0.000}$ & $0.950_{\pm 0.011}$ & $0.053_{\pm 0.002}$\\
    \bottomrule
  \end{tabular}
  }
\label{tab:MCAR}
\end{minipage}

\vspace{3mm}

\begin{minipage}[t]{0.99\textwidth}
  \centering
  \resizebox{0.55\columnwidth}{!}{
  \begin{tabular}{lrrrrrrrrrrrrrrrr}
    \toprule
    & \multicolumn{3}{c}{MAR}\\
    \midrule
    Model & Bias $\downarrow$ & Coverage & Width $\downarrow$ \\
    \midrule
MICE & $ 0.010 _{\pm 0.001 }$ & $ 0.845 _{\pm 0.019 }$ & $ 0.040 _{\pm 0.002 }$ \\
GAIN & $ 0.019 _{\pm 0.002 }$ & $ 0.633 _{\pm 0.033 }$ & $ 0.040 _{\pm 0.002 }$ \\
missMDA & $ 0.015 _{\pm 0.001 }$ & $ 0.700 _{\pm 0.022 }$ & $ 0.043 _{\pm 0.002 }$ \\
VAEAC & $ 0.008 _{\pm 0.001 }$ & $ 0.905 _{\pm 0.016 }$ & $ 0.040 _{\pm 0.002 }$ \\
MIWAE & $ 0.006 _{\pm 0.000 }$ & $ 0.952 _{\pm 0.012 }$ & $ 0.043 _{\pm 0.002 }$ \\
notMIWAE & $ 0.006 _{\pm 0.000 }$ & $ 0.949 _{\pm 0.012 }$ & $ 0.042 _{\pm 0.002 }$ \\
EGC & $ 0.006 _{\pm 0.000 }$ & $ 0.996 _{\pm 0.004 }$ & $ 0.058 _{\pm 0.002 }$ \\
    \midrule
MaCoDE(MAR) & $0.006_{\pm 0.000}$ & $0.963_{\pm 0.009}$ & $0.051_{\pm 0.003}$\\
    \bottomrule
  \end{tabular}
  }
\label{tab:MAR}
\end{minipage}

\vspace{3mm}

\begin{minipage}[t]{0.99\textwidth}
  \centering
  \resizebox{0.55\columnwidth}{!}{
  \begin{tabular}{lrrrrrrrrrrrrrrrr}
    \toprule
    & \multicolumn{3}{c}{MNARL}\\
    \midrule
    Model & Bias $\downarrow$ & Coverage & Width $\downarrow$ \\
    \midrule
MICE & $ 0.012_{\pm 0.001 }$ & $ 0.790 _{\pm 0.026 }$ & $ 0.037 _{\pm 0.002 }$ \\
GAIN & $ 0.026_{\pm 0.002 }$ & $ 0.538_{\pm 0.033 }$ & $ 0.040 _{\pm 0.002 }$ \\
missMDA & $ 0.020_{\pm 0.001 }$ & $ 0.592_{\pm 0.025 }$ & $ 0.045 _{\pm 0.002 }$ \\
VAEAC & $ 0.011 _{\pm 0.001 }$ & $ 0.851_{\pm 0.023 }$ & $ 0.041 _{\pm 0.002 }$ \\
MIWAE & $ 0.009_{\pm 0.001 }$ & $ 0.920_{\pm 0.015}$ & $ 0.045 _{\pm 0.002 }$ \\
notMIWAE & $ 0.009_{\pm 0.001 }$ & $ 0.908_{\pm 0.012}$ & $ 0.044 _{\pm 0.002 }$ \\

EGC & $ 0.007_{\pm 0.0 00}$ & $ 0.994_{\pm 0.004 }$ & $ 0.061 _{\pm 0.002 }$ \\

    \midrule
MaCoDE(MNARL) & $0.008_{\pm 0.000}$ & $0.948_{\pm 0.013}$ & $0.052_{\pm 0.003}$\\
    \bottomrule
  \end{tabular}
  }
\label{tab:MNARL}
\end{minipage}
\vspace{3mm}

\begin{minipage}[t]{0.99\textwidth}
  \centering
  \resizebox{0.55\columnwidth}{!}{
  \begin{tabular}{lrrrrrrrrrrrrrrrr}
    \toprule
    & \multicolumn{3}{c}{MNARQ}\\
    \midrule
    Model & Bias $\downarrow$ & Coverage & Width $\downarrow$ \\
    \midrule
MICE & $ 0.007 _{\pm 0.001 }$ & $ 0.880 _{\pm 0.018 }$ & $ 0.040 _{\pm 0.002 }$ \\
GAIN & $ 0.011 _{\pm 0.001 }$ & $ 0.819 _{\pm 0.020 }$ & $ 0.040 _{\pm 0.002 }$ \\
missMDA & $ 0.011 _{\pm 0.001 }$ & $ 0.785 _{\pm 0.016 }$ & $ 0.043 _{\pm 0.002 }$ \\
VAEAC & $ 0.006 _{\pm 0.001 }$ & $ 0.909 _{\pm 0.020 }$ & $ 0.040 _{\pm 0.002 }$ \\
MIWAE & $ 0.005 _{\pm 0.000 }$ & $ 0.969 _{\pm 0.009 }$ & $ 0.043 _{\pm 0.002 }$ \\
notMIWAE & $ 0.005 _{\pm 0.000 }$ & $ 0.958 _{\pm 0.010 }$ & $ 0.043 _{\pm 0.002 }$ \\

EGC & $ 0.004 _{\pm 0.000 }$ & $ 1.000 _{\pm 0.000 }$ & $ 0.054 _{\pm 0.002 }$ \\
    \midrule
MaCoDE(MNARQ) & $0.005_{\pm 0.000}$ & $0.944_{\pm 0.012}$ & $0.051_{\pm 0.003}$\\
    \bottomrule
  \end{tabular}
  }
\label{tab:MNARQ}
\end{minipage}
\caption{\textbf{Q3}: Multiple imputation under \textbf{MCAR}, \textbf{MAR}, \textbf{MNARL}, and \textbf{MNARQ} at 0.3 missingness. The means and standard errors of the mean across 5 datasets and 10 repeated experiments are reported. $\downarrow$ denotes the lower is better.}
\end{table}

\begin{table}[ht]
\centering
\begin{minipage}[t]{0.99\textwidth}
  \centering
  \resizebox{0.45\columnwidth}{!}{
  \begin{tabular}{lrrrrrrrrrrrrrrrr}
    \toprule
Dataset & & \texttt{abalone} &\\
    \midrule
Model & Bias $\downarrow$ & Coverage & Width $\downarrow$ \\
    \midrule	
MICE& $0.005_{\pm 0.000}$ & $1.000_{\pm 0.000}$ & $0.030_{\pm 0.000}$  \\
GAIN& $0.009_{\pm 0.001}$ & $0.829_{\pm 0.042}$ & $0.030_{\pm 0.000}$  \\
missMDA& $0.014_{\pm 0.000}$ & $0.686_{\pm 0.036}$ & $0.034_{\pm 0.001}$  \\ 
VAEAC & $0.005_{\pm 0.002}$ & $0.943_{\pm 0.074}$ & $0.030_{\pm 0.000}$\\
MIWAE& $0.006_{\pm 0.001}$ & $0.971_{\pm 0.019}$ & $0.034_{\pm 0.001}$  \\
not-MIWAE& $0.009_{\pm 0.001}$ & $0.886_{\pm 0.019}$ & $0.037_{\pm 0.001}$  \\
EGC & $0.004_{\pm 0.000}$ & $1.000_{\pm 0.000}$ & $0.040_{\pm 0.000}$  \\
    \midrule
MaCoDE & $0.007_{\pm 0.001}$ & $0.914_{\pm 0.023}$ & $0.036_{\pm 0.002}$\\
    \bottomrule
  \end{tabular}
  }
\end{minipage}

\vspace{3mm}

\begin{minipage}[t]{0.99\textwidth} 
  \centering
  \resizebox{0.45\columnwidth}{!}{
  \begin{tabular}{lrrrrrrrrrrrrrrrr}
    \toprule
Dataset & & \texttt{banknote} &\\
    \midrule
Model & Bias $\downarrow$ & Coverage & Width $\downarrow$ \\
    \midrule	
MICE& $0.018_{\pm 0.001}$ & $0.725_{\pm 0.025}$ & $0.053_{\pm 0.000}$  \\ 
GAIN& $0.009_{\pm 0.001}$ & $0.829_{\pm 0.042}$ & $0.030_{\pm 0.000}$  \\
missMDA& $0.018_{\pm 0.000}$ & $0.625_{\pm 0.042}$ & $0.056_{\pm 0.000}$  \\
VAEAC & $0.009_{\pm 0.002}$ & $1.000_{\pm 0.000}$ & $0.054_{\pm 0.000}$\\
MIWAE& $0.006_{\pm 0.000}$ & $1.000_{\pm 0.000}$ & $0.055_{\pm 0.000}$  \\
not-MIWAE& $0.006_{\pm 0.000}$ & $1.000_{\pm 0.000}$ & $0.055_{\pm 0.000}$  \\ 
EGC & $0.004_{\pm 0.001}$ & $1.000_{\pm 0.000}$ & $0.066_{\pm 0.001}$  \\
    \midrule
MaCoDE & $0.006_{\pm 0.001}$ & $1.000_{\pm 0.000}$ & $0.054_{\pm 0.000}$\\
    \bottomrule
    \end{tabular}
    }
\end{minipage}

\vspace{3mm}

\begin{minipage}[t]{0.99\textwidth}
  \centering
  \resizebox{0.45\columnwidth}{!}{
  \begin{tabular}{lrrrrrrrrrrrrrrrr}
    \toprule
Dataset & & \texttt{breast} &\\
    \midrule
Model & Bias $\downarrow$ & Coverage & Width $\downarrow$ \\
    \midrule	
MICE& $0.007_{\pm 0.000}$ & $1.000_{\pm 0.000}$ & $0.080_{\pm 0.000}$  \\
GAIN& $0.012_{\pm 0.000}$ & $0.990_{\pm 0.005}$ & $0.080_{\pm 0.000}$  \\
missMDA& $0.025_{\pm 0.000}$ & $0.883_{\pm 0.013}$ & $0.086_{\pm 0.000}$  \\ 
VAEAC& $0.008_{\pm 0.000}$ & $1.000_{\pm 0.000}$ & $0.082_{\pm 0.000}$  \\
MIWAE& $0.005_{\pm 0.000}$ & $1.000_{\pm 0.000}$ & $0.084_{\pm 0.000}$  \\
not-MIWAE& $0.006_{\pm 0.000}$ & $1.000_{\pm 0.000}$ & $0.082_{\pm 0.000}$  \\
EGC& $0.005_{\pm 0.000}$ & $1.000_{\pm 0.000}$ & $0.084_{\pm 0.000}$  \\
    \midrule
MaCoDE & $0.009_{\pm 0.000}$ & $1.000_{\pm 0.000}$ & $0.083_{\pm 0.000}$\\
    \bottomrule
  \end{tabular}
  }
\end{minipage}

\vspace{3mm}

\begin{minipage}[t]{0.99\textwidth}
  \centering
  \resizebox{0.45\columnwidth}{!}{
  \begin{tabular}{lrrrrrrrrrrrrrrrr}
    \toprule
Dataset & & \texttt{redwine} &\\
    \midrule
Model & Bias $\downarrow$ & Coverage & Width $\downarrow$ \\
    \midrule	
MICE& $0.020_{\pm 0.000}$ & $0.764_{\pm 0.020}$ & $0.048_{\pm 0.000}$  \\
GAIN& $0.024_{\pm 0.001}$ & $0.627_{\pm 0.050}$ & $0.048_{\pm 0.000}$  \\
missMDA& $0.026_{\pm 0.000}$ & $0.618_{\pm 0.012}$ & $0.055_{\pm 0.000}$  \\
VAEAC& $0.014_{\pm 0.001}$ & $0.855_{\pm 0.028}$ & $0.050_{\pm 0.001}$  \\
MIWAE& $0.009_{\pm 0.001}$ & $0.945_{\pm 0.020}$ & $0.054_{\pm 0.000}$  \\
not-MIWAE& $0.010_{\pm 0.000}$ & $0.909_{\pm 0.000}$ & $0.053_{\pm 0.000}$  \\
EGC& $0.007_{\pm 0.000}$ & $1.000_{\pm 0.000}$ & $0.070_{\pm 0.000}$  \\
    \midrule
MaCoDE & $0.008_{\pm 0.001}$ & $0.973_{\pm 0.014}$ & $0.056_{\pm 0.000}$\\
    \bottomrule
  \end{tabular}
  }
\end{minipage}

\vspace{3mm}
\begin{minipage}[t]{0.99\textwidth} 
  \centering
  \resizebox{0.45\columnwidth}{!}{
  \begin{tabular}{lrrrrrrrrrrrrrrrr}
    \toprule
Dataset & & \texttt{whitewine} &\\
    \midrule
Model & Bias $\downarrow$ & Coverage & Width $\downarrow$ \\
    \midrule	
MICE& $0.011_{\pm 0.000}$ & $0.691_{\pm 0.031}$ & $0.028_{\pm 0.000}$  \\
GAIN& $0.026_{\pm 0.003}$ & $0.482_{\pm 0.036}$ & $0.028_{\pm 0.000}$  \\
missMDA& $0.015_{\pm 0.000}$ & $0.645_{\pm 0.016}$ & $0.035_{\pm 0.000}$  \\ 
VAEAC& $0.010_{\pm 0.001}$ & $0.700_{\pm 0.038}$ & $0.028_{\pm 0.000}$  \\
MIWAE& $0.011_{\pm 0.001}$ & $0.827_{\pm 0.029}$ & $0.036_{\pm 0.001}$  \\
not-MIWAE& $0.009_{\pm 0.000}$ & $0.845_{\pm 0.033}$ & $0.033_{\pm 0.001}$  \\
EGC& $0.006_{\pm 0.000}$ & $1.000_{\pm 0.000}$ & $0.062_{\pm 0.000}$  \\
    \midrule
MaCoDE & $0.009_{\pm 0.001}$ & $0.864_{\pm 0.034}$ & $0.037_{\pm 0.001}$\\
    \bottomrule
    \end{tabular}
    }
\end{minipage}
\caption{\textbf{Q3}: Multiple imputation for each dataset under \textbf{MCAR} at 0.3 missingness. The means and standard errors of the mean across 5 datasets and 10 repeated experiments are reported. $\downarrow$ denotes lower is better.}
\label{tab:MCAR_bydata}
\end{table}

\begin{table}[ht]
\centering
\begin{minipage}[t]{0.99\textwidth}
  \centering
  \resizebox{0.45\columnwidth}{!}{
  \begin{tabular}{lrrrrrrrrrrrrrrrr}
    \toprule
Dataset & & \texttt{abalone} &\\
    \midrule
Model & Bias $\downarrow$ & Coverage & Width $\downarrow$ \\
    \midrule	
MICE& $0.005_{\pm 0.001}$ & $0.971_{\pm 0.019}$ & $0.030_{\pm 0.000}$  \\
GAIN& $0.011_{\pm 0.002}$ & $0.729_{\pm 0.072}$ & $0.030_{\pm 0.000}$  \\
missMDA& $0.015_{\pm 0.001}$ & $0.571_{\pm 0.048}$ & $0.032_{\pm 0.001}$  \\
VAEAC& $0.003_{\pm 0.001}$ & $0.986_{\pm 0.014}$ & $0.031_{\pm 0.000}$  \\
MIWAE& $0.005_{\pm 0.001}$ & $0.971_{\pm 0.019}$ & $0.033_{\pm 0.001}$  \\
not-MIWAE& $0.006_{\pm 0.001}$ & $0.943_{\pm 0.023}$ & $0.032_{\pm 0.001}$  \\
EGC& $0.004_{\pm 0.001}$ & $0.986_{\pm 0.014}$ & $0.035_{\pm 0.001}$  \\
    \midrule
MaCoDE & $0.005_{\pm 0.001}$ & $0.957_{\pm 0.022}$ & $0.033_{\pm 0.001}$\\
    \bottomrule
  \end{tabular}
  }
\end{minipage}

\vspace{3mm}
\begin{minipage}[t]{0.99\textwidth} 
  \centering
  \resizebox{0.45\columnwidth}{!}{
  \begin{tabular}{lrrrrrrrrrrrrrrrr}
    \toprule
Dataset & & \texttt{banknote} &\\
    \midrule
Model & Bias $\downarrow$ & Coverage & Width $\downarrow$ \\
    \midrule	
MICE& $0.012_{\pm 0.002}$ & $0.800_{\pm 0.033}$ & $0.052_{\pm 0.000}$  \\
GAIN& $0.021_{\pm 0.004}$ & $0.675_{\pm 0.084}$ & $0.052_{\pm 0.000}$  \\ 
missMDA& $0.014_{\pm 0.001}$ & $0.800_{\pm 0.033}$ & $0.055_{\pm 0.000}$  \\ 
VAEAC& $0.009_{\pm 0.001}$ & $0.925_{\pm 0.038}$ & $0.053_{\pm 0.000}$  \\ 
MIWAE& $0.004_{\pm 0.001}$ & $1.000_{\pm 0.000}$ & $0.054_{\pm 0.000}$  \\
not-MIWAE& $0.004_{\pm 0.001}$ & $1.000_{\pm 0.000}$ & $0.054_{\pm 0.000}$  \\
EGC& $0.005_{\pm 0.001}$ & $1.000_{\pm 0.000}$ & $0.071_{\pm 0.003}$  \\
    \midrule
MaCoDE & $0.004_{\pm 0.001}$ & $1.000_{\pm 0.000}$ & $0.054_{\pm 0.000}$\\
    \bottomrule
    \end{tabular}
    }
\end{minipage}

\vspace{3mm}

\begin{minipage}[t]{0.99\textwidth}
  \centering
  \resizebox{0.45\columnwidth}{!}{
  \begin{tabular}{lrrrrrrrrrrrrrrrr}
    \toprule
Dataset & & \texttt{breast} &\\
    \midrule
Model & Bias $\downarrow$ & Coverage & Width $\downarrow$ \\
    \midrule	
MICE& $0.005_{\pm 0.000}$ & $1.000_{\pm 0.000}$ & $0.080_{\pm 0.000}$  \\
GAIN& $0.011_{\pm 0.001}$ & $0.960_{\pm 0.016}$ & $0.080_{\pm 0.000}$  \\
missMDA& $0.019_{\pm 0.001}$ & $0.887_{\pm 0.015}$ & $0.084_{\pm 0.000}$  \\ 
VAEAC& $0.006_{\pm 0.000}$ & $0.997_{\pm 0.002}$ & $0.081_{\pm 0.000}$  \\
MIWAE& $0.004_{\pm 0.000}$ & $1.000_{\pm 0.000}$ & $0.082_{\pm 0.000}$  \\
not-MIWAE& $0.005_{\pm 0.001}$ & $0.993_{\pm 0.007}$ & $0.081_{\pm 0.000}$  \\
EGC& $0.006_{\pm 0.000}$ & $1.000_{\pm 0.000}$ & $0.083_{\pm 0.000}$  \\
    \midrule
MaCoDE & $0.007_{\pm 0.000}$ & $0.997_{\pm 0.003}$ & $0.083_{\pm 0.000}$\\
    \bottomrule
  \end{tabular}
  }
\end{minipage}

\vspace{3mm}

\begin{minipage}[t]{0.99\textwidth}
  \centering
  \resizebox{0.45\columnwidth}{!}{
  \begin{tabular}{lrrrrrrrrrrrrrrrr}
    \toprule
Dataset & & \texttt{redwine} &\\
    \midrule
Model & Bias $\downarrow$ & Coverage & Width $\downarrow$ \\
    \midrule	
MICE& $0.015_{\pm 0.001}$ & $0.836_{\pm 0.035}$ & $0.048_{\pm 0.000}$  \\
GAIN& $0.020_{\pm 0.001}$ & $0.627_{\pm 0.021}$ & $0.048_{\pm 0.000}$  \\
missMDA& $0.020_{\pm 0.002}$ & $0.709_{\pm 0.035}$ & $0.053_{\pm 0.000}$  \\
VAEAC& $0.011_{\pm 0.001}$ & $0.882_{\pm 0.019}$ & $0.049_{\pm 0.000}$  \\
MIWAE& $0.008_{\pm 0.001}$ & $0.945_{\pm 0.024}$ & $0.053_{\pm 0.001}$  \\
not-MIWAE& $0.009_{\pm 0.001}$ & $0.909_{\pm 0.030}$ & $0.051_{\pm 0.000}$  \\
EGC& $0.007_{\pm 0.001}$ & $1.000_{\pm 0.000}$ & $0.067_{\pm 0.001}$  \\
    \midrule
MaCoDE & $0.008_{\pm 0.000}$ & $0.980_{\pm 0.013}$ & $0.054_{\pm 0.001}$\\
    \bottomrule
  \end{tabular}
  }
\end{minipage}

\vspace{3mm}
\begin{minipage}[t]{0.99\textwidth} 
  \centering
  \resizebox{0.45\columnwidth}{!}{
  \begin{tabular}{lrrrrrrrrrrrrrrrr}
    \toprule
Dataset & & \texttt{whitewine} &\\
    \midrule
Model & Bias $\downarrow$ & Coverage & Width $\downarrow$ \\
    \midrule	
MICE& $0.008_{\pm 0.001}$ & $0.773_{\pm 0.034}$ & $0.028_{\pm 0.000}$  \\
GAIN& $0.025_{\pm 0.003}$ & $0.500_{\pm 0.053}$ & $0.028_{\pm 0.000}$  \\
missMDA& $0.011_{\pm 0.000}$ & $0.718_{\pm 0.032}$ & $0.033_{\pm 0.001}$  \\
VAEAC& $0.008_{\pm 0.001}$ & $0.827_{\pm 0.029}$ & $0.029_{\pm 0.000}$  \\
MIWAE& $0.007_{\pm 0.001}$ & $0.891_{\pm 0.030}$ & $0.033_{\pm 0.001}$  \\
not-MIWAE& $0.006_{\pm 0.001}$ & $0.945_{\pm 0.024}$ & $0.031_{\pm 0.000}$  \\
EGC& $0.006_{\pm 0.000}$ & $1.000_{\pm 0.000}$ & $0.058_{\pm 0.001}$  \\
    \midrule
MaCoDE & $0.007_{\pm 0.003}$ & $0.882_{\pm 0.075}$ & $0.033_{\pm 0.003}$\\
    \bottomrule
    \end{tabular}
    }
\end{minipage}
\caption{\textbf{Q3}: Multiple imputation for each dataset under \textbf{MAR} at 0.3 missingness. The means and standard errors of the mean across 5 datasets and 10 repeated experiments are reported. $\downarrow$ denotes lower is better. $\downarrow$ denotes lower is better.}
\label{tab:MAR_bydata}
\end{table}

\begin{table}[ht]
\centering
\begin{minipage}[t]{0.99\textwidth}
  \centering
  \resizebox{0.45\columnwidth}{!}{
  \begin{tabular}{lrrrrrrrrrrrrrrrr}
    \toprule
Dataset & & \texttt{abalone} &\\
    \midrule
Model & Bias $\downarrow$ & Coverage & Width $\downarrow$ \\
    \midrule	
MICE & $0.007_{\pm 0.001}$ & $0.929_{\pm 0.024}$ & $0.030_{\pm 0.000}$  \\
GAIN& $0.015_{\pm 0.002}$ & $0.629_{\pm 0.061}$ & $0.030_{\pm 0.000}$  \\
missMDA& $0.018_{\pm 0.001}$ & $0.471_{\pm 0.057}$ & $0.034_{\pm 0.001}$  \\ 
VAEAC& $0.005_{\pm 0.001}$ & $0.957_{\pm 0.022}$ & $0.030_{\pm 0.000}$  \\ 
MIWAE& $0.008_{\pm 0.001}$ & $0.900_{\pm 0.043}$ & $0.034_{\pm 0.001}$  \\
not-MIWAE& $0.010_{\pm 0.002}$ & $0.886_{\pm 0.060}$ & $0.035_{\pm 0.003}$  \\
EGC& $0.005_{\pm 0.000}$ & $0.986_{\pm 0.014}$ & $0.039_{\pm 0.001}$  \\
    \midrule
MaCoDE & $0.007_{\pm 0.001}$ & $0.929_{\pm 0.024}$ & $0.035_{\pm 0.001}$\\
    \bottomrule
  \end{tabular}
  }
\end{minipage}

\vspace{3mm}
\begin{minipage}[t]{0.99\textwidth} 
  \centering
  \resizebox{0.45\columnwidth}{!}{
  \begin{tabular}{lrrrrrrrrrrrrrrrr}
    \toprule
Dataset & & \texttt{banknote} &\\
    \midrule
Model & Bias $\downarrow$ & Coverage & Width $\downarrow$ \\
    \midrule	
MICE & $0.017_{\pm 0.001}$ & $0.775_{\pm 0.025}$ & $0.053_{\pm 0.000}$  \\
GAIN& $0.032_{\pm 0.005}$ & $0.625_{\pm 0.077}$ & $0.053_{\pm 0.000}$  \\
missMDA& $0.019_{\pm 0.001}$ & $0.725_{\pm 0.045}$ & $0.056_{\pm 0.000}$  \\
VAEAC& $0.010_{\pm 0.001}$ & $0.975_{\pm 0.025}$ & $0.053_{\pm 0.000}$  \\
MIWAE& $0.005_{\pm 0.001}$ & $1.000_{\pm 0.000}$ & $0.055_{\pm 0.000}$  \\
not-MIWAE& $0.005_{\pm 0.003}$ & $1.000_{\pm 0.000}$ & $0.055_{\pm 0.000}$  \\
EGC& $0.007_{\pm 0.001}$ & $1.000_{\pm 0.000}$ & $0.072_{\pm 0.002}$  \\
    \midrule
MaCoDE & $0.006_{\pm 0.001}$ & $1.000_{\pm 0.000}$ & $0.054_{\pm 0.000}$\\
    \bottomrule
    \end{tabular}
    }
\end{minipage}

\vspace{3mm}

\begin{minipage}[t]{0.99\textwidth}
  \centering
  \resizebox{0.45\columnwidth}{!}{
  \begin{tabular}{lrrrrrrrrrrrrrrrr}
    \toprule
Dataset & & \texttt{breast} &\\
    \midrule
Model & Bias $\downarrow$ & Coverage & Width $\downarrow$ \\
    \midrule	
MICE& $0.008_{\pm 0.000}$ & $1.000_{\pm 0.000}$ & $0.080_{\pm 0.000}$  \\
GAIN& $0.014_{\pm 0.001}$ & $0.960_{\pm 0.011}$ & $0.080_{\pm 0.000}$  \\
missMDA& $0.026_{\pm 0.001}$ & $0.853_{\pm 0.015}$ & $0.086_{\pm 0.000}$  \\
VAEAC& $0.009_{\pm 0.000}$ & $1.000_{\pm 0.000}$ & $0.082_{\pm 0.000}$  \\ 
MIWAE& $0.006_{\pm 0.000}$ & $1.000_{\pm 0.000}$ & $0.084_{\pm 0.000}$  \\
not-MIWAE& $0.007_{\pm 0.002}$ & $0.990_{\pm 0.016}$ & $0.082_{\pm 0.000}$  \\
EGC& $0.007_{\pm 0.000}$ & $1.000_{\pm 0.000}$ & $0.084_{\pm 0.000}$  \\
    \midrule
MaCoDE & $0.009_{\pm 0.000}$ & $1.000_{\pm 0.000}$ & $0.084_{\pm 0.000}$\\
    \bottomrule
  \end{tabular}
  }
\end{minipage}

\vspace{3mm}

\begin{minipage}[t]{0.99\textwidth}
  \centering
  \resizebox{0.45\columnwidth}{!}{
  \begin{tabular}{lrrrrrrrrrrrrrrrr}
    \toprule
Dataset & & \texttt{redwine} &\\
    \midrule
Model & Bias $\downarrow$ & Coverage & Width $\downarrow$ \\
    \midrule	
MICE& $0.021_{\pm 0.001}$ & $0.673_{\pm 0.034}$ & $0.048_{\pm 0.000}$  \\
GAIN& $0.026_{\pm 0.002}$ & $0.473_{\pm 0.057}$ & $0.048_{\pm 0.000}$  \\
missMDA& $0.028_{\pm 0.000}$ & $0.600_{\pm 0.028}$ & $0.055_{\pm 0.000}$  \\
VAEAC& $0.016_{\pm 0.001}$ & $0.782_{\pm 0.036}$ & $0.050_{\pm 0.000}$  \\ 
MIWAE& $0.012_{\pm 0.001}$ & $0.909_{\pm 0.019}$ & $0.055_{\pm 0.001}$  \\
not-MIWAE& $0.013_{\pm 0.001}$ & $0.855_{\pm 0.047}$ & $0.053_{\pm 0.001}$  \\
EGC& $0.009_{\pm 0.000}$ & $1.000_{\pm 0.000}$ & $0.072_{\pm 0.001}$  \\
    \midrule
MaCoDE & $0.011_{\pm 0.001}$ & $0.990_{\pm 0.010}$ & $0.055_{\pm 0.001}$\\
    \bottomrule
  \end{tabular}
  }
\end{minipage}

\vspace{3mm}

\begin{minipage}[t]{0.99\textwidth} 
  \centering
  \resizebox{0.45\columnwidth}{!}{
  \begin{tabular}{lrrrrrrrrrrrrrrrr}
    \toprule
Dataset & & \texttt{whitewine} &\\
    \midrule
Model & Bias $\downarrow$ & Coverage & Width $\downarrow$ \\
    \midrule	
MICE& $0.012_{\pm 0.001}$ & $0.673_{\pm 0.039}$ & $0.028_{\pm 0.000}$  \\
GAIN& $0.029_{\pm 0.004}$ & $0.427_{\pm 0.045}$ & $0.028_{\pm 0.000}$  \\
missMDA& $0.015_{\pm 0.000}$ & $0.582_{\pm 0.034}$ & $0.035_{\pm 0.001}$  \\
VAEAC& $0.012_{\pm 0.001}$ & $0.691_{\pm 0.024}$ & $0.029_{\pm 0.000}$  \\
MIWAE& $0.010_{\pm 0.001}$ & $0.873_{\pm 0.024}$ & $0.036_{\pm 0.001}$  \\
not-MIWAE& $0.008_{\pm 0.001}$ & $0.891_{\pm 0.072}$ & $0.033_{\pm 0.002}$  \\
EGC& $0.008_{\pm 0.000}$ & $0.991_{\pm 0.009}$ & $0.063_{\pm 0.001}$  \\
    \midrule
MaCoDE & $0.009_{\pm 0.001}$ & $0.827_{\pm 0.037}$ & $0.034_{\pm 0.001}$\\
    \bottomrule
    \end{tabular}
    }
\end{minipage}
\caption{\textbf{Q3}: Multiple imputation for each dataset under \textbf{MNARL} at 0.3 missingness. The means and standard errors of the mean across 5 datasets and 10 repeated experiments are reported. $\downarrow$ denotes lower is better.}
\label{tab:MNARL_bydata}
\end{table}

\begin{table}[ht]
\centering
\begin{minipage}[t]{0.99\textwidth}
  \centering
  \resizebox{0.45\columnwidth}{!}{
  \begin{tabular}{lrrrrrrrrrrrrrrrr}
    \toprule
Dataset & & \texttt{abalone} &\\
    \midrule
Model & Bias $\downarrow$ & Coverage & Width $\downarrow$ \\
    \midrule	
MICE& $0.003_{\pm 0.000}$ & $1.000_{\pm 0.000}$ & $0.030_{\pm 0.000}$  \\
GAIN& $0.008_{\pm 0.001}$ & $0.886_{\pm 0.036}$ & $0.031_{\pm 0.001}$  \\
missMDA& $0.009_{\pm 0.000}$ & $0.800_{\pm 0.032}$ & $0.033_{\pm 0.000}$  \\ 
VAEAC& $0.002_{\pm 0.000}$ & $1.000_{\pm 0.000}$ & $0.030_{\pm 0.000}$  \\
MIWAE& $0.005_{\pm 0.001}$ & $0.957_{\pm 0.022}$ & $0.032_{\pm 0.001}$  \\
not-MIWAE& $0.007_{\pm 0.001}$ & $0.943_{\pm 0.023}$ & $0.035_{\pm 0.001}$  \\
EGC& $0.003_{\pm 0.000}$ & $1.000_{\pm 0.000}$ & $0.038_{\pm 0.001}$  \\
    \midrule
MaCoDE & $0.007_{\pm 0.001}$ & $0.886_{\pm 0.019}$ & $0.033_{\pm 0.001}$\\
    \bottomrule
  \end{tabular}
  }
\end{minipage}

\vspace{3mm}
\begin{minipage}[t]{0.99\textwidth} 
  \centering
  \resizebox{0.45\columnwidth}{!}{
  \begin{tabular}{lrrrrrrrrrrrrrrrr}
    \toprule
Dataset & & \texttt{banknote} &\\
    \midrule
Model & Bias $\downarrow$ & Coverage & Width $\downarrow$ \\
    \midrule	
MICE& $0.008_{\pm 0.001}$ & $0.875_{\pm 0.042}$ & $0.052_{\pm 0.000}$  \\
GAIN& $0.013_{\pm 0.002}$ & $0.825_{\pm 0.053}$ & $0.052_{\pm 0.000}$  \\
missMDA& $0.009_{\pm 0.001}$ & $0.775_{\pm 0.045}$ & $0.054_{\pm 0.000}$  \\ 
VAEAC& $0.005_{\pm 0.000}$ & $1.000_{\pm 0.000}$ & $0.053_{\pm 0.000}$  \\
MIWAE& $0.003_{\pm 0.000}$ & $1.000_{\pm 0.000}$ & $0.054_{\pm 0.000}$  \\
not-MIWAE& $0.003_{\pm 0.000}$ & $1.000_{\pm 0.000}$ & $0.054_{\pm 0.000}$  \\
EGC & $0.004_{\pm 0.001}$ & $1.000_{\pm 0.000}$ & $0.066_{\pm 0.001}$  \\
    \midrule
MaCoDE & $0.003_{\pm 0.000}$ & $1.000_{\pm 0.000}$ & $0.053_{\pm 0.000}$\\
    \bottomrule
    \end{tabular}
    }
\end{minipage}

\vspace{3mm}

\begin{minipage}[t]{0.99\textwidth}
  \centering
  \resizebox{0.45\columnwidth}{!}{
  \begin{tabular}{lrrrrrrrrrrrrrrrr}
    \toprule
Dataset & & \texttt{breast} &\\
    \midrule
Model & Bias $\downarrow$ & Coverage & Width $\downarrow$ \\
    \midrule	
MICE& $0.004_{\pm 0.000}$ & $0.997_{\pm 0.003}$ & $0.080_{\pm 0.000}$  \\
GAIN& $0.007_{\pm 0.000}$ & $0.997_{\pm 0.003}$ & $0.080_{\pm 0.000}$  \\
missMDA& $0.017_{\pm 0.001}$ & $0.927_{\pm 0.012}$ & $0.084_{\pm 0.000}$  \\
VAEAC& $0.006_{\pm 0.000}$ & $1.000_{\pm 0.000}$ & $0.082_{\pm 0.000}$  \\
MIWAE& $0.004_{\pm 0.000}$ & $1.000_{\pm 0.000}$ & $0.082_{\pm 0.000}$  \\
not-MIWAE& $0.004_{\pm 0.000}$ & $0.997_{\pm 0.003}$ & $0.081_{\pm 0.000}$  \\
EGC& $0.004_{\pm 0.000}$ & $1.000_{\pm 0.000}$ & $0.082_{\pm 0.000}$  \\
    \midrule
MaCoDE & $0.005_{\pm 0.000}$ & $1.000_{\pm 0.000}$ & $0.083_{\pm 0.000}$\\
    \bottomrule
  \end{tabular}
  }
\end{minipage}

\vspace{3mm}

\begin{minipage}[t]{0.99\textwidth}
  \centering
  \resizebox{0.45\columnwidth}{!}{
  \begin{tabular}{lrrrrrrrrrrrrrrrr}
    \toprule
Dataset & & \texttt{redwine} &\\
    \midrule
Model & Bias $\downarrow$ & Coverage & Width $\downarrow$ \\
    \midrule	
MICE& $0.010_{\pm 0.001}$ & $0.873_{\pm 0.028}$ & $0.048_{\pm 0.000}$  \\
GAIN& $0.011_{\pm 0.001}$ & $0.827_{\pm 0.032}$ & $0.048_{\pm 0.000}$  \\
missMDA& $0.016_{\pm 0.002}$ & $0.755_{\pm 0.024}$ & $0.052_{\pm 0.000}$  \\ 
VAEAC& $0.009_{\pm 0.001}$ & $0.900_{\pm 0.025}$ & $0.050_{\pm 0.000}$  \\
MIWAE& $0.006_{\pm 0.001}$ & $0.955_{\pm 0.020}$ & $0.052_{\pm 0.000}$  \\
not-MIWAE& $0.006_{\pm 0.001}$ & $0.973_{\pm 0.014}$ & $0.050_{\pm 0.000}$  \\
EGC& $0.005_{\pm 0.001}$ & $1.000_{\pm 0.000}$ & $0.063_{\pm 0.001}$  \\
    \midrule
MaCoDE & $0.006_{\pm 0.001}$ & $0.982_{\pm 0.018}$ & $0.053_{\pm 0.000}$\\
    \bottomrule
  \end{tabular}
  }
\end{minipage}

\vspace{3mm}
\begin{minipage}[t]{0.99\textwidth} 
  \centering
  \resizebox{0.45\columnwidth}{!}{
  \begin{tabular}{lrrrrrrrrrrrrrrrr}
    \toprule
Dataset & & \texttt{whitewine} &\\
    \midrule
Model & Bias $\downarrow$ & Coverage & Width $\downarrow$ \\
    \midrule	
MICE& $0.008_{\pm 0.000}$ & $0.773_{\pm 0.020}$ & $0.028_{\pm 0.000}$  \\
GAIN& $0.012_{\pm 0.002}$ & $0.736_{\pm 0.029}$ & $0.028_{\pm 0.000}$  \\
missMDA& $0.010_{\pm 0.000}$ & $0.809_{\pm 0.021}$ & $0.033_{\pm 0.001}$  \\
VAEAC& $0.009_{\pm 0.001}$ & $0.736_{\pm 0.029}$ & $0.028_{\pm 0.000}$  \\ 
MIWAE& $0.006_{\pm 0.001}$ & $0.964_{\pm 0.020}$ & $0.033_{\pm 0.001}$  \\
not-MIWAE& $0.006_{\pm 0.000}$ & $0.918_{\pm 0.025}$ & $0.031_{\pm 0.001}$  \\  
EGC& $0.004_{\pm 0.000}$ & $1.000_{\pm 0.000}$ & $0.055_{\pm 0.001}$  \\
    \midrule
MaCoDE & $0.006_{\pm 0.001}$ & $0.855_{\pm 0.031}$ & $0.032_{\pm 0.001}$\\
    \bottomrule
    \end{tabular}
    }
\end{minipage}
\caption{\textbf{Q3}: Multiple imputation for each dataset under \textbf{MNARQ} at 0.3 missingness. The means and standard errors of the mean across 5 datasets and 10 repeated experiments are reported. $\downarrow$ denotes lower is better.}
\label{tab:MNARQ_bydata}
\end{table}

\end{document}